\documentclass[10pt,twocolumn,letterpaper]{article}

\usepackage[pagenumbers]{cvpr} 

\usepackage{graphicx}
\usepackage{multirow}
\usepackage{amsmath}
\usepackage{floatrow}
\usepackage{wrapfig}
\usepackage[ruled,linesnumbered]{algorithm2e}
\usepackage{algpseudocode}
\usepackage{wrapfig}
\usepackage{booktabs}
\usepackage{enumitem}
\usepackage[accsupp]{axessibility}

\usepackage[dvipsnames]{xcolor}

\definecolor{cvprblue}{rgb}{0.21,0.49,0.74}
\usepackage[pagebackref,breaklinks,colorlinks,citecolor=cvprblue]{hyperref}

\def\paperID{9969} 
\def\confName{CVPR}
\def\confYear{2024}

\title{Residual Denoising Diffusion Models}

\author{%
  Jiawei Liu$^{1,2,3}$, Qiang Wang$^{1,4}$, Huijie Fan$^{1,2}$\footnotemark[1], Yinong Wang$^5$, Yandong Tang$^{1,2}$, Liangqiong Qu$^{5}$\footnotemark[1] \\
  $^1$State Key Laboratory of Robotics, Shenyang Institute of Automation, Chinese Academy of Sciences\\
  $^2$Institutes for Robotics and Intelligent Manufacturing, Chinese Academy of
  Science\\
  $^3$University of Chinese Academy of Sciences
  $^4$Shenyang University
  $^5$The University of Hong Kong\\
  \texttt{\{liujiawei,wangqiang,fanhuijie,ytang\}@sia.cn, liangqqu@hku.hk} \\
}

\begin{document}
\maketitle
\begin{abstract}
    We propose residual denoising diffusion models (RDDM), a novel dual diffusion process that decouples the traditional single denoising diffusion process into residual diffusion and noise diffusion. This dual diffusion framework expands the denoising-based diffusion models, initially uninterpretable for image restoration, into a unified and interpretable model for both image generation and restoration by introducing residuals. Specifically, our residual diffusion represents directional diffusion from the target image to the degraded input image and explicitly guides the reverse generation process for image restoration, while noise diffusion represents random perturbations in the diffusion process.  The residual prioritizes certainty, while the noise emphasizes diversity, enabling RDDM to effectively unify tasks with varying certainty or diversity requirements, such as image generation and restoration. We demonstrate that our sampling process is consistent with that of DDPM and DDIM through coefficient transformation, and propose a partially path-independent generation process to better understand the reverse process. Notably, our RDDM enables a generic UNet, trained with only an L1 loss and a batch size of 1, to compete with state-of-the-art image restoration methods. We provide code and pre-trained models to encourage further exploration, application, and development of our innovative framework (\href{https://github.com/nachifur/RDDM}{https://github.com/nachifur/RDDM}).
\end{abstract}    
\section{Introduction}\label{Introduction}
\renewcommand{\thefootnote}{\fnsymbol{footnote}}
\footnotetext[1]{Corresponding author.}
\renewcommand{\thefootnote}{\arabic{footnote}}

\begin{figure}[t]
    \setlength{\abovecaptionskip}{0.0cm}
    \centering
    \includegraphics[width=1\linewidth]{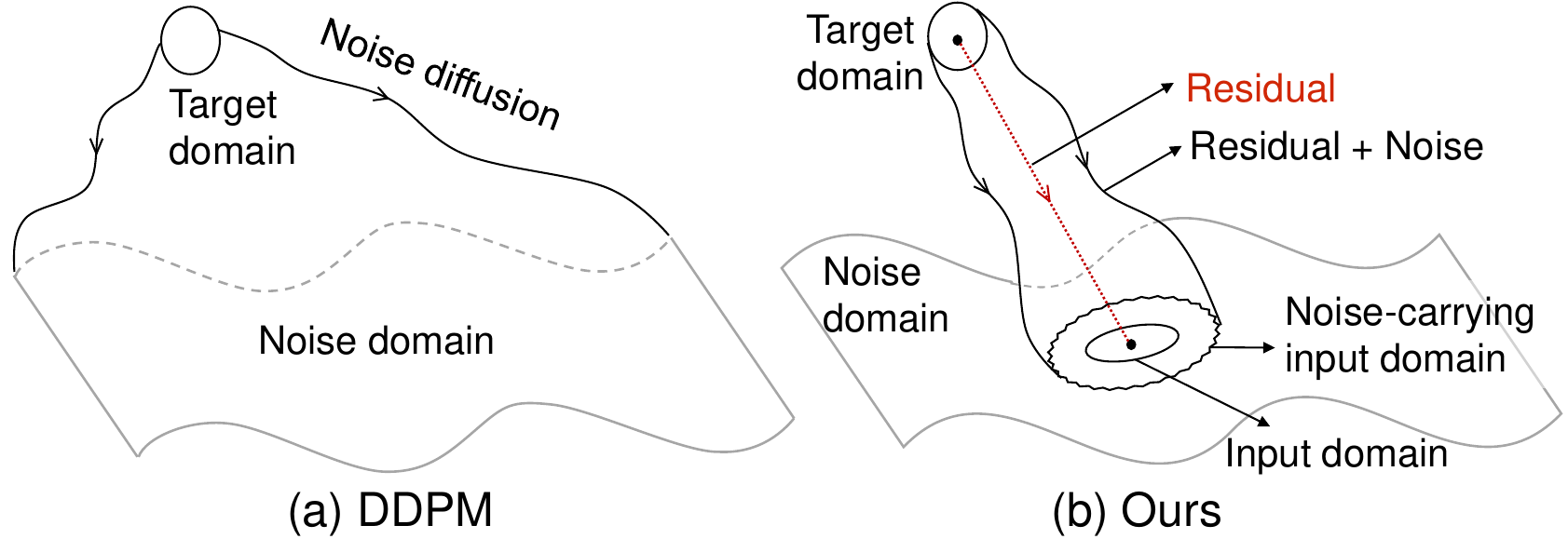}
    \caption{Denoising diffusion process - DDPM~\citep{ho2020denoising} (a) and our residual denoising diffusion process (b). For image restoration, we introduce residual diffusion to represent the diffusion direction from the target image to the input image.\vspace{-0.4cm}}
    \label{fig:2}
 \end{figure}

In real-life scenarios, diffusion often occurs in complex forms involving multiple, concurrent processes, such as the dispersion of multiple gases or the propagation of different types of waves or fields. This leads us to ponder whether the denoising-based diffusion models~\citep{ho2020denoising,song2020denoising} have limitations in focusing solely on denoising.
Current diffusion-based image restoration methods~\citep{lugmayr2022repaint,saharia2022image,rombach2022high,jin2022shadowdiffusion,ozdenizci2023restoring} extend the diffusion model to image restoration tasks by using degraded images as a condition input to implicitly guide the reverse generation process, without modifying the original denoising diffusion process~\citep{ho2020denoising,song2020denoising}.
However, the reverse process starting from noise seems to be unnecessary, as the degraded image is already known. The forward process is non-interpretability for image restoration, as the diffusion process does not contain any information about the degraded image, as shown in Fig.~\ref{fig:2}(a).

In this paper, we explore a novel dual diffusion process and propose Residual Denoising Diffusion Models (RDDM), which can tackle the non-interpretability of a single denoising process for image restoration.
In RDDM,  we decouple the previous diffusion process into residual diffusion and noise diffusion.
Residual diffusion prioritizes certainty and represents a directional diffusion from the target image to the conditional input image, and noise diffusion emphasizes diversity and represents random perturbations in the diffusion process.
Thus, our RDDM can unify different tasks that require different certainty or diversity, e.g., image generation and restoration.
Compared to denoising-based diffusion models for image restoration, the residuals in RDDM clearly indicate the forward diffusion direction and explicitly guide the reverse generation process for image restoration, as shown in Fig.~\ref{fig:2}(b).

\begin{figure}[t]
    \setlength{\abovecaptionskip}{0.0cm}
    \centering
    \includegraphics[width=0.7\linewidth]{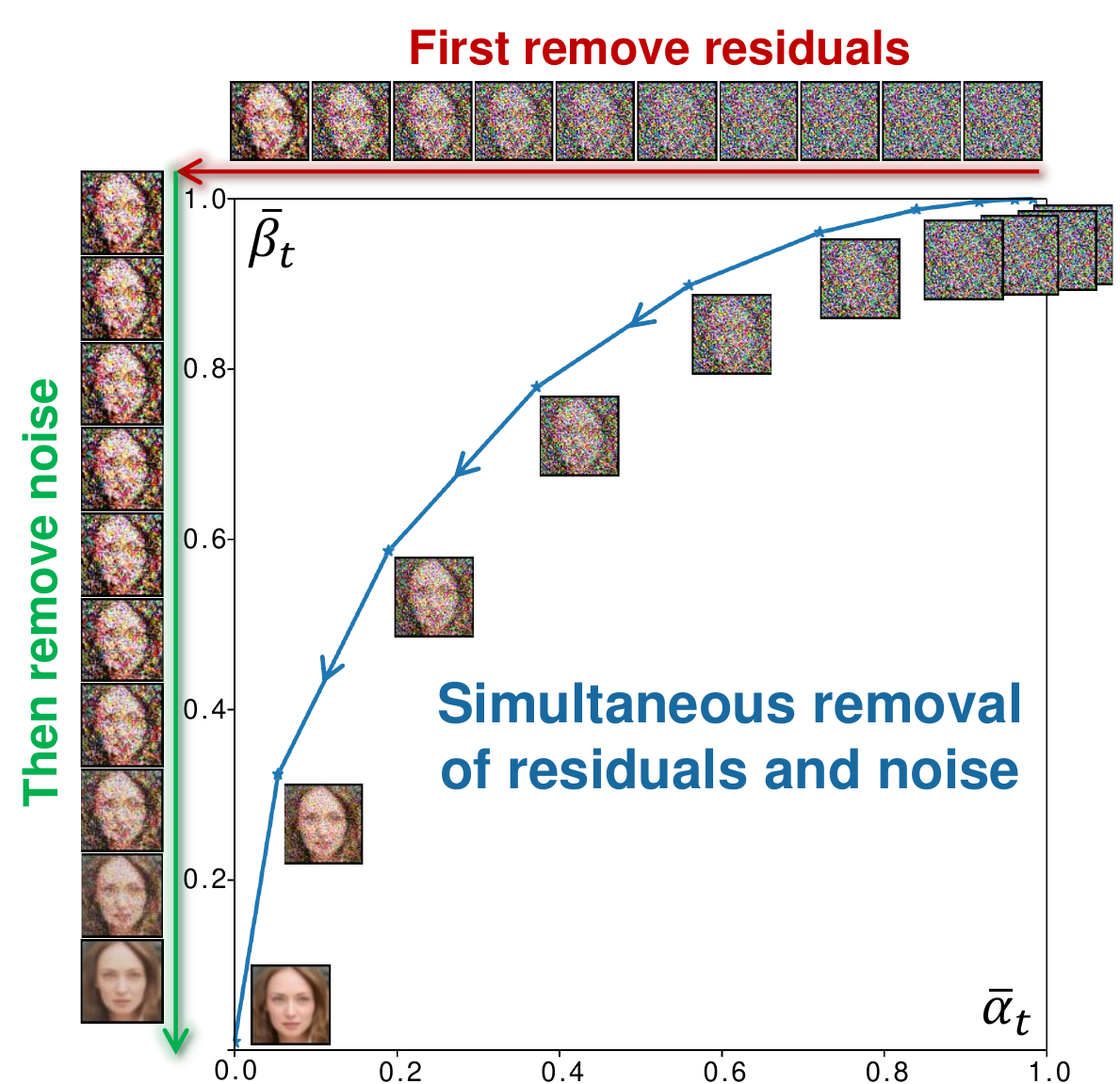}
    \caption{Decoupled dual diffusion framework. The previous forward diffusion process is decoupled into residual diffusion and noise diffusion, while in the reverse process, the simultaneous sampling can be decoupled into first removing the residuals and then removing noise. \vspace{-0.4cm}}
    \label{fig:decouple}
\end{figure}

Specifically, we redefine a new forward process that allows simultaneous diffusion of residuals and noise, wherein the target image progressively diffuses into a purely noisy image for image generation or a noise-carrying input image for image restoration. Unlike the previous denoising diffusion model~\citep{ho2020denoising,song2020denoising}, which uses one coefficient schedule to control the mixing ratio of noise and images, our RDDM employs two independent coefficient schedules to control the diffusion speed of residuals and noise.
We found that this independent diffusion property is also evident in the reverse generation process, e.g., readjusting the coefficient schedule within a certain range during testing does not affect the image generation results, and removing the residuals firstly, followed by denoising (see Fig.~\ref{fig:decouple}), can also produce semantically consistent images.
Our RDDM is compatible with widely used denoising diffusion models, i.e., our sampling process is consistent with that of DDPM~\citep{ho2020denoising} and DDIM~\citep{song2020denoising} by transforming coefficient schedules. In addition, our RDDM natively supports conditional inputs, enabling networks trained with only an $\ell _1$ loss and a batch size of 1 to compete with state-of-the-art image restoration methods.
We envision that our models can facilitate a unified and interpretable image-to-image distribution transformation methodology, highlighting that residuals and noise are equally important for diffusion models, e.g., the residual prioritizes certainty while the noise emphasizes diversity.
The contributions of this paper are summarized as follows:
\begin{itemize}
    \item We propose a novel dual diffusion framework to tackle the non-interpretability of a single denoising process for image restoration by introducing residuals. Our residual diffusion represents a directional diffusion from the target image to the conditional input image.
    \item We introduce a partially path-independent generation process that decouples residuals and noise, highlighting their roles in controlling directional residual shift (certainty) and random perturbation (diversity), respectively.
    \item We design an automatic objective selection algorithm to choose whether to predict residuals or noise for unknown new tasks. 
    \item Extensive experiments demonstrate that our method can be adapted to different tasks, e.g., image generation, restoration, inpainting and translation, focusing certainty or diversity, and involving paired or unpaired data.
\end{itemize}

\vspace{-0.3cm}
\section{Related Work} \label{Sec:6.0}

Denoising diffusion models (e.g., DDPM~\citep{ho2020denoising}, SGM~\citep{song2019generative,song2020score}, and DDIM~\citep{song2020denoising}) were initially developed for image generation. Subsequent image restoration methods~\citep{lugmayr2022repaint,rombach2022high,guo2023shadowdiffusion} based on DDPM and DDIM feed a degraded image as a conditional input to a denoising network, e.g., DvSR~\citep{whang2022deblurring}, SR3~\citep{saharia2022image}, and WeatherDiffusion~\citep{ozdenizci2023restoring}, which typically require large sampling steps and batch sizes. Additionally, the reverse process starting from noise in these methods seems unnecessary and inefficient for image restoration tasks. Thus, SDEdit~\citep{meng2021sdedit}, ColdDiffusion~\citep{bansal2022cold}, InDI~\citep{delbracio2023inversion}, and I2SB~\citep{liu20232} propose generating a clear image directly from a degraded image or noise-carrying degraded image. InDI~\citep{delbracio2023inversion} and I2SB~\citep{liu20232}, which also present unified image generation and restoration frameworks, are the most closely related to our proposed RDDM.
Specifically, the forward diffusion of InDI, I2SB, and our RDDM consistently employs a mixture of three terms (i.e., input images $I_{in}$, target images $I_0$, and noise $\epsilon$), extending beyond the denoising-based diffusion model~\citep{ho2020denoising,song2020denoising} which incorporates a mixture of two terms (i.e., $I_0$ and $\epsilon$). However, InDI~\citep{delbracio2023inversion} and I2SB~\citep{liu20232} opt for estimating the target image or its linear transformation term to replace the noise estimation, akin to a special case of our RDDM (SM-Res). In contrast, we introduce residual estimation while also embracing noise for both generation and restoration tasks. Our RDDM can further extend DDPM~\citep{ho2020denoising}, DDIM~\citep{song2020denoising}, InDI~\citep{delbracio2023inversion}, and I2SB~\citep{liu20232} to independent double diffusion processes, and pave the way for the multi-dimensional diffusion process.
We highlight that residuals and noise are equally important, e.g., the residual prioritizes certainty while the noise emphasizes diversity.
In addition, our work is related to coefficient schedule design~\citep{rombach2022high,nichol2021improved}, variance strategy optimization~\citep{kingma2021variational,nichol2021improved,bao2022analytic,bao2022estimating}, superimposed image decomposition~\citep{zou2020deep,duan2022develop}, curve integration~\citep{riley_hobson_bence_2006}, stochastic differential equations~\citep{song2020score}, and residual learning~\citep{he2016deep} for image restoration~\citep{zhang2017beyond,zhang2020residual,anwar2020densely,zamir2021multi,tu2022maxim,liu2023shadow}. See Appendix~\ref{Appendix:a.5} for detailed comparison.


\begin{figure*}[t]
    \setlength{\abovecaptionskip}{0.0cm}
    \centering
    \includegraphics[width=0.7\linewidth]{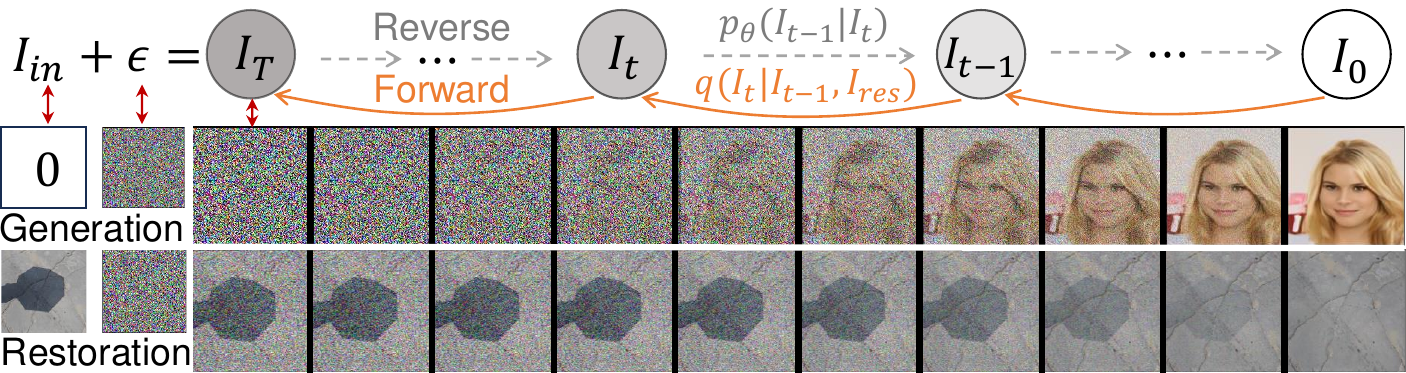}
    \caption{
    The proposed residual denoising diffusion model (RDDM) is a unified framework for image generation and restoration (a shadow removal task is shown here). We introduce residuals ($I_{res}$) in RDDM, redefining the forward diffusion process to involve simultaneous diffusion of residuals and noise. The residuals ($I_{res}=I_{in}-I_0$) diffusion represents the directional diffusion from the target image $I_0$ to the degraded input image  $I_{in}$, while the noise ($\epsilon$) diffusion  represents the random perturbations in the diffusion process.
    In RDDM, $I_0$ gradually diffuses into $I_T=I_{in}+\epsilon$, $\epsilon \sim \mathcal{N} (\mathbf{0} ,\mathbf{I})$. In the third columns, $I_T$ is a purely noisy image for image generation since $I_{in}=0$, and a noise-carrying degraded image for image restoration as $I_{in}$ is the degraded image.\vspace{-0.4cm}}
    \label{fig:1}
\end{figure*}

\section{Background}\label{Background}
Denoising diffusion models~\citep{ho2020denoising,sohl2015deep} aim to  learn a distribution $p_\theta (I_0):=\int p_\theta (I_{0:T})dI_{1:T}$\protect\footnotemark[1]\footnotetext[1]{To understand diffusion from an image perspective, we use $I$ instead of $x$ in DDPM~\citep{ho2020denoising}.} to approximate a target data distribution $q(I_0)$, where $I_0$ are target images and $I_1, \dots , I_T$ ($T=1000$) are latent images of the same dimension as $I_0$. In the forward process, $q(I_0)$ is diffused into a Gaussian noise distribution using a fixed Markov chain,
\begin{align}
    q(I _{1:T}|I _{0}) & := {\textstyle \prod_{t=1}^{T}} q(I _t|I _{t-1}),                                 \\
    q(I_t|I_{t-1})     & :=\mathcal{N}(I_t;\sqrt{\alpha _t}I_{t-1},(1-\alpha _t) \mathbf{I} ),\label{Eq:1}
\end{align}
where $\alpha _{1:T}\in (0,1]^T$. $q(I_t|I_{t-1})$ can also be written as $I_t=\sqrt{\alpha  _t}I_{t-1}+\sqrt{1-\alpha _t}\epsilon_{t-1}$. In fact, it is simpler to sampling $I_t$ from $I_0$ by reparameterization~\citep{Kingma2014,Diederik2019},
\begin{align}
    I_t=\sqrt{\bar{\alpha}  _t}I_{0}+\sqrt{1-\bar{\alpha} _t}\epsilon,\label{Eq:4}
\end{align}
where $\epsilon \sim \mathcal{N}(\mathbf{0},\mathbf{I}), \bar{\alpha} _t:= {\textstyle \prod_{s=1}^{t}}\alpha_s$.
The reverse process is also a Markov chain starting at $p_\theta(I_T)\sim \mathcal{N}(I_T;\mathbf{0},\mathbf{I})$,
\begin{align}
    p_\theta (I_{0:T})    & :=p_\theta(I_T) {\textstyle \prod_{t=1}^{T}}p_\theta(I_{t-1}|I_t),           \\
    p_\theta(I_{t-1}|I_t) & :=\mathcal{N} (I_{t-1};\mu _\theta (I_t,t),\Sigma_t\mathbf{I}), \label{Eq:3}
\end{align}
where $p_\theta(I_{t-1}|I_t)$ is a learnable transfer probability (the variance schedule $\Sigma_t$ is fixed). A simplified loss function~\cite{ho2020denoising} is derived from the maximum likelihood of $p_\theta (I_0)$, i.e., $L(\theta ):=\mathbb{E} _{I_0\sim q(I_0),\epsilon \sim \mathcal{N} (\mathbf{0} ,\mathbf{I} )}\left [ \left \| \epsilon-\epsilon _{\theta }(I_t,t) \right \|^2  \right ]$. The estimated noise $\epsilon _{\theta }$ can be used to represent $\mu _\theta$ in $p_\theta(I_{t-1}|I_t)$, thus $I_{t-1}$ can be sampled from $p_\theta(I_{t-1}|I_t)$ step by step.


\section{Residual Denoising Diffusion Models}\label{Residual}
Our goal is to develop a dual diffusion process to unify and interpret image generation and restoration. We modify the representation of $I_T=\epsilon$ in traditional DDPM to $I_T=I_{in}+\epsilon$ in our RDDM, where $I_{in}$ is a degraded image (e.g., a shadow, low-light, or blurred image) for image restoration and is set to $0$ for image generation. This modification is compatible with the widely used denoising diffusion model, e.g., $I_T = 0 + \epsilon $ is the pure noise ($\epsilon$) for generation. For image restoration, $I_T$ is a noisy-carrying degraded image ($I_{in}+\epsilon$), as shown in the third column in Fig.~\ref{fig:1}. The modified forward process from $I_0$ to $I_T=I_{in}+\epsilon$ involves progressively degrading $I_0$ to $I_{in}$, and injecting noise $\epsilon$.
This naturally results in a dual diffusion process, a residual diffusion to model the transition from $I_0$ to $I_{in}$ and a noise diffusion. For example, the forward diffusion process from the shadow-free image $I_0$ to the noisy carrying shadow image $I_T$ involves progressively adding shadows and noise, as shown in the second row in Fig.~\ref{fig:1}.

In the following subsections, we detail the underlying theory and the methodology behind our RDDM. Inspired by residual learning~\citep{he2016deep,liu2023decoupled,liu2023shadow}, we redefine each forward diffusion process step in Section ~\ref{Sec:3.1}.
For the reverse process, we present a training objective to predict the residuals and noise injected in the forward process in Section ~\ref{Sec:3.2}.
In Section ~\ref{Sec:3.3}, we propose three sampling methods, i.e., residual prediction (SM-Res), noise prediction (SM-N), and ``residual and noise prediction'' (SM-Res-N).

\subsection{Directional Residual Diffusion Process with Perturbation}\label{Sec:3.1}
To model the gradual degradation of image quality and the increment of noise, we define the single forward process step in our RDDM as follows:
\begin{align}
    I_t & =I_{t-1}+I_{res}^t, \qquad I_{res}^t\sim \mathcal{N} (\alpha _tI_{res},\beta ^2_t\mathbf{I} ), \label{Eq:7}
\end{align}
where $I_{res}^t$ represents a directional mean shift (residual diffusion) with random perturbation (noise diffusion) from state $I_{t-1}$ to state $I_{t}$,
the residuals $I_{res}$ in $I_{res}^t$ is the difference between $I_{in}$ and $I_{0}$ (i.e., $I_{res}=I_{in}-I_{0}$), and two independent coefficient schedules $\alpha _t$ and $\beta _t$ control the residual and noise diffusion, respectively. In fact, it is simpler to sample $I_t$ from $I_0$ (like Eq.~\ref{Eq:4}),
\begin{align}
    \begin{split}
        I_t=&I_{t-1}+\alpha_t I_{res}+\beta_t \epsilon_{t-1},  \\
        =&I_{t-2}+(\alpha_{t-1}+\alpha_{t}) I_{res}+(\sqrt{\beta_{t-1}^2+\beta_t^2}) \epsilon_{t-2}\\
        =&\dots\\
        =&I_0+\bar{\alpha}_t I_{res}+\bar{\beta}_t \epsilon, \label{Eq:8}
    \end{split}
\end{align}
where $\epsilon_{t-1},\dots \epsilon \sim \mathcal{N}(\mathbf{0},\mathbf{I})$, $\bar{\alpha}_t= {\textstyle \sum_{i=1}^{t}} \alpha_i$ and $\bar{\beta}_t=  \sqrt{{\textstyle \sum_{i=1}^{t}}\beta_i^2 }$.
If $t=T$, $\bar{\alpha}_T=1$ and $I_T=I_{in}+\bar{\beta}_T \epsilon$. $\bar{\beta}_T$ can control the intensity of noise perturbation for image restoration (e.g., $\bar{\beta}_T^2=0.01$ for shadow removal), while $\bar{\beta}_T^2=1$ for image generation.
From Eq.~\ref{Eq:7}, the joint probability distributions in the forward process can be defined as:
\begin{align}
    q(I _{1:T}|I _{0},I _{res}) & := {\textstyle \prod_{t=1}^{T}} q(I _t|I _{t-1},I _{res}),                        \\
    q(I_t|I_{t-1},I_{res})      & :=\mathcal{N} (I_{t};I_{t-1}+\alpha_t I_{res},\beta_t^2\mathbf{I}  ).\label{Eq:9}
\end{align}
Eq.~\ref{Eq:8} defines the marginal probability distribution $q(I_t|I_0,I_{res})=\mathcal{N} (I_{t};I_0+\bar{\alpha }_t I_{res},\bar{\beta }_t^2\mathbf{I})$. In fact, the forward diffusion of our RDDM is a mixture of three terms (i.e., $I_0$, $I_{res}$, and $\epsilon$), extending beyond the widely used denoising diffusion model that is a mixture of two terms, i.e, $I_0$ and $\epsilon$. A similar mixture form of three terms can be seen in several concurrent works, e.g., InDI~\citep{delbracio2023inversion}, I2SB~\citep{liu20232}, IR-SDE~\citep{luo2023image}, and ResShift~\citep{yue2023resshift}.
\begin{table*}[t]
    \resizebox{0.9\columnwidth}{!}{
        \setlength{\tabcolsep}{1.5mm}{
            \begin{tabular}{l|cc|ccc|cc|cc}
                \toprule
                \multirow{2}*{Sampling Method} & \multicolumn{2}{c|}{Generation (CelebA)} & \multicolumn{3}{c|}{Shadow removal (ISTD)} & \multicolumn{2}{c}{Low-light (LOL)} & \multicolumn{2}{c}{Deraining (RainDrop)}                                                                                                      \\
                                               & FID ($\downarrow $)                      & IS ($\uparrow $)                           & MAE($\downarrow $)                  & PSNR($\uparrow $)                        & SSIM($\uparrow $) & PSNR($\uparrow $) & SSIM($\uparrow $) & PSNR($\uparrow $) & SSIM($\uparrow $)  \\
                \midrule
                SM-Res                         & 31.47                                    & 1.73                                       & \underline{4.76}                    & \underline{30.72}                        & \underline{0.959} & {\bf 25.39}       & {\bf 0.937}       & \underline{31.96} & \underline{0.9509}
                \\
                SM-N                           & {\bf 23.25}                              & {\bf 2.05}                                 & 81.01                               & 11.34                                    & 0.175             & 16.30             & 0.649             &  19.15             & 0.7179 \\
                SM-Res-N                       & \underline{28.90}                        & \underline{1.78}                           & {\bf 4.67}                          & {\bf 30.91}                              & {\bf 0.962}       & \underline{23.90} & \underline{0.931} & {\bf 32.51}       & {\bf 0.9563}       \\
                \bottomrule
            \end{tabular}
        }}
    \setlength{\belowcaptionskip}{0.0cm}
    \caption{Sampling method analysis. The sampling steps are 10 on the CelebA 64 $\times$ 64~\citep{liu2015faceattributes} dataset, 5 on the ISTD~\citep{wang2018stacked} dataset, 2 on the LOL~\citep{wei2018deep} dataset, and 5 on the RainDrop~\citep{qian2018attentive} dataset.\vspace{-0.4cm}}
    \label{table:7}
    \vspace{-0.3cm}
\end{table*}

\subsection{Generation Process and Training Objective} \label{Sec:3.2}
In the forward process (Eq.~\ref{Eq:8}), residuals ($I_{res}$) and noise ($\epsilon$) are gradually added to $I_{0}$, and then synthesized into $I_t$, while the reverse process from $I_T$ to $I_{0}$ involves the estimation of the residuals and noise injected in the forward process. We can train a residual network $ I^{\theta }_{res}(I_t,t,I_{in})$ to predict $I_{res}$ and a noise network $\epsilon _{\theta }(I_t,t,I_{in})$ to estimate $\epsilon$. Using Eq.~\ref{Eq:8}, we obtain the estimated target images $I_0^\theta=I_t-\bar{\alpha}_t I_{res}^\theta-\bar{\beta}_t \epsilon_\theta$. If $I_0^\theta$ and $I_{res}^\theta$ are given, the generation process is defined as,
\begin{align}
    p_\theta(I_{t-1}|I_t):=q_\sigma (I_{t-1}|I_t,I_0^\theta,I_{res}^\theta),\label{Eq:12}
\end{align}
where the transfer probability $q_\sigma (I_{t-1}|I_t,I_0,I_{res})$\protect\footnotemark[2]\footnotetext[2]{Eq.~\ref{Eq:11} does not change $q(I_t|I_0,I_{res})$ in Appendix~\ref{Appendix:a.2}.} from $I_t$ to $I_{t-1}$ is,
\begin{align}
    \begin{split}
        &q_\sigma (I_{t-1}|I_t,I_0,I_{res})=\mathcal{N} (I_{t-1};I_0+\bar{\alpha }_{t-1}I_{res}\\&+\sqrt{\bar{\beta }_{t-1}^2 -\sigma _t^2} \frac{I_t-(I_0+\bar{\alpha }_{t}I_{res})}{\bar{\beta }_{t}} ,\sigma _t^2\mathbf{I}  ),\label{Eq:11}
    \end{split}
\end{align}
where $\sigma _t^2=\eta \beta_t^2\bar{\beta} _{t-1}^2/\bar{\beta} _{t}^2$ and $\eta$ controls whether the generation process is random ($\eta=1$) or deterministic ($\eta=0$).
Using Eq.~\ref{Eq:12} and Eq.~\ref{Eq:11}, $I_{t-1}$ can be sampled from $I_t$ via:
\begin{align}
    \begin{split}
        I_{t-1}&=I_t-(\bar{\alpha} _t-\bar{\alpha} _{t-1})I_{res}^\theta \\&-(\bar{\beta }_t-\sqrt{\bar{\beta }_{t-1}^2-\sigma _t^2}  )\epsilon _\theta +\sigma _t\epsilon _t, \label{Eq:13}
    \end{split}
\end{align}
where $\epsilon_{t} \sim \mathcal{N}(\mathbf{0},\mathbf{I})$. When $\eta =1$, our RDDM has the sum-constrained variance, while DDPM has preserving variance (see Appendix~\ref{Appendix:a.4}). When $\eta =0$ (i.e., $\sigma _t=0$), the sampling process is deterministic,
\begin{align}
    I_{t-1}=I_t-(\bar{\alpha} _t-\bar{\alpha} _{t-1})I_{res}^\theta -(\bar{\beta }_t-\bar{\beta }_{t-1} )\epsilon _\theta. \label{Eq:15}
\end{align}
We derive the following simplified loss function for training (Appendix~\ref{Appendix:a.1}):
\begin{align}
    L_{res}(\theta ) & :=\mathbb{E} \left [ \lambda _{res}\left \| I_{res}-I_{res}^{\theta }(I_t,t,I_{in}) \right \|^2 \right ],\label{Eq:16-1}\\
    L_{\epsilon}(\theta )&:=\mathbb{E} \left [ \lambda _{\epsilon}\left \| \epsilon-\epsilon _{\theta }(I_t,t,I_{in}) \right \|^2 \right ],\label{Eq:16}
\end{align}
where the hyperparameters $\lambda _{res}$, $\lambda _{\epsilon}$ $\in\{0,1\}$, and the training input image $I_t$ is synthesized using $I_0$, $I_{res}$, and $\epsilon$ by Eq.~\ref{Eq:8}. $I_t$ can also be synthesized using $I_{in}$ (replace $I_0$ in Eq.~\ref{Eq:8} by $I_0=I_{in}-I_{res}$),
\begin{align}
    I_t=I_{in}+(\bar{\alpha}_t-1) I_{res}+\bar{\beta}_t\epsilon.\label{Eq:17}
\end{align}

\subsection{Sampling Method Selection Strategies}\label{Sec:3.3}



For the generation process (from $I_t$ to $I_{t-1}$), $I_t$ and $I_{in}$ are known, and thus $I_{res}$ and $\epsilon$ can represent each other by Eq.~\ref{Eq:17}. From Eq.~\ref{Eq:16-1}, \ref{Eq:16}, \ref{Eq:17}, we propose three sampling methods as follows.\\
{\bf SM-Res.} When $\lambda _{res}=1$ and $\lambda _{\epsilon}=0$, the residuals $I_{res}^\theta$ are predicted by a network, while the noise $\epsilon_\theta$ is represented as a transformation of $I_{res}^\theta$ using Eq.~\ref{Eq:17}.\\
{\bf SM-N.} When $\lambda _{res}=0$ and $\lambda _{\epsilon}=1$, the noise $\epsilon_\theta$ is predicted by a network, while the residuals $I_{res}^\theta$ are represented as a transformation of $\epsilon_\theta$ using Eq.~\ref{Eq:17}.\\
{\bf SM-Res-N.} When $\lambda _{res}=1$ and $\lambda _{\epsilon}=1$, both the residuals and the noise are predicted by networks.\\
To determine the optimal sampling method for real-world applications, we give empirical strategies and automatic selection algorithms in the following.

    {\bf Empirical Research.} Table~\ref{table:7} presents that the SM-Res shows better results for image restoration but offers a poorer FID for generation.
On the other hand, the SM-N yields better frechet inception distance (FID in~\cite{heusel2017gans}) and inception scores (IS), but is ineffective in image restoration (e.g., PSNR 11.34 for shadow and 16.30 for low-light). This may be due to the inadequacy of using $\epsilon_\theta$ to represent $I_{res}^\theta$ in Eq.~\ref{Eq:17} for restoration tasks. We attribute these inconsistent results to the fact that {\bf residual predictions prioritize certainty, whereas noise predictions emphasize diversity}.
In our experiments, we use SM-N for image generation, SM-Res for low-light (LOL~\citep{wei2018deep}), and SM-Res-N for other image restoration tasks. For an unknown new task, we empirically recommend using SM-N for those requiring greater diversity and SM-Res for tasks that demand higher certainty.

    {\bf Automatic Objective Selection Algorithm (AOSA).} 
To automatically choose between SM-Res or SM-N for an unknown task, we develop an automatic sampling selection algorithm in Appendix~\ref{Appendix:b.3}. This algorithm requires only a single network and learns the hyperparameter in Eq.~\ref{Eq:16}, enabling a gradual transition from combined residual and noise training (akin to SM-Res-N) to individual prediction (SM-Res or SM-N). This plug-and-play training strategy requires less than 1000 additional training iterations and is fully compatible with the current denoising-based diffusion methods~\citep{ho2020denoising}. Our RDDM using AOSA has the potential to provide a unified and interpretable methodology for modeling, training, and inference pipelines for unknown target tasks.

    {\bf Comparison with Other Prediction Methods.}
Our SM-N is similar to DDIM~\citep{song2020denoising} (or DDPM~\citep{ho2020denoising}), which only estimates the noise, and is consistent with DDPM and DDIM by transforming the coefficient/variance schedules in Eq.~\ref{Eq:13} (the proof in Appendix~\ref{Appendix:a.3}),\footnotetext[3]{$\bar{\alpha } _{DDIM}^t$ here is $\alpha _t$ of DDIM~\citep{song2020denoising}.}
\begin{align}
    \begin{split}
        &\bar{\alpha } _t=1-\sqrt{\bar{\alpha } _{DDIM}^t}\,\protect\footnotemark[3], \bar{\beta } _t=\sqrt{1-\bar{\alpha } _{DDIM}^t}, \\&\sigma _t^2=\sigma _t^2(DDIM). \label{Eq:19}
    \end{split}
\end{align}

In fact, current research has delved into numerous diffusion forms that extend beyond noise estimation.
For example, IDDPM~\citep{nichol2021improved} proposes that it is feasible to estimate noise ($\epsilon_\theta$), clean target images ($I_0^\theta$), or the mean term ($\mu _\theta$) to represent the transfer probabilities (i.e., $p_\theta(I_{t-1}|I_t)$ in Eq.~\ref{Eq:3}).
The score-based generative model (SGM)~\citep{song2019generative} and Schr\" {o}dinger Bridge (I2SB~\citep{liu20232}) estimate the score of noisy data (i.e., the sum of residuals and noise ${\textstyle \sum_{i=1}^{t}} I_{res}^t$). ColdDiffusion~\citep{bansal2022cold} and InDI~\citep{delbracio2023inversion} estimate the clean target images ($I_0$). Rectified Flow~\citep{liu2023flow} predicts the residuals ($I_{res}$) to align with the image linear interpolation process without noise diffusion (i.e., $I_T=I_{in}$). A detailed comparison can be found in Appendix~\ref{Appendix:a.5}.

These previous/concurrent works choose to estimate the noise, the residual, the target image, or its linear transformation term. In contrast, we introduce residual estimation while also embracing noise for both generation and restoration. Residuals and noise have equal and independent status, which is reflected in the forward process (Eq.~\ref{Eq:8}), the reverse process (Eq.~\ref{Eq:15}), and the loss function (Eq.~\ref{Eq:16}). This independence means that the noise diffusion can even be removed and only the residual diffusion retained to model the image interpolation process (when $\bar{\beta}_T= 0$ in Eq.~\ref{Eq:8}, RDDM degenerates to Rectified Flow~\citep{liu2023flow}). In addition, this property derives a decoupled dual diffusion framework in Section ~\ref{Sec:4.0}.



\section{Decoupled Dual Diffusion Framework} \label{Sec:4.0}
Upon examining DDPM from the perspective of RDDM, we discover that DDPM indeed involves the simultaneous diffusion of residuals and noise, which is evident as Eq.~\ref{Eq:49} becomes equivalent to Eq.~\ref{Eq:45} in Appendix~\ref{Appendix:a.3}.
We find that it is possible to decouple these two types of diffusion.
Section ~\ref{Sec:4.1} presents a decoupled forward diffusion process.
In Section ~\ref{Sec:4.2}, we propose a partially path-independent generation process and decouple the simultaneous sampling into first removing the residuals and then removing noise (see Fig.~\ref{fig:6}(d) and Fig.~\ref{fig:appendix_fig8}). This decoupled dual diffusion framework sheds light on the roles of deresidual and denoising in the DDPM generation process.\footnotetext[4]{Our RDDM is implemented based on the popular diffusion repository github.com/lucidrains/denoising-diffusion-pytorch. Differences in network structure and training details may lead to poorer FID. We have verified sampling consistency with DDIM~\citep{song2020denoising} in Table~\ref{table:all}(a) and Appendix~\ref{Appendix:a.3}.}

\subsection{Decoupled Forward Diffusion Process}\label{Sec:4.1}
\begin{table}[t]
    \resizebox{\columnwidth}{!}{
        \setlength{\abovecaptionskip}{0.0cm}
        \setlength{\belowcaptionskip}{0.0cm}
        \caption{Coefficient schedules analysis on CelebA ($64\times 64$) \citep{liu2015faceattributes}. In our RDDM, the residual diffusion and noise diffusion are decoupled, so one may design a better schedule in the decoupled coefficient space, e.g., $\alpha_t$ (linearly decreasing), $\beta_t^2$ (linearly increasing). To be fair, all coefficient schedules were retrained using the same network structure, training, and evaluation. The sampling method is SM-N with 10 sampling steps using Eq.~\ref{Eq:15}.}\label{table:alpha_schedules}
        \setlength{\tabcolsep}{0.1cm}{
            \begin{tabular}{lcc}
                \toprule
                Schedules                                                           & FID ($\downarrow $)           & IS ($\uparrow $) \\
                \midrule
                Linear (DDIM~\cite{song2020denoising})                              & 28.39\protect\footnotemark[4] & 2.05             \\
                Scaled linear~\citep{rombach2022high}                               & 28.15                         & 2.00             \\
                Squared cosine~\citep{nichol2021improved}                           & 47.21                         & {\bf 2.64}       \\
                \hline
                $\alpha_t$ (mean), $\beta_t^2$ (mean)                               & 38.35                         & 2.22             \\
                $\alpha_t$ (linearly increasing), $\beta_t^2$ (linearly increasing) & 40.03                         & \underline{2.45} \\
                $\alpha_t$ (linearly decreasing), $\beta_t^2$ (linearly decreasing) & \underline{27.82}             & 2.26             \\
                $\alpha_t$ (linearly decreasing), $\beta_t^2$ (linearly increasing) & {\bf 23.25}                   & 2.05             \\
                \bottomrule
            \end{tabular}
        }}
    \vspace{-0.3cm}
\end{table}
\begin{figure}[t]
    \setlength{\abovecaptionskip}{0.0cm}
    \centering
    \includegraphics[width=0.33\linewidth,height=0.33\textwidth]{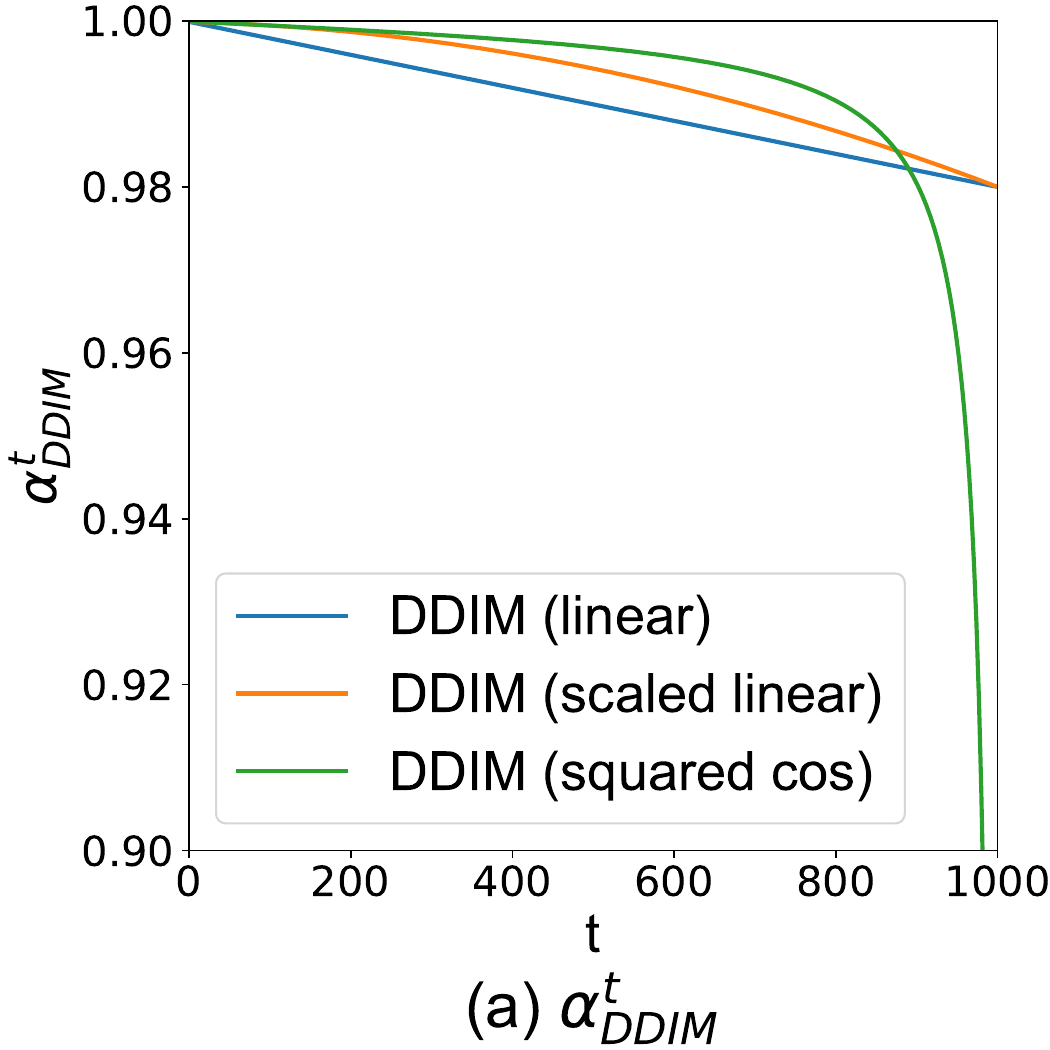}
    \hspace{-0.19cm}
    \includegraphics[width=0.33\linewidth,height=0.33\textwidth]{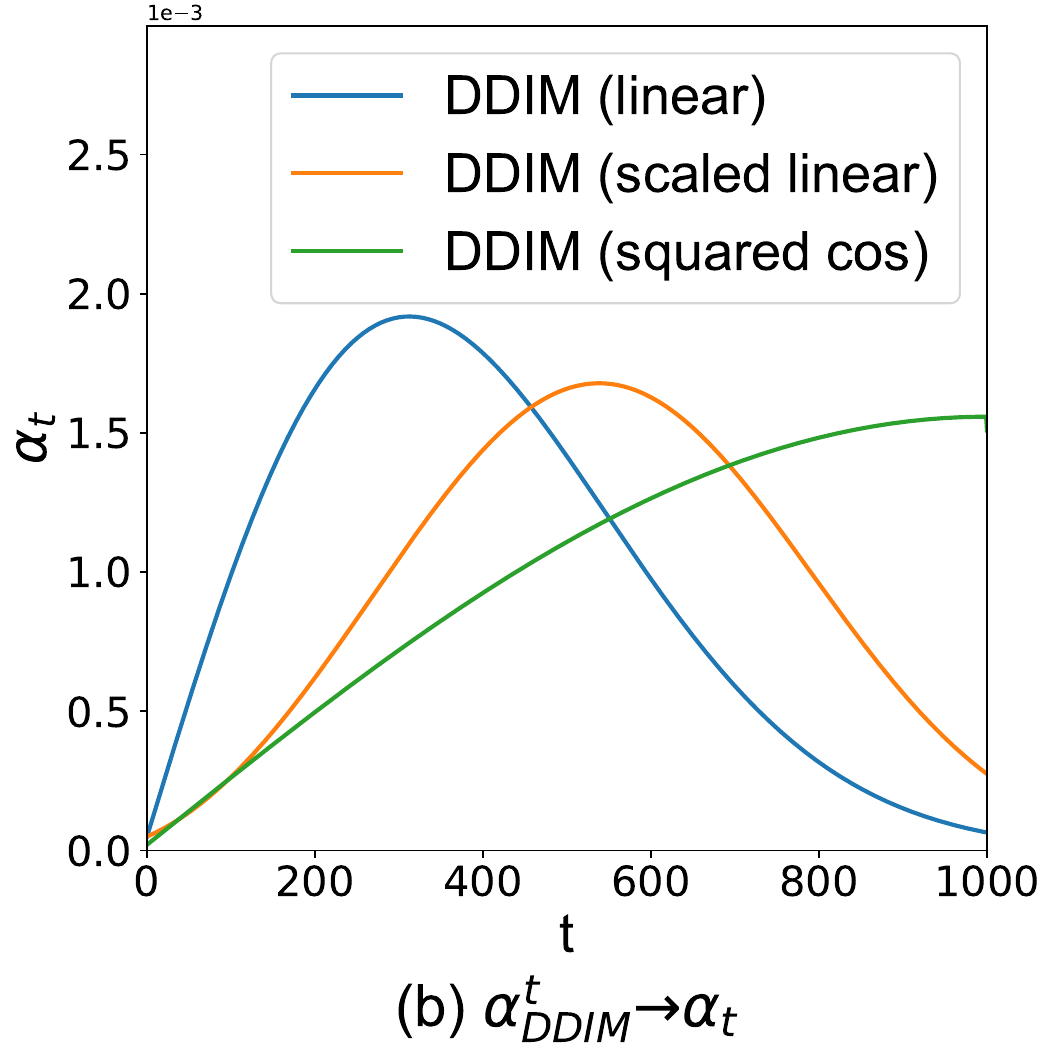}
    \hspace{-0.2cm}
    \includegraphics[width=0.33\linewidth,height=0.33\textwidth]{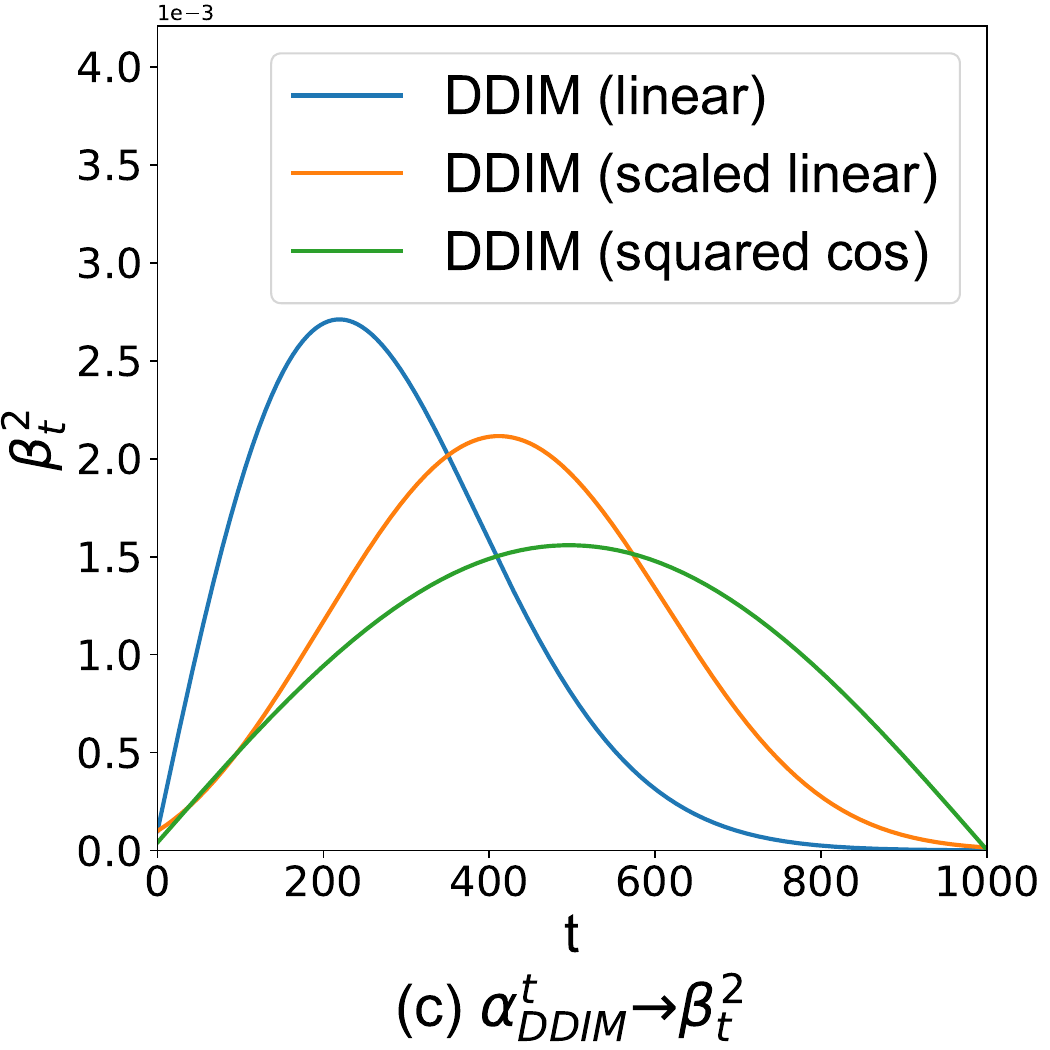}
    \hspace{-0.2cm}
    \includegraphics[width=0.33\linewidth,height=0.33\textwidth]{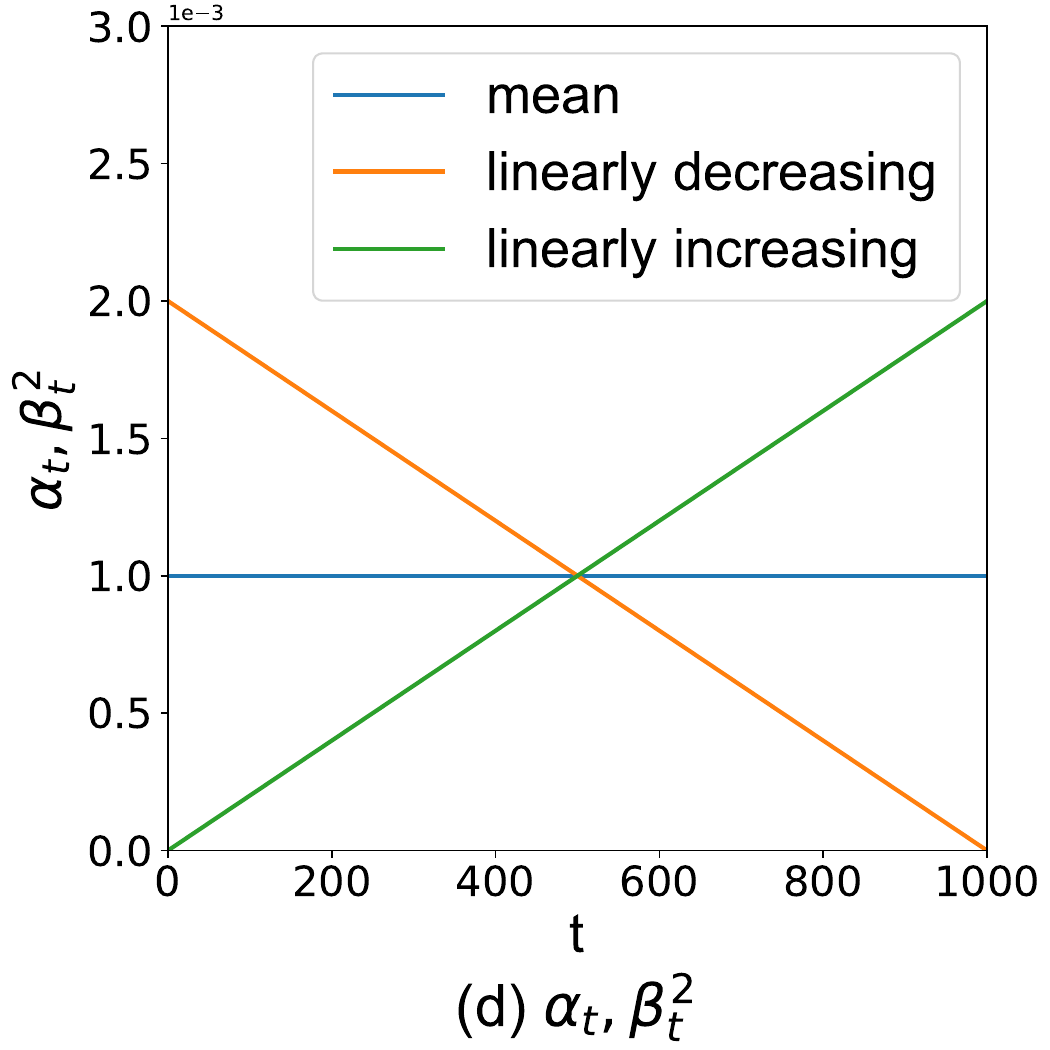}
    \includegraphics[width=0.33\linewidth,height=0.33\textwidth]{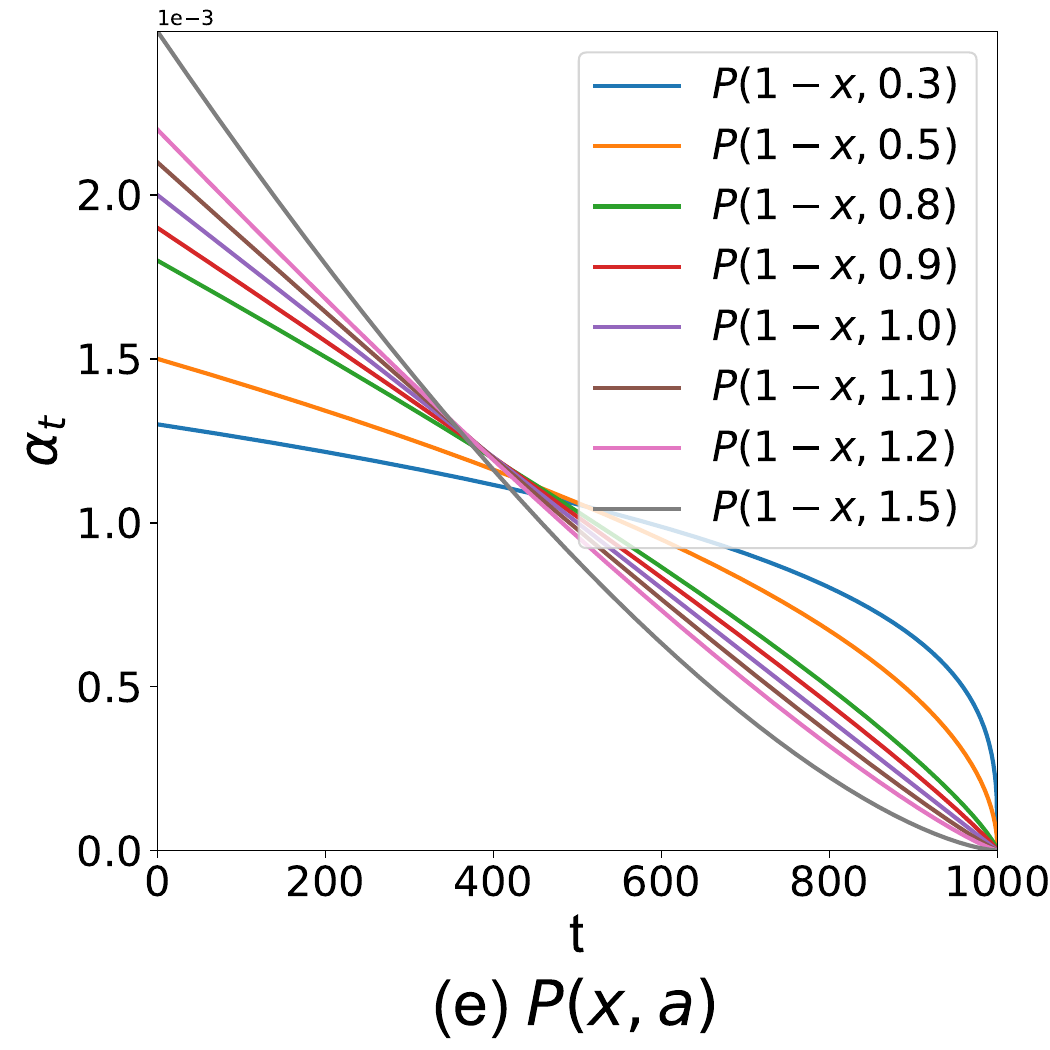}
    \caption{Coefficient transformation from DDIM~\citep{song2020denoising} to RDDM using Eq.~\ref{Eq:19}. (a) We show several schedules for $\alpha^t_{DDIM}$, e.g., linear~\citep{song2020denoising}, scaled linear~\citep{rombach2022high}, and squared cosine~\citep{nichol2021improved}. (b) We transform $\alpha_{DDIM}^t$ into $\alpha_t$ in our RDDM. (c) We transform $\alpha_{DDIM}^t$ into $\beta_t^2$ in our RDDM.
    (d) A few simple schedules. (e) $P(x,a)$ is a normalized power function (see Eq.~\ref{Eq:20}). "mean", "linearly increasing", and "linearly decreasing" in (d) can be denoted as $P(x,0)$, $P(x,1)$ and $P(1-x,1)$, respectively. See Algorithm~\ref{Algorithm1} in Appendix~\ref{Appendix:a.3} for more details of (b) and (c).\vspace{-0.3cm}}
    \label{fig:alpha}
\end{figure}
Our defined coefficients ($\alpha_t,\beta_t^2$) offer a distinct physical interpretation. In the forward diffusion process (Eq.~\ref{Eq:8}), $\alpha_t$ controls the speed of residual diffusion and $\beta_t^2$ regulates the speed of noise diffusion. In the reverse generation process (Eq.~\ref{Eq:15}), $\bar{\alpha}_t$ and $\bar{\beta}_t$ are associated with the speed of removing residual and noise, respectively.
In fact, there are no constraints on $\alpha^t$ and $\beta_t^2$ in Eq.~\ref{Eq:8}, meaning that the residual diffusion and noise diffusion are independent of each other. Utilizing this decoupled property and the difference between these two diffusion processes, we should be able to design a better coefficient schedule, e.g., $\alpha_t$ (linearly decreasing) and $\beta_t^2$ (linearly increasing) in Table~\ref{table:alpha_schedules}. This aligns with the intuition that, during the reverse generation process (from $T$ to $0$), the estimated residuals become increasingly accurate while the estimated noise should also weaken progressively.
Therefore, when $t$ is close to $0$, the deresidual pace should be faster and the denoising pace should be slower. Since our $\alpha_t$ and $\beta_t^2$ represent the speed of diffusion, we name the curve in Fig.~\ref{fig:alpha} (b-d) the \textit{diffusion speed curve}.

\begin{figure}[t]
    \setlength{\abovecaptionskip}{0.1cm}
    \setlength{\belowcaptionskip}{0.cm}
    \centering
    \includegraphics[width=1\linewidth]{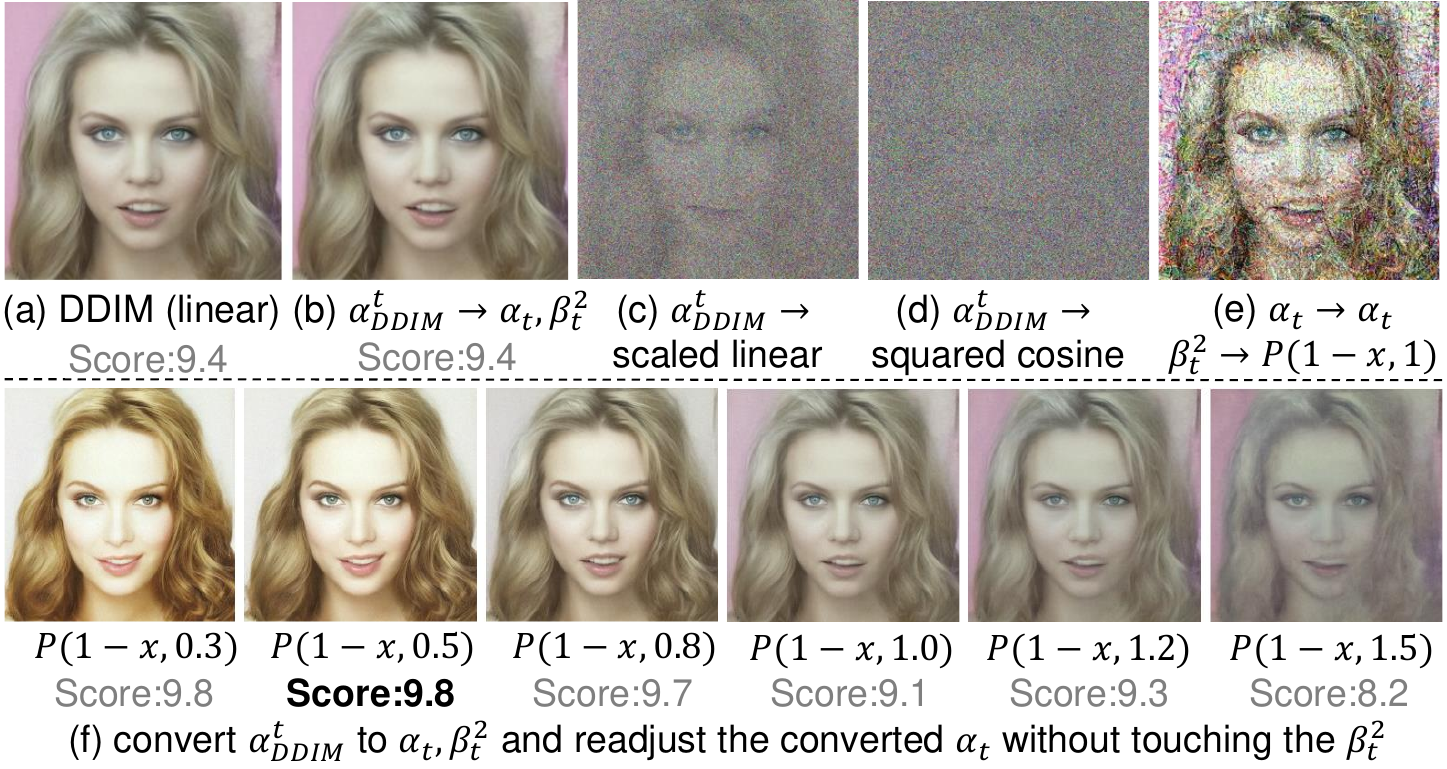}
    \caption{Analysis of readjusting coefficient schedules.
        We find that changing the $\alpha_t$ schedule barely affects the denoising process in (f) and edited faces may have higher face scores when assessed using AI face scoring software\protect\footnotemark[5].
        These images were generated using a pre-trained UNet on the CelebA ($256\times 256$) dataset \citep{liu2015faceattributes} with 10 sampling steps.\vspace{-0.3cm}}
    \label{fig:4}
    \vspace{-0.3cm}
\end{figure}

\subsection{Partially Path-independent Generation Process}\label{Sec:4.2}

In the original DDPM~\citep{ho2020denoising} or DDIM~\citep{song2020denoising}, when the $\alpha^t_{DDIM}$ schedule changes, it is necessary to retrain the denoising network because this alters the diffusion process~\citep{rombach2022high,nichol2021improved}.
As shown in Fig.~\ref{fig:4}(c)(d), directly changing the $\alpha^t_{DDIM}$ schedule causes denoising to fail. Here, we propose a path-independent generation process, i.e., modifying the diffusion speed curve does not cause the image generation process to fail.
We try to readjust the diffusion speed curve in the generation process.
First, we convert the $\alpha^t_{DDIM}$ schedule of a pre-trained DDIM into the $\alpha_t$ and $\beta_t^2$ schedules of our RDDM using Eq.~\ref{Eq:19} (from Fig.~\ref{fig:4}(a) to Fig.~\ref{fig:4}(b). We then readjust the converted  $\alpha_t$ schedules using the normalized power function ($P(x,a)$ in Fig.~\ref{fig:4}(f)), without touching the $\beta^2_t$ schedule that controls noise diffusion, as shown in Fig.~\ref{fig:4}(f). $P(x,a)$ is defined as ($a$ is a parameter of the power function),
\begin{align}
    P(x,a):=x^a/{\int_{0}^{1}} x^adx, \mathrm{where} \,\, x=t/T.\label{Eq:20}
\end{align}
These schedule modifications shown in Fig.~\ref{fig:4} lead to the following key findings.

    {\bf 1.} Fig.~\ref{fig:4}(f) shows that modifying the residual diffusion speed curve ($\alpha_t$) leads to a drastic change in the generation results, probably due to $I_{res}^\theta$ being represented as a transformation of $\epsilon_\theta$ using Eq.~\ref{Eq:17}.

{\bf 2.} As the time condition $t$ represents the current noise intensity in the denoising network ($\epsilon_\theta(I_t,t,0)$), modifying the noise diffusion speed curve ($\beta_t^2$) causes $t$ to deviate from accurately indicating the current noise intensity, leading to denoising failure, as shown in Fig.~\ref{fig:4}(e).

Nonetheless, we believe that, corresponding to the decoupled forward diffusion process, there should also be a path-independent reverse generation process. To develop a path-independent generation process, we improve the generation process based on the above two key findings:\footnotetext[5]{https://ux.xiaoice.com/beautyv3}

{\bf 1.} Two networks are used to estimate $I_{res}^\theta$ and $\epsilon_\theta$ separately, i.e., SM-Res-N-2Net in Appendix~\ref{Appendix:b.3}.

{\bf 2.} $\bar{\alpha}_t$ and $\bar{\beta}_t$ are used for the time conditions embedded in the network, i.e., $I_{res}^\theta(I_t,t,0) \to I_{res}^\theta(I_t,\bar{\alpha}_t\cdot T,0)$, $\epsilon_\theta(I_t,t,0) \to \epsilon_\theta(I_t,\bar{\beta}_t\cdot T,0)$.

\begin{figure}[t]
    \setlength{\abovecaptionskip}{0.cm}
    \setlength{\belowcaptionskip}{0.cm}
    \centering
    \includegraphics[width=1\linewidth]{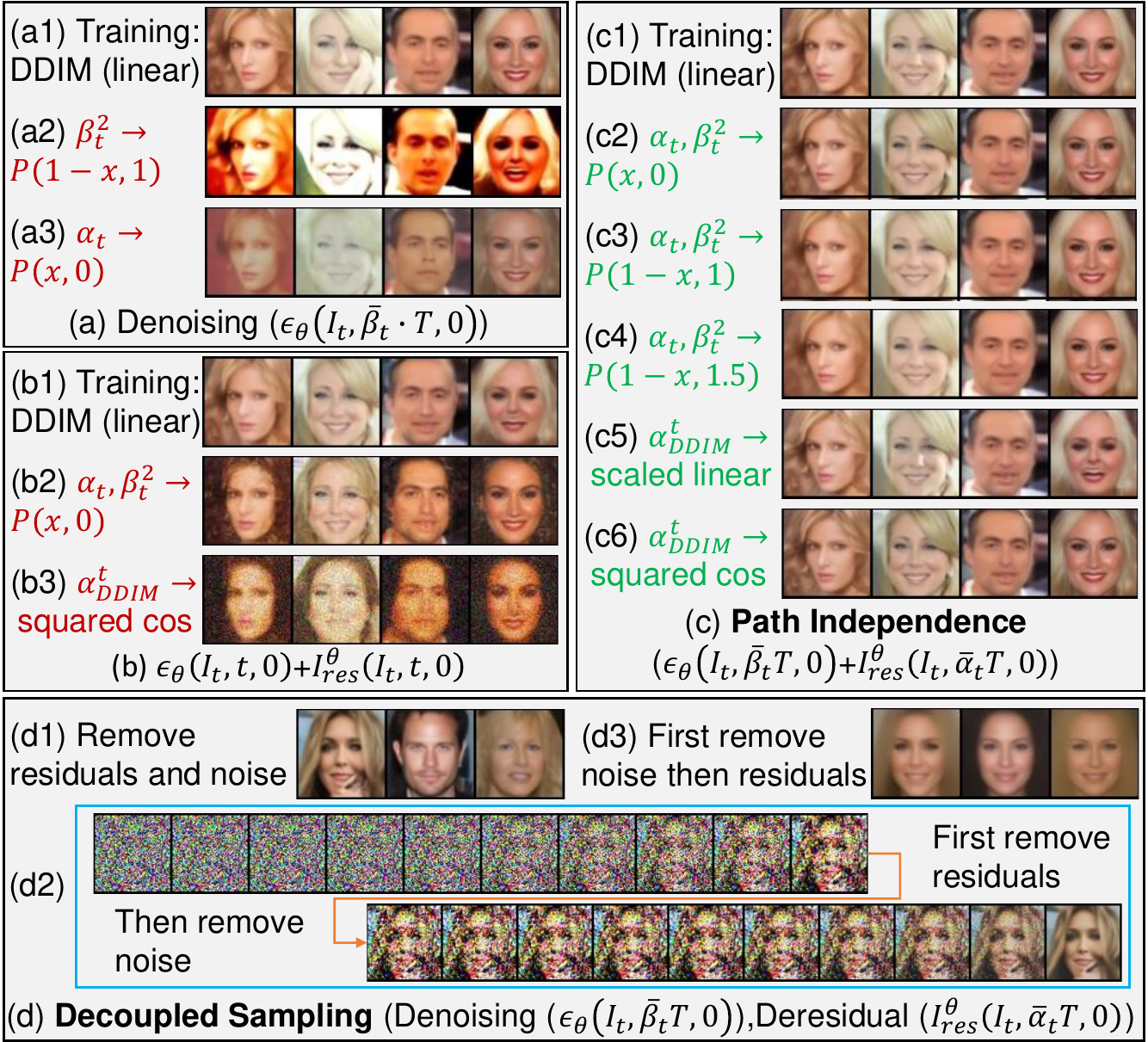}
    \caption{Partially path-independent generation process. (a1) We trained a denoising network using the DDIM linear schedule~\citep{song2020denoising}. (a2-a3) We modified the $\alpha_t$ and $\beta^2_t$ schedules during testing. (b) We trained two networks to remove noise and residuals. In contrast to the sharply varying images in (a2-a3) and the noisy images in (b2-b3), (c) shows that we constructed a path independent generation process where modifications to the diffusion speed curve can generate a noise-free image with little variation in image semantics. (d) The simultaneous sampling in (d1) or (c) can be decomposed into first removing residuals and then noise (d2), or removing noise and then residuals (d3). In (d3), diversity is significantly reduced because noise is removed first.}
    \label{fig:6}
\end{figure}

These improvements lead to a partially path-independent generation process, as evidenced by the results shown in Fig.~\ref{fig:6}(c).

\begin{figure*}[t]
    \centering
    \includegraphics[width=1\linewidth]{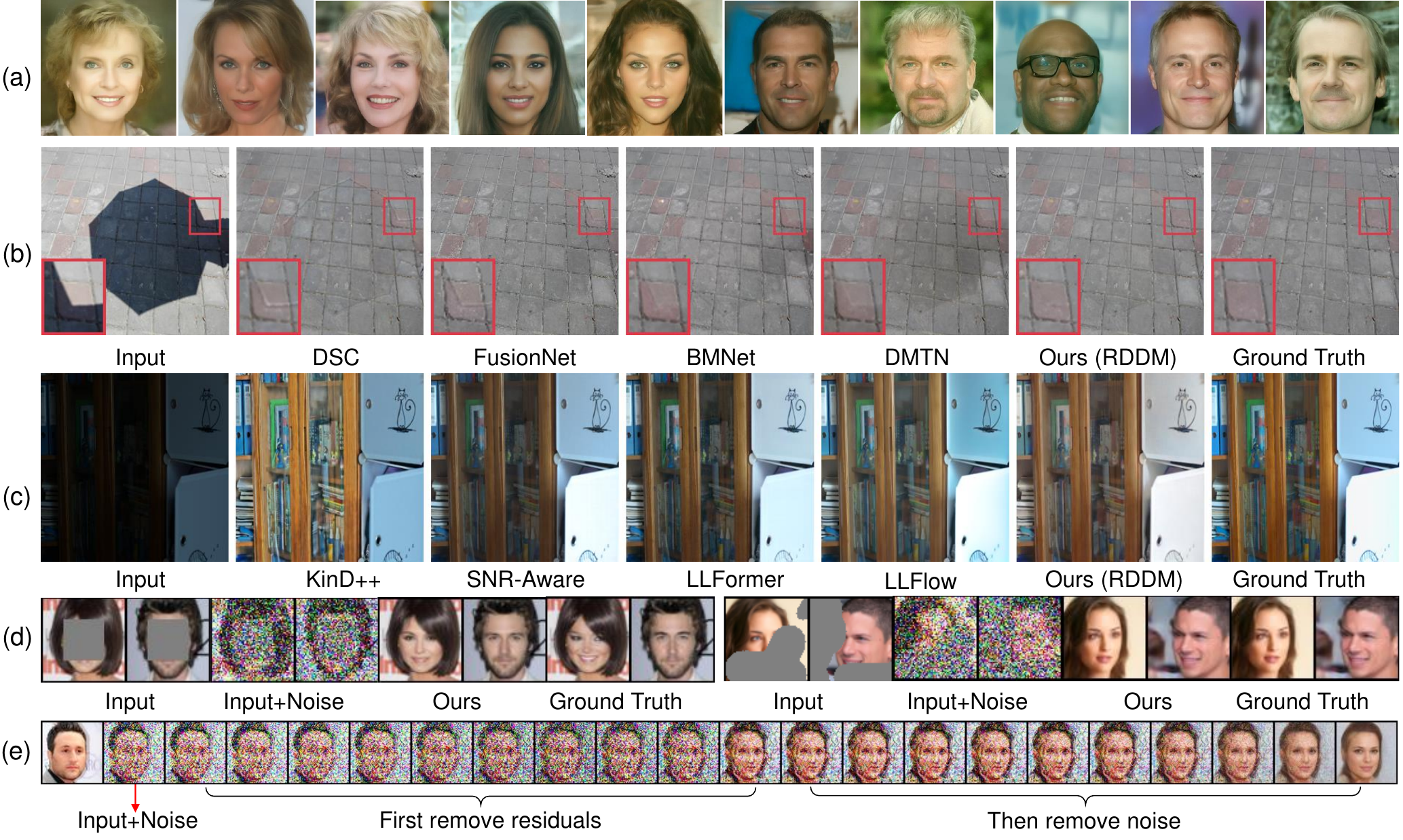}
    \caption{Application of our RDDM. (a) Image generation on the CelebA dataset \citep{liu2015faceattributes}. (b) Shadow removal on the ISTD dataset~\citep{wang2018stacked}. (c) Low-light enhancement on the LOL dataset~\citep{wei2018deep}. (d) Image inpainting (center and irregular mask). (e) The image translation process can be regarded as first translating the semantics and then generating the details. These images in (b) are magnified using MulimgViewer \citep{MulimgViewer}.}
    \label{fig:7}
\end{figure*}

{\bf Analysis of Partially Path-independence via Green's Theorem.} ``Path-independence" reminds us of Green's theorem in curve integration~\citep{riley_hobson_bence_2006}.
From Eq.~\ref{Eq:15}, we have:
\begin{align}
    I_t-I_{t-1}= & (\bar{\alpha} _t-\bar{\alpha} _{t-1})I_{res}^\theta +(\bar{\beta }_t-\bar{\beta }_{t-1} )\epsilon _\theta, \\
    \begin{split}
        \mathrm{d} I(t) =& I_{res}^\theta(I(t),\bar{\alpha}(t)\cdot T,0) \mathrm{d} \bar{\alpha}(t)
        \\&+\epsilon _\theta (I(t),\bar{\beta}(t)\cdot T,0) \mathrm{d}\bar{\beta}(t),
    \end{split}
\end{align}
where $I(t)=I(0)+\bar{\alpha}(t) I_{res}+\bar{\beta}(t) \epsilon$.
Given inputs $I(t)$ and $\bar{\alpha}(t)$, the denoising network learns to approximate the noise $\epsilon$ in $I(t)$ by estimating  $\epsilon_{\theta}$. If this network is trained well and robust enough, it should be able to avoid the interference of the residual terms $\bar{\alpha}(t) I_{res}$ in $I(t)$. This also applies to a robust residual estimation network. Thus, we have
\begin{align}
    \frac{\partial I_{res}^\theta(I(t),\bar{\alpha}(t)\cdot T)}{\partial \bar{\beta}(t)}\approx  0, \frac{\partial \epsilon _\theta (I(t),\bar{\beta}(t)\cdot T)}{\partial \bar{\alpha}(t)}\approx  0. \label{Eq:22}
\end{align}
If the equation in Formula~\ref{Eq:22} holds true, it serves as a necessary and sufficient condition for path independence in curve integration, which provides an explanation for why Fig.~\ref{fig:6}(c) achieves a partially path-independent generation process.
The path-independent property is related to the network's resilience to disturbances and applies to disturbances that vary within a certain range. However, excessive disturbances can lead to visual inconsistencies, e.g., readjusting $\alpha_t$ and $\beta_t^2$ to $P(x,5)$. Thus, we refer to this generative property as partially path-independent.
We also investigated two reverse paths to gain insight into the implications of the proposed partial path independence. In the first case, the residuals are removed first, followed by the noise: $I(T)\overset{-I_{res}}{\rightarrow} I(0)+\bar{\beta }_T\epsilon \overset{-\bar{\beta }_T\epsilon }{\rightarrow}I(0)$. The second case involves removing the noise first and then the residuals: $I(T)\overset{-\bar{\beta }_T\epsilon }{\rightarrow} I_{in} \overset{-I{res}}{\rightarrow} I(0)$. The first case (Fig.~\ref{fig:6}(d2)) shows that removing residuals controls semantic transitions, while the second case (Fig.~\ref{fig:6}(d3)) shows that diversity is significantly reduced because noise is removed first. Fig.~\ref{fig:6}(d) validates our argument that residuals control directional semantic drift (certainty) and noise controls random perturbation (diversity). See Appendix~\ref{Appendix:c.1} for more details.

\section{Experiments} \label{Sec:5.0}

\begin{table}[t]

    \resizebox{1\columnwidth}{!}{
        \setlength{\tabcolsep}{1.2mm}{
            \begin{tabular}{l|c|c|c|c|c}
                \toprule
                {\bf (a)} CelebA (FID) & 5 steps & 10 steps & 15 steps & 20 steps & 100 steps \\
                \midrule
                DDIM                   & 69.60   & 40.45    & 32.67    & 30.61    & 23.66     \\
                DDIM$\to$RDDM          & 69.60   & 40.41    & 32.71    & 30.77    & 24.92     \\
                \hline
            \end{tabular}
        }}
\vspace{0.1cm}
    \resizebox{1\columnwidth}{!}{
        \setlength{\tabcolsep}{0.8mm}{
            \begin{tabular}{l|c|c|c|c|c|c|c|c|c}
                \toprule
                {\bf (b)} Shadow                 & \multicolumn{3}{c|}{MAE($\downarrow $)} & \multicolumn{3}{c|}{SSIM($\uparrow $)} & \multicolumn{3}{c}{PSNR($\uparrow $)}                                                                                                                         \\
                \cline{2-10}
                Removal                          & S                                       & NS                                     & ALL                                   & S                 & NS                & ALL               & S                 & NS                & ALL               \\
                \hline
                DSC~\citep{hu2019direction} $\P$ & 9.48                                    & 6.14                                   & 6.67                                  & 0.967             & -                 & -                 & 33.45             & -                 & -                 \\
                FusionNet~\citep{fu2021auto}     & 7.77                                    & 5.56                                   & 5.92                                  & 0.975             & 0.880             & 0.945             & 34.71             & 28.61             & 27.19             \\
                BMNet~\citep{zhu2022bijective}   & 7.60                                    & 4.59                                   & 5.02                                  & \underline{0.988} & \underline{0.976} & 0.959             & 35.61             & 32.80             & 30.28             \\
                DMTN~\citep{liu2023decoupled}    & \underline{7.00}                        & \underline{4.28}                       & \underline{4.72}                      & {\bf 0.990}       & {\bf 0.979}       & {\bf 0.965}       & \underline{35.83} & \underline{33.01} & \underline{30.42} \\
                Ours (RDDM)                      & {\bf 6.67}                              & {\bf 4.27}                             & {\bf 4.67}                            & \underline{0.988} & {\bf 0.979}       & \underline{0.962} & {\bf 36.74}       & {\bf 33.18}       & {\bf 30.91}       \\
                \hline
            \end{tabular}
        }}
    \vspace{0.1cm}
    \resizebox{1\columnwidth}{!}{
        \setlength{\tabcolsep}{0.1mm}{
            \begin{tabular}{l|c|c|c||l|c|c}
                \toprule
                {\bf (c)}  Low-light           & PSNR($\uparrow $) & SSIM($\uparrow $) & LPIPS     ($\downarrow $) & {\bf (d)} Deraining                        & PSNR($\uparrow $) & SSIM($\uparrow $)  \\
                \midrule
                KinD++~\citep{zhang2021beyond} & 17.752            & 0.760             & 0.198                     & AttnGAN~\citep{qian2018attentive}          & 31.59             & 0.9170             \\
                KinD++-SKF~\citep{wu2023skf}   & 20.363            & 0.805             & 0.201                     & DuRN~\citep{liu2019dual}                   & 31.24             & 0.9259             \\
                DCC-Net~\citep{zhang2022deep}  & 22.72             & 0.81              & -                         & RainAttn~\citep{quan2019deep}              & 31.44             & 0.9263             \\
                SNR-Aware~\citep{xu2022snr}    & 24.608            & 0.840             & 0.151                     & IDT~\citep{xiao2022image}                  & 31.87             & 0.9313             \\
                LLFlow~\citep{wang2022low}     & \underline{25.19} & \underline{0.93}  & {\bf 0.11}                & RainDiff64~\citep{ozdenizci2023restoring}  & 32.29             & \underline{0.9422} \\
                LLFormer~\citep{wang2023ultra} & 23.649            & 0.816             & 0.169                     & RainDiff128~\citep{ozdenizci2023restoring} & \underline{32.43} & 0.9334             \\ 
                Ours (RDDM)                    & {\bf 25.392}      & {\bf 0.937}       & \underline{0.116}         & Ours (RDDM)                                & {\bf 32.51}       & {\bf 0.9563}       \\
                \bottomrule
            \end{tabular}
        }}{
        \caption{Quantitative comparison results of image generation on the CelebA ($256\times 256$) dataset \citep{liu2015faceattributes}, shadow removal on the ISTD dataset~\citep{wang2018stacked}, low-light enhancement on the LOL~\citep{wei2018deep} dataset, and deraining on the RainDrop~\citep{qian2018attentive} dataset. ``S, NS, ALL'' in (b) denote shadow area (S), non-shadow area (NS) and whole image (ALL). The sampling steps are 5 for shadow removal and deraining, 2 for low-light.}
        \label{table:all}
    }
    \vspace{-0.3cm}
\end{table}

{\bf Image Generation.} We can convert a pre-trained\protect\footnotemark[6]\footnotetext[6]{https://huggingface.co/google/ddpm-celebahq-256} DDIM~\citep{song2020denoising} to RDDM by coefficient transformation using Eq.~\ref{Eq:19}, and generate images by Eq.~\ref{Eq:13}. Table~\ref{table:all}(a) verifies that the quality of the generated images before and after the conversion is nearly the same\protect\footnotemark[7]\footnotetext[7]{The subtle differences in larger sampling steps may stem from errors introduced by numerical representation limitations during coefficient transformation, which may accumulate and amplify in larger sampling steps.}.
We show the generated face images with 10 sampling steps in Fig.~\ref{fig:7}(a).

{\bf Image Restoration.} We extensively evaluate our method on several image restoration tasks, including shadow removal, low-light enhancement, deraining, and deblurring on 5 datasets.
Notably, our RDDM uses an identical UNet and is trained with a batch size of 1 for all these tasks. In contrast, SOAT methods often involve elaborate network architectures, such as multi-stage~\citep{fu2021auto,zhu2022efficient,wang2022low}, multi-branch~\citep{cun2020towards}, Transformer~\citep{wang2023ultra}, and GAN~\citep{kupyn2019deblurgan}, or sophisticated loss functions like the chromaticity~\citep{jin2021dc}, texture similarity~\citep{zhang2019kindling}, and edge loss~\citep{zamir2021multi}. Table~\ref{table:all} and Fig.~\ref{fig:7}(b-c) show that our RDDM is competitive with the SOTA restoration methods.
See Appendix~\ref{Appendix:b.0} for more training details and comparison results.

We extend DDPM~\citep{ho2020denoising}/DDIM~\citep{song2020denoising}, initially uninterpretable for image restoration, into a unified and interpretable diffusion model for both image generation and restoration by introducing residuals. However, the residual diffusion process represents the directional diffusion from target images to conditional input images, which does not involve a priori information about the image restoration task, and therefore is not limited to it. Beyond image generation and restoration, we show examples of image inpainting and image translation to verify that our RDDM has the potential to be a unified and interpretable methodology for image-to-image distribution transformation. {\bf We do not intend to achieve optimal performance on all tasks by tuning all hyperparameters.} The current experimental results show that RDDM 1) achieves consistent {\bf image generation} performance with DDIM after coefficient transformation, 2) competes with state-of-the-art {\bf image restoration} methods using a generic UNet with only an $\ell _1$ loss, a batch size of 1, and fewer than 5 sampling steps, and 3) has satisfactory visual results of {\bf image inpainting} and {\bf image translation} (see Fig.~\ref{fig:7}(d-e), Fig.~\ref{fig:appendix_inpainting}, or Fig.~\ref{fig:appendix_translation} in Appendix~\ref{Appendix:b.2}), which validates our RDDM.

\section{Conclusions}
We present a unified dual diffusion model called Residual Denoising Diffusion Models (RDDM) for image restoration and image generation. This is a three-term mixture framework beyond the previous denoising diffusion framework with two-term mixture. We demonstrate that our sampling process is consistent with that of DDPM and DDIM through coefficient schedule transformation, and propose a partially path-independent generation process. Our experimental results on four different image restoration tasks show that RDDM achieves SOTA performance in no more than five sampling steps.
We believe that our model and framework hold the potential to provide a unified methodology for image-to-image distribution transformation and pave the way for the multi-dimensional diffusion process.
\section{Acknowledgments}
This work was supported by National Natural Science Foundation of China under Grants 61991413, 62273339, 62073205, 62306253.

{
    \small
    \bibliographystyle{ieeenat_fullname}
    \bibliography{main}
}



\clearpage
\onecolumn
\appendix
\section*{Appendix}

\section{Derivations and Proofs}
\subsection{Perturbed Generation Process}\label{Appendix:a.1}


For the reverse generation process from $I_{t}$ to $I_{t-1}$,
we can represent the transfer probabilities $q(I_{t-1}|I_{t},I_0,I_{res})$ by Bayes' rule:
\begin{align}
   q(I_{t-1}|I_{t},I_0,I_{res})=q(I_{t}|I_{t-1},I_0,I_{res})\frac{q(I_{t-1}|I_0,I_{res})}{q(I_t|I_0,I_{res})},\label{Eq:22-1}
\end{align}
where $q(I_{t-1}|I_0,I_{res})=\mathcal{N} (I_{t-1};I_0+\bar{\alpha }_{t-1} I_{res},\bar{\beta }_{t-1}^2\mathbf{I}  )$ from Eq.~\ref{Eq:8}, and $q(I_{t}|I_{t-1},I_0,I_{res})=q(I_{t}|I_{t-1},I_{res})$\protect\footnotemark[8]\footnotetext[8]{Each step in Eq.~\ref{Eq:8} adds a new random Gaussian noise in the random forward diffusion. Thus for simplicity, we assume $q(I_{t}|I_{t-1},I_0,I_{res})=q(I_{t}|I_{t-1},I_{res})$, it follows that $I_0$ is not important for $I_t$ when $I_{t-1}$ presents as a condition. }$=\mathcal{N} (I_{t};I_{t-1}+\alpha_t I_{res},\beta_t^2\mathbf{I})$ from Eq.~\ref{Eq:9}. Thus, we have (considering only the exponential term)
\begin{align}
   \begin{split}
      &q(I_{t-1}|I_{t},I_0,I_{res})=\mathcal{N} (I_{t-1};\mu _t (x_t,I_0,I_{res}),\Sigma_t (x_t,I_0,I_{res})\mathbf{I}  )\\
      &\propto exp\left (-\frac{1}{2}(\frac{(I_t-I_{t-1}-\alpha_{t} I_{res})^2}{\beta _t^2} +\frac{(I_{t-1}-I_{0}-\bar{\alpha}_{t-1} I_{res})^2}{\bar{\beta} _{t-1}^2})-\frac{(I_{t}-I_{0}-\bar{\alpha}_{t} I_{res})^2}{\bar{\beta} _{t}^2})  \right)\\
      &= exp\left (-\frac{1}{2} ((\frac{\bar{\beta}_{t}^2}{\beta_t^2\bar{\beta}_{t-1}^2} )I_{t-1}^2-2(\frac{I_t-\alpha_tI_{res}}{\beta_t^2}+\frac{\bar{\alpha}_{t-1}I_{res}+I_0}{\bar{\beta}_{t-1}^2}  )I_{t-1}+C(I_t,I_0,I_{res}))\right),\label{Eq:23}\\
   \end{split}
\end{align}
where the $C(I_t,I_0,I_{res})$ term is not related to $I_{t-1}$. From Eq.~\ref{Eq:23}, $\mu _t (x_t,I_0,I_{res})$ and $\Sigma_t (x_t,I_0,I_{res})$ are represented as follows,
\begin{align}
   \mu _t (x_t,I_0,I_{res})   & =(\frac{I_t-\alpha_tI_{res}}{\beta_t^2}+\frac{\bar{\alpha}_{t-1}I_{res}+I_0}{\bar{\beta}_{t-1}^2}  )/\frac{\bar{\beta}_{t}^2}{\beta_t^2\bar{\beta}_{t-1}^2}\label{Eq:24}                            \\
                              & =\frac{\bar{\beta}_{t-1}^2}{\bar{\beta}_{t}^2} I_t+\frac{\beta_t^2\bar{\alpha}_{t-1}-\bar{\beta}_{t-1}^2\alpha_t}{\bar{\beta}_{t}^2} I_{res}+\frac{\beta_{t}^2}{\bar{\beta}_{t}^2} I_0\label{Eq:25} \\
                              & =I_t-\alpha_tI_{res}-\frac{\beta_{t}^2}{\bar{\beta}_{t}}\epsilon,\label{Eq:26}                                                                                                                      \\
   \Sigma_t (x_t,I_0,I_{res}) & =\frac{\beta_t^2\bar{\beta}_{t-1}^2}{\bar{\beta}_{t}^2}.
\end{align}
Eq.~\ref{Eq:8} is used for the derivation from Eq.~\ref{Eq:25} to Eq.~\ref{Eq:26}.
Then, we define the generation process to start from $p_\theta(I_T)\sim \mathcal{N}(I_T;\mathbf{0},\mathbf{I})$,
\begin{align}
   p_\theta(I_{t-1}|I_t) = q(I_{t-1}|I_{t},I_0^\theta,I_{res}^\theta),
\end{align}
where $I_0^\theta=I_t-\bar{\alpha}_t I_{res}^\theta-\bar{\beta}_t \epsilon_\theta$ by Eq.~\ref{Eq:8}. Here we only consider $L_{t-1}$ in~\citep{ho2020denoising},
\begin{align}
   L_{t-1} & =D_{KL}(q(I_{t-1}|I_{t},I_0,I_{res})||p_\theta(I_{t-1}|I_t))                                                                                                                                                   \\
           & =\mathbb{E} \left [\left \| I_t-\alpha_tI_{res}-\frac{\beta_{t}^2}{\bar{\beta}_{t}}\epsilon-(I_t-\alpha_tI_{res}^\theta-\frac{\beta_{t}^2}{\bar{\beta}_{t}}\epsilon_\theta) \right \|^2 \right ],\label{Eq:30}
\end{align}
where $D_{KL}$ denotes KL divergence.
The noise $\epsilon$ can be represented as a transformation of $I_{res}$ using Eq.~\ref{Eq:17}, and then Eq.~\ref{Eq:30} simplifies to:
\begin{align}
   L_{res}(\theta ):=\mathbb{E} \left [ \lambda _{res}\left \| I_{res}-I_{res}^{\theta }(I_t,t,I_{in}) \right \|^2 \right ]\label{Eq:31}
\end{align}
In addition, the residuals $I_{res}$ can be represented as a transformation of $\epsilon$ using Eq.~\ref{Eq:17}, and then Eq.~\ref{Eq:30} simplifies to:\\
\begin{align}
   L_{\epsilon}(\theta ):=\mathbb{E} \left [ \lambda _{\epsilon}\left \| \epsilon-\epsilon _{\theta }(I_t,t,I_{in}) \right \|^2 \right ],\label{Eq:32-2}
\end{align}
We reset the weights, i.e., $\lambda _{res}, \lambda _{\epsilon} \in\{0,1\}$, similar to DDPM~\cite{ho2020denoising} discarding weights (or regarding them as reweighting) in simplified training objective.

\subsection{Deterministic Implicit Sampling}\label{Appendix:a.2}
If $q_\sigma (I_{t-1}|I_t,I_0,I_{res})$ is defined in Eq.~\ref{Eq:11}, we have:
\begin{align}
   q(I_t|I_0,I_{res})=\mathcal{N} (I_{t};I_0+\bar{\alpha }_t I_{res},\bar{\beta }_t^2\mathbf{I}).\label{Eq:33}
\end{align}
{\it Proof}. Similar to the evolution from DDPM~\citep{ho2020denoising} to DDIM~\citep{song2020denoising}, we can prove the statement with an induction argument for $t$ from $T$ to $1$. Assuming that Eq.~\ref{Eq:33} holds at $T$,  we just need to verify $q(I_{t-1}|I_0,I_{res})=\mathcal{N} (I_{t-1};I_0+\bar{\alpha }_{t-1} I_{res},\bar{\beta }_{t-1}^2\mathbf{I})$ at $t-1$ from $q(I_t|I_0,I_{res})$ at $t$ using Eq.~\ref{Eq:33}. Given:
\begin{align}
    & q(I_t|I_0,I_{res})=\mathcal{N} (I_{t};I_0+\bar{\alpha }_t I_{res},\bar{\beta }_t^2\mathbf{I}),\label{Eq:35}                                                                                                                         \\
    & q_\sigma (I_{t-1}|I_t,I_0,I_{res})=\mathcal{N} (I_{t-1};I_0+\bar{\alpha }_{t-1}I_{res}+\sqrt{\bar{\beta }_{t-1}^2 -\sigma _t^2} \frac{I_t-(I_0+\bar{\alpha }_{t}I_{res})}{\bar{\beta }_{t}} ,\sigma _t^2\mathbf{I}  ),\label{Eq:36} \\
    & q(I_{t-1}|I_0,I_{res}):=\mathcal{N}(\tilde{\mu}_{t-1},\tilde{\Sigma} _{t-1})\label{Eq:34}
\end{align}
Similar to obtaining $p(y)$ from $p(x)$ and $p(y|x)$ using Eq.2.113-Eq.2.115 in~\citep{bishop2006pattern}, the values of $ \tilde{\mu}_{t-1}$ and $\tilde{\Sigma} _{t-1}$ are derived as following:
\begin{align}
   \tilde{\mu}_{t-1}     & =I_0+\bar{\alpha }_{t-1}I_{res}+\sqrt{\bar{\beta }_{t-1}^2 -\sigma _t^2} \frac{(I_0+\bar{\alpha }_t I_{res})-(I_0+\bar{\alpha }_{t}I_{res})}{\bar{\beta }_{t}}=I_0+\bar{\alpha }_{t-1}I_{res}, \\
   \tilde{\Sigma} _{t-1} & =\sigma _t^2\mathbf{I}+(\frac{\sqrt{\bar{\beta }_{t-1}^2 -\sigma _t^2} }{\bar{\beta }_{t}})^2\bar{\beta }_t^2\mathbf{I}=\beta _{t-1}^2\mathbf{I}.
\end{align}
Therefore, $q(I_{t-1}|I_0,I_{res})=\mathcal{N} (I_{t-1};I_0+\bar{\alpha }_{t-1} I_{res},\bar{\beta }_{t-1}^2\mathbf{I})$.
In fact, the case ($t = T$) already holds, thus Eq.~\ref{Eq:33} holds for all $t$.

   {\bf Simplifying Eq.~\ref{Eq:11}.} Eq.~\ref{Eq:11} can also be written as:
\begin{align}
   I_{t-1} & =I_0+\bar{\alpha }_{t-1}I_{res}+\sqrt{\bar{\beta }_{t-1}^2 -\sigma _t^2} \frac{I_t-(I_0+\bar{\alpha }_{t}I_{res})}{\bar{\beta }_{t}} +\sigma _t\epsilon _t,\label{Eq:39}                                                                                                                        \\
           & =\frac{\sqrt{\bar{\beta }_{t-1}^2 -\sigma _t^2} }{\bar{\beta }_{t}}I_t+(1-\frac{\sqrt{\bar{\beta }_{t-1}^2 -\sigma _t^2} }{\bar{\beta }_{t}})I_0+(\bar{\alpha}_{t-1}-\frac{\sqrt{\bar{\alpha}_t\bar{\beta }_{t-1}^2 -\sigma _t^2} }{\bar{\beta }_{t}})I_{res}+\sigma _t\epsilon _t\label{Eq:40} \\
           & =I_t-(\bar{\alpha} _t-\bar{\alpha} _{t-1})I_{res} -(\bar{\beta }_t-\sqrt{\bar{\beta }_{t-1}^2-\sigma _t^2}  )\epsilon +\sigma _t\epsilon _t,\label{Eq:41}
\end{align}
where $\epsilon_{t} \sim \mathcal{N}(\mathbf{0},\mathbf{I})$. Eq.~\ref{Eq:41} is consistent with Eq.~\ref{Eq:13}, and Eq.~\ref{Eq:8} is used for the derivation from Eq.~\ref{Eq:40} to Eq.~\ref{Eq:41}.

\subsection{Coefficient Transformation}\label{Appendix:a.3}
For image generation, $I_{in}=0$, thus Eq.~\ref{Eq:17} can also be written as:
\begin{align}
    & I_t=(\bar{\alpha}_t-1) I_{res}+\bar{\beta}_t\epsilon                     \\
    & I_{res}=\frac{I_t-\bar{\beta}_t\epsilon}{\bar{\alpha}_t-1}.\label{Eq:44}
\end{align}
If the residuals $I_{res}^\theta$ are represented as a transformation of $\epsilon_\theta$ using Eq.~\ref{Eq:44}, Eq.~\ref{Eq:13} is simplified to
\begin{align}
   I_{t-1} & =I_t-(\bar{\alpha} _t-\bar{\alpha} _{t-1})I_{res}^\theta -(\bar{\beta }_t-\sqrt{\bar{\beta }_{t-1}^2-\sigma _t^2}  )\epsilon _\theta +\sigma _t\epsilon _t\label{Eq:45}                                                                                                                          \\
           & =I_t-(\bar{\alpha} _t-\bar{\alpha} _{t-1})\frac{I_t-\bar{\beta}_t\epsilon_\theta}{\bar{\alpha}_t-1} -(\bar{\beta }_t-\sqrt{\bar{\beta }_{t-1}^2-\sigma _t^2}  )\epsilon _\theta +\sigma _t\epsilon _t                                                                                            \\
           & =\frac{1-\bar{\alpha}_{t-1}}{1-\bar{\alpha}_t}I_t-(\frac{1-\bar{\alpha}_{t-1}}{1-\bar{\alpha}_t}\bar{\beta }_t-\sqrt{\bar{\beta }_{t-1}^2-\sigma _t^2}  )\epsilon _\theta +\sigma_t\epsilon _t\label{Eq:47}                                                                                      \\
           & =\frac{\sqrt{\bar{\alpha } _{DDIM}^{t-1}}}{\sqrt{\bar{\alpha } _{DDIM}^t}}I_t-(\frac{\sqrt{\bar{\alpha } _{DDIM}^{t-1}}\sqrt{1-\bar{\alpha } _{DDIM}^{t}}}{\sqrt{\bar{\alpha } _{DDIM}^t}}-\sqrt{1-\bar{\alpha } _{DDIM}^{t-1}-\sigma _t^2}  )\epsilon _\theta +\sigma_t\epsilon _t\label{Eq:48} \\
           & =\sqrt{\bar{\alpha } _{DDIM}^{t-1}}\left (\frac{I_t-\sqrt{1-\bar{\alpha } _{DDIM}^{t}}\epsilon _\theta}{\sqrt{\bar{\alpha } _{DDIM}^{t}}}\right ) +\sqrt{1-\bar{\alpha } _{DDIM}^{t-1}-\sigma _t^2}  \epsilon _\theta +\sigma_t\epsilon _t.\label{Eq:49}
\end{align}
Eq.~\ref{Eq:49} is consistent with Eq.12 in DDIM~\citep{song2020denoising} by replacing $\sigma _t^2$ with $\sigma _t^2(DDIM)$, and Eq.~\ref{Eq:19} is used for the derivation from Eq.~\ref{Eq:47} to Eq.~\ref{Eq:48}. Thus, our sampling process is consistent with that of DDPM~\citep{song2020denoising} and DDIM~\citep{ho2020denoising} by transforming coefficient/variance schedules.

\begin{algorithm}[t]
   \SetKwData{Left}{left}\SetKwData{This}{this}\SetKwData{Up}{up}
   \SetKwFunction{Loss}{Loss}\SetKwFunction{Rmse}{Rmse}\SetKwFunction{Argmin}{Argmin}\SetKwFunction{Aug}{Aug}\SetKwFunction{Linspace}{Linspace}\SetKwFunction{Cumprod}{Cumprod} \SetKwFunction{return}{return}\SetKwFunction{pass}{pass}\SetKwFunction{Power}{Power}\SetKwFunction{Cumsum}{Cumsum}
   \SetKwInOut{Input}{Input}\SetKwInOut{Output}{Output}
   \Input{The initial conditions $\bar{\alpha}_T=1$, $\bar{\beta}_T^2>0$, $T=1000$, and $t\in \left \{ 1,2,\dots T \right \}$. The hyperparameter $\eta=1$ for the random generation process and $\eta=0$ for deterministic implicit sampling. Variance modes have ``DDIM'' and ``DDIM$\to$RDDM''. The coefficient adjustment mode Adjust=``Alpha'', ``Beta'', or ``Alpha+Beta''.}
   \Output{The adjusted coefficients $\bar{\alpha}_t^*$, $\bar{\beta}_t^*$ and $\sigma _t^*$.}
   \BlankLine
   \tcp{(a) Coefficient initialization of DDIM~\citep{song2020denoising}}
   $\beta^t_{DDIM}\leftarrow$ \Linspace(0.0001, 0.02, $T$) \Comment{linear schedule~\citep{song2020denoising}}\\
   $\alpha^t_{DDIM}\leftarrow1-\beta^t_{DDIM}$\\
   $\bar{\alpha}^t_{DDIM}\leftarrow$ \Cumprod($\alpha^t_{DDIM}$) \Comment{cumulative multiplication}\\
   $\sigma _t^2(DDIM)\leftarrow\eta\frac{(1-\bar{\alpha}^{t-1}_{DDIM})}{1-\bar{\alpha}^{t}_{DDIM}} (1-\frac{\bar{\alpha}^{t}_{DDIM}}{\bar{\alpha}^{t-1}_{DDIM}} )$\\
   \tcp{(b) Coefficient transformation from DDIM~\citep{song2020denoising} to our RDDM}
   $\bar{\alpha}_t\leftarrow 1-\sqrt{\bar{\alpha } _{DDIM}^t}$\Comment{Eq.~\ref{Eq:19}}\\
   $\bar{\beta}_t\leftarrow \sqrt{1-\bar{\alpha } _{DDIM}^t}$\Comment{Eq.~\ref{Eq:19}}\\
   $\sigma _t^2(RDDM)\leftarrow \eta\frac{(\bar{\beta}_t^2-\bar{\beta}_{t-1}^2)\bar{\beta}_{t-1}^2}{\bar{\beta}_{t}^2}$\\
   \tcp{(c) Select variance schedule}
   \uIf{Variance==``DDIM''}
   {
      $\sigma _t^*\leftarrow \sqrt{\sigma _t^2(DDIM)}$ \Comment{consistent sampling process with DDIM~\citep{song2020denoising} and DDPM~\citep{ho2020denoising}}
   }\ElseIf{Variance==``DDIM$\to$RDDM''}{
      $\sigma _t^*\leftarrow \sqrt{\sigma _t^2(RDDM)}$ \Comment{sum-constrained variance schedule}
   }
   \tcp{(d) Coefficient adjustment}
   $\alpha_t\leftarrow$\Power($1-t/T$, $1$)  \Comment{linearly decreasing by Eq.~\ref{Eq:20}}\\
   $\beta_t^2\leftarrow$\Power($t/T$, $1$)$\cdot\bar{\beta}_T^2$ \Comment{control the noise intensity in $I_T$ by $\bar{\beta}_T^2$}\\
   \uIf{Adjust==``Alpha''}{
      $\bar{\alpha}_t^*\leftarrow$\Cumsum($\alpha_t$), $\bar{\beta}_t^*\leftarrow\bar{\beta}_t$ \Comment{cumulative sum}
   }\uElseIf{Adjust==``Beta''}{
      $\bar{\alpha}_t^*\leftarrow\bar{\alpha}_t$, $\bar{\beta}_t^*\leftarrow$$\sqrt{\Cumsum(\beta_t^2)}$ \Comment{cumulative sum}
   }\uElseIf{Adjust==``Alpha+Beta''}{
      $\bar{\alpha}_t^*\leftarrow$\Cumsum($\alpha_t$), $\bar{\beta}_t^*\leftarrow$$\sqrt{\Cumsum(\beta_t^2)}$ \Comment{coefficient reinitialization}
   }\Else{
      $\bar{\alpha}_t^*\leftarrow\bar{\alpha}_t$, $\bar{\beta}_t^*\leftarrow\bar{\beta}_t$
   }
   \return $\bar{\alpha}_t^*$, $\bar{\beta}_t^*$, $\sigma _t^*$\Comment{sampling with adjusted coefficients by Eq.~\ref{Eq:13}}\\
   \caption{Coefficient initialization, transformation, and adjustment.}\label{Algorithm1}
\end{algorithm}

\begin{figure*}[t]
   \setlength{\abovecaptionskip}{0.cm}
   \setlength{\belowcaptionskip}{0.cm}
   \centering
   \includegraphics[width=1\linewidth]{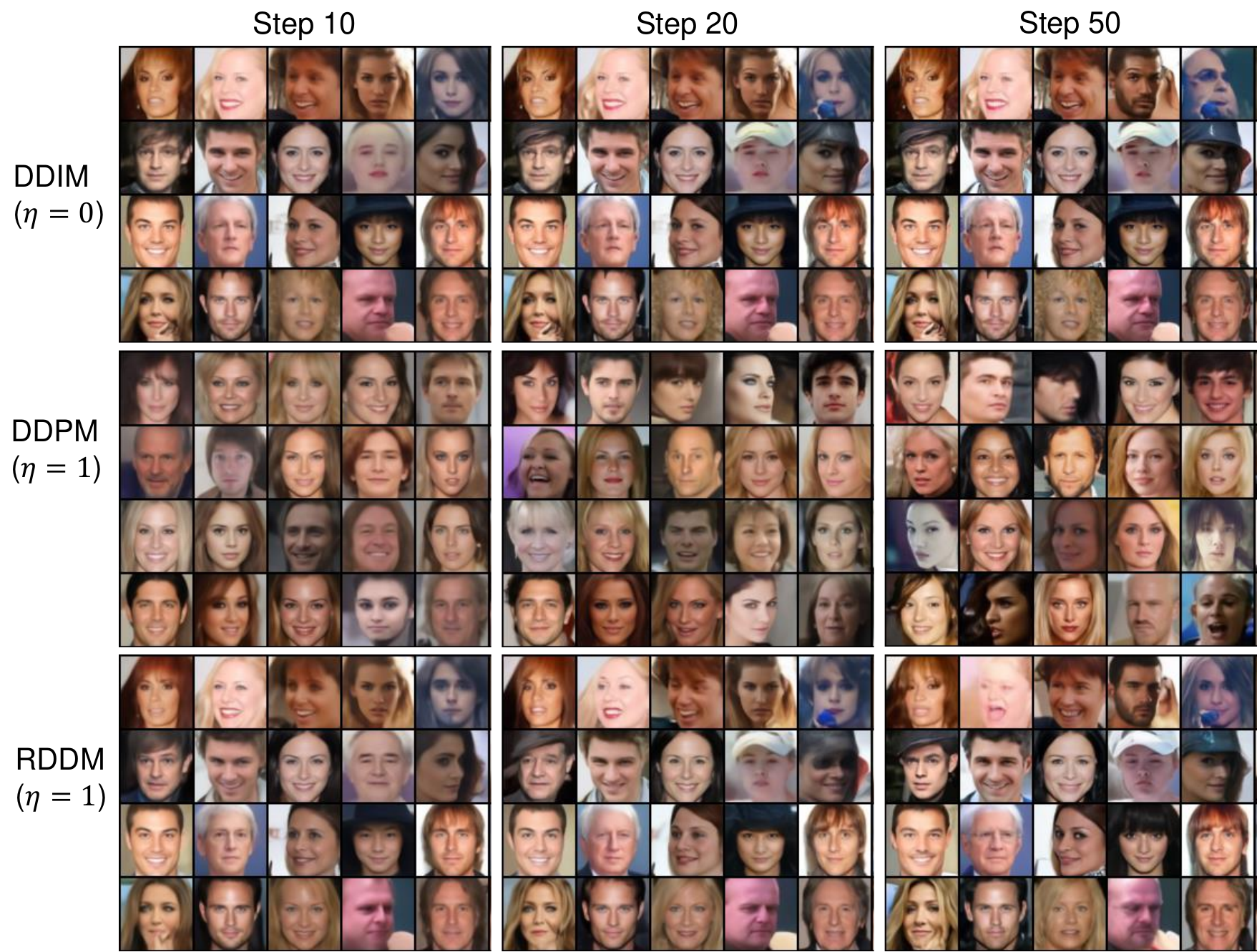}
   \caption{Perturbed generation process with sum-constrained variance on the CelebA 64$\times$64 dataset~\citep{liu2015faceattributes}. When the coefficient schedules ($\bar{\alpha}_t,\bar{\beta}_t$) are consistent, RDDM ($\eta=0$) is the same as DDIM ($\eta=0$), but RDDM ($\eta=1$) is different from DDPM ($\eta=1$) due to different variance schedules ($\sigma_t^2$). At different sampling steps, our RDDM has the same total noise, while DDPM has a different total noise. Notably, all the results in Fig.~\ref{fig:appendix_fig6} can be generated from the same pre-trained model via variance transformation in Appendix~\ref{Appendix:a.3}. In other words, the RDDM provides a sum-constrained variance strategy, which can be used directly in the pre-trained DDPM without re-training the model.\vspace{-0.3cm}}
   \label{fig:appendix_fig6}
\end{figure*}

We present the pipeline of coefficient transformation in Algorithm~\ref{Algorithm1}. Fig.~\ref{fig:alpha} (b)(c) show the result of coefficient transformation. In Eq.~\ref{Eq:19}, in addition to the coefficient $\bar{\alpha}_t,\bar{\beta}_t^2$ being replaced by $\bar{\alpha}_{DDIM}^t$, the variance $\sigma _t^2$ is also replaced with $\sigma _t^2(DDIM)$ to be consistent with DDIM~\citep{song2020denoising} ($\eta=0$) and DDPM~\citep{ho2020denoising} ($\eta=1$).
In fact, for DDIM~\citep{song2020denoising} ($\eta=0$), the variance is equal to 0 and does not need to be converted. Therefore, we analyze the difference between the variance of our RDDM and the variance of DDPM~\citep{ho2020denoising} in Appendix~\ref{Appendix:a.4}.

\subsection{Perturbed Generation Process with Sum-constrained Variance}\label{Appendix:a.4}


From Eq.~\ref{Eq:13}, the variance of our RDDM ($\eta=1$) is
\begin{align}
   \sigma^2_t(RDDM) = \eta\frac{\beta _t^2 \bar{\beta }_{t-1}^2}{\bar{\beta }_{t}^2}.\label{Eq:49-1}
\end{align}
We replace $\bar{\beta}_t^2$ by $\bar{\alpha}_{DDIM}^t$ using Eq.~\ref{Eq:19} and replace $\beta_t^2$ by $\bar{\beta}_t^2-\bar{\beta}_{t-1}^2$,
\begin{align}
   \sigma^2_t(RDDM) = \eta\bar{\alpha } _{DDIM}^{t-1}\frac{ (1-\bar{\alpha } _{DDIM}^{t-1})}{1-\bar{\alpha } _{DDIM}^t}(1-\frac{\bar{\alpha}^{t}_{DDIM}}{\bar{\alpha}^{t-1}_{DDIM}} ),
\end{align}
while the variance of DDPM~\citep{ho2020denoising} ($\eta=1$) is
\begin{align}
   \sigma _t^2(DDIM)=\eta\frac{ (1-\bar{\alpha } _{DDIM}^{t-1})}{1-\bar{\alpha } _{DDIM}^t}(1-\frac{\bar{\alpha}^{t}_{DDIM}}{\bar{\alpha}^{t-1}_{DDIM}} ). \label{Eq:51}
\end{align}
Our variance is much smaller than the variance of DDPM~\citep{ho2020denoising} because $0<\bar{\alpha } _{DDIM}^{t-1}<\alpha_{DDIM}^{1}<1$ (e.g., $\alpha_{DDIM}^{1}=0.02$ in linear schedule~\citep{song2020denoising}). Compared to $\sigma _t^2(DDIM)\approx 1$~\citep{song2020score}, the variance of our RDDM is sum-constrained,
\begin{align}
   \sum_{i=1}^{T} \sigma^2_t(RDDM) =  \sum_{i=1}^{T} \eta\beta _t^2\frac{ \bar{\beta }_{t-1}^2}{\bar{\beta }_{t}^2}\le \sum_{i=1}^{T}\beta _t^2\le 1, \label{Eq:52}
\end{align}
where $\sum_{i=1}^{T}\beta _t^2=\bar{\beta} _T^2=1$ for image generation.
This is also consistent with the directional residual diffusion process with perturbation defined in Eq.~\ref{Eq:8}. A qualitative comparison of our RDDM ($\eta=1$) with DDIM~\citep{song2020denoising} ($\eta=0$) and DDPM~\citep{ho2020denoising} ($\eta=1$) is shown in Fig.~\ref{fig:appendix_fig6}. Notably, for $\eta=0$, our RDDM is consistent with DDIM~\citep{song2020denoising} (in Fig.~\ref{fig:4}(a)(b)).

\subsection{Comparison With Other Methods}\label{Appendix:a.5}

The main difference is that to adapt the denoising diffusion, score, flow, or Schrödinger's bridge to image restoration, they choose the noise (Shadow Diffusion~\citep{guo2023shadowdiffusion}, SR3~\citep{saharia2022image}, and WeatherDiffusion~\citep{ozdenizci2023restoring}), the residual (DvSR~\citep{whang2022deblurring} and Rectified Flow~\citep{liu2023flow}), the target image (ColdDiffusion~\citep{bansal2022cold} and InDI~\citep{delbracio2023inversion}), {\bf or} its linear transformation term (I2SB~\citep{liu20232}), which is similar to a special case of our RDDM when it only predicts noise (SM-N) or residuals (SM-Res), while we introduce residual estimation but also embrace noise both for generation and restoration (SM-Res-N). We highlight that residuals and noise are equally important, e.g., the residual prioritizes certainty while the noise emphasizes diversity.

   {\bf Differences from DDPM~\citep{ho2020denoising}.} 1) DDPM is not interpretable for image restoration, while our RDDM is a unified, interpretable diffusion model for both image generation and restoration. 2) Differences in the definition of the forward process lead to different variance strategies. Our RDDM has sum-constrained variance (much smaller than the variance of DDPM), while DDPM has preserving variance~\citep{song2020score} (see Appendix~\ref{Appendix:a.4}).

{\bf Differences from IDDPM~\citep{nichol2021improved}.} In the original DDPM~\citep{ho2020denoising}, for the transfer probabilities $p_\theta(I_{t-1}|I_t)$ in Eq.~\ref{Eq:3}, the mean $\mu_{\theta}(I_{t}, t)$ is learnable, while the variance $\Sigma_t$ is fixed.
IDDPM~\citep{nichol2021improved} highlights the importance of {\bf estimating both the mean and variance}, demonstrating that learning variances allow for fewer sampling steps with negligible differences in sample quality. However, IDDPM~\citep{nichol2021improved} still only involves denoising procedures, and crucially, IDDPM, like DDPM, is thus not interpretable for image restoration.
In addition, IDDPM~\citep{nichol2021improved} proposes three alternative ways to parameterize $\mu_{\theta}(I_{t}, t)$, i.e., predict mean  $\mu_{\theta}(I_{t}, t)$ directly with a neural network, predict noise $\epsilon$, or predict clean image $I_0$. IDDPM~\citep{nichol2021improved} does not predict the clean image $I_0$ and noise $\epsilon$ at the same time, while both the residuals and the noise are predicted for our SM-Res-N.

The {\bf essential difference} is that, our RDDM contains a mixture of three terms (i.e., input images $I_{in}$, target images $I_0$, and noise $\epsilon$) beyond DDPM/IDDPM (a mixture of two terms, i.e, $I_0$ and $\epsilon$). {\bf We emphasize that residuals and noise are equally important}: the residual prioritizes certainty, while the noise emphasizes diversity. Furthermore, our RDDM preserves the original DDPM generation framework by coefficient/variance transformation (Eq.~\ref{Eq:19}), enabling seamless transfer of improvement techniques from DDPM, such as variance optimization from IDDPM.

   {\bf Differences from ColdDiffusion~\citep{bansal2022cold}.}
1) ColdDiffusion aims to remove the random noise entirely from the diffusion model, and replace it with other transforms (e.g., blur, masking), while our RDDM still embraces noise diffusion. Notably, we argue that noise is necessary for generative tasks that emphasize diversity (see Table~\ref{table:7}). In fact, since ColdDiffusion discards random noise, extra noise injection is required to improve generation diversity.
2) To simulate the degradation process for different restoration tasks, ColdDiffusion attempts to use a Gaussian blur operation for deblurring, a snowification transform for snow removal., etc. These explorations may lose generality and differ fundamentally from our residual learning. RDDM represents directional diffusion from target images to input images using residuals, without designing specific degradation operators for each task. Additionally, RDDM provides solid theoretical derivation, while ColdDiffusion lacks theoretical justification.

   {\bf Differences from DvSR~\citep{whang2022deblurring}.} \cite{whang2022deblurring} indeed use residual. But they 1) predict the initial clean image from a blurring image via a traditional (non-diffusion) network, calculate the residuals between the ground truth of the clean image and the predicted clean image 2) use denoising-based diffusion models predict noise like DDPM~\citep{ho2020denoising} and use a linear transformation of the noise to represent the residuals. They treat the residual predictions as an image generation task, aiming to produce diverse and plausible outputs based on the initial predicted clean image. Beyond simply building a diffusion model on top of residuals, we redefine a new forward process that allows simultaneous diffusion of residuals and noise, wherein the target image progressively diffuses into a purely noise or a noise-carrying input image.

   {\bf Differences from InDI~\citep{delbracio2023inversion} and I2SB~\citep{liu20232}.} We can conclude that the forward diffusion of InDI, I2SB, and our RDDM is consistent in the form of a mixture of three terms (i.e., input images $I_{in}$, target images $I_0$, and noise $\epsilon$) beyond the denoising-based diffusion (a mixture of two terms, i.e, $I_0$ and $\epsilon$). Substituting $I_{res}=I_{in}-I_0$ into Eq.~\ref{Eq:17} results in $I_t=\bar{\alpha_t}I_{in}+(1-\bar{\alpha_t})I_0+\bar{\beta_t}\epsilon$. This resulted $I_t$ has the same format as Eq.8 in InD ($x_t=ty+(1-t)x+\sqrt{t}\epsilon_t\eta_t$), and is the same format as Eq.11 in I2SB. Similar to Eq.~\ref{Eq:19} (from our RDDM to DDPM/DDIM), transforming coefficients leads to complete consistency. However, our RDDM can further extend DDPM/DDIM, InD, and I2SB to independent double diffusion processes, and holds the potential to pave the way for the multi-dimensional diffusion process. From the initial stages of constructing a new forward process, our RDDM uses independent coefficient schedules to control the diffusion of residuals and noise. This provides a more general, flexible, and scalable framework, and inspires our partially path-independent generation process, demonstrated in Fig.~\ref{fig:6} and Fig.~\ref{fig:appendix_fig7}(b-f) with stable generation across various diffusion rates and path variations.

\section{Experiments}\label{Appendix:b.0}

\begin{table}[t]
   \resizebox{1\columnwidth}{!}{
   \setlength{\tabcolsep}{0.1mm}{
      \begin{tabular}{l|c|c|c|c|c|c|c}
         \toprule
         \multirow{2}*{Tasks}     & \multicolumn{4}{c|}{Image Restoration} & Image               & Image                  & Image                                                                                               \\
                                  & Shadow Removal                         & Low-light           & Deblurring             & Deraining               & Generation            & Inpainting               & Translation            \\
         \midrule
         \multirow{2}*{Datasets}  & \multirow{2}*{ISTD}                    & \multirow{2}*{LOL}  & \multirow{2}*{GoPro}   & \multirow{2}*{RainDrop} & \multirow{2}*{CelebA} & \multirow{2}*{CelebA-HQ} & CelebA-HQ              \\
                                  &                                        &                     &                        &                         &                       &                          & AFHQ                   \\
         \hline
         Batch size               & 1                                      & 1                   & 1                      & 1                       & 128                   & 64                       & 64                     \\
         Image size               & 256                                    & 256                 & 256                    & 256                     & 64                    & 64                       & 64                     \\
         $\bar{\beta}_T^2$        & 0.01                                   & 1                   & 0.01                   & 1                       & 1                     & 1                        & 1                      \\
         $I_{in}$                 & $I_{in}$                               & $I_{in}$            & $I_{in}$               & $I_{in}$                & 0                     & 0                        & 0                      \\
         Sampling steps           & 5                                      & 2                   & 2                      & 5                       & 10                    & 10                       & 10                     \\
         Loss type                & $\ell _1$                              & $\ell _1$           & $\ell _1$              & $\ell _1$               & $\ell _2$             & $\ell _2$                & $\ell _2$              \\
         Loss                     & $L_{res}+L_{\epsilon}$                 & $L_{res}$           & $L_{res}+L_{\epsilon}$ & $L_{res}+L_{\epsilon}$  & $L_{\epsilon}$        & $L_{res},L_{\epsilon}$   & $L_{res},L_{\epsilon}$ \\
         Sampling Method          & SM-Res-N-2Net                          & SM-Res              & SM-Res-N-2Net          & SM-Res-N-2Net           & SM-N                  & SM-Res-N-2Net            & SM-Res-N-2Net          \\
         Optimizer                & Adam                                   & Adam                & Adam                   & Adam                    & RAdam                 & RAdam                    & RAdam                  \\
         Learning rate            & 8e-5                                   & 8e-5                & 8e-5                   & 8e-5                    & 2e-4                  & 2e-4                     & 2e-4                   \\
         Training iterations      & 80k                                    & 80k                 & 400k                   & 120k                    & 100k                  & 100k                     & 100k                   \\
         \multirow{2}*{Schedules} & $\alpha_t:P(1-x,1)$                    & $\alpha_t:P(1-x,1)$ & $\alpha_t:P(1-x,1)$    & $\alpha_t:P(1-x,1)$     & $\alpha^t_{DDIM}\to$  & $\alpha_t:P(1-x,1)$      & $\alpha^t_{DDIM}\to$   \\
                                  & $\beta_t^2:P(x,1)$                     & $\beta_t^2:P(x,1)$  & $\beta_t^2:P(x,1)$     & $\beta_t^2:P(x,1)$      & $\alpha_t,\beta_t^2$  & $\beta_t^2:P(x,1)$       & $\alpha_t,\beta_t^2$   \\
         \bottomrule
      \end{tabular}
   }}{\caption{Experimental settings for training our RDDM.``SM-Res-N-2Net'' is described in Appendix~\ref{Appendix:b.3}. Two optimizers can be implemented in $L_{res},L_{\epsilon}$. We use ``SM-Res-N-2Net'' and $L_{res}+L_{\epsilon}$ on the SID-RGB dataset~\citep{xu2020learning}.\label{table:Experimental_settings}}}
   \vspace{-0.3cm}
\end{table}



\subsection{Training Details}\label{Appendix:b.1}
We use a UNet architecture\footnotemark[9]\footnotetext[9]{Our RDDM is implemented by modifying https://github.com/lucidrains/denoising-diffusion-pytorch repository.} for both residual prediction and noise prediction in our RDDM. The UNet settings remain consistent across all tasks, including the channel size (64) and channel multiplier (1,2,4,8). Detailed experimental settings can be found in Table~\ref{table:Experimental_settings}.
Training and testing for all experiments in Table~\ref{table:Experimental_settings} can be conducted on a single Nvidia GTX 3090.

   {\bf Image Generation.} For comparison with DDIM~\cite{song2020denoising}, we convert the $\alpha^t_{DDIM}$ schedule of DDIM~\cite{song2020denoising} into the $\alpha_t$ and $\beta_t^2$ schedules of our RDDM using Eq.~\ref{Eq:19} in Section ~\ref{Sec:4.2} and Section ~\ref{Sec:5.0}. In fact, a better coefficient schedule can be used for training in our RDDM, e.g., $\alpha_t$ (linearly decreasing) and $\beta_t^2$ (linearly increasing) in Table~\ref{table:alpha_schedules}. The quantitative results were evaluated by Frechet Inception Distance (FID) and Inception Score (IS).

   {\bf Image Restoration.} We extensively evaluate our method on several image restoration tasks, including shadow removal, low-light enhancement, image deraining, and image deblurring on 5 different datasets. For fair comparisons, the results of other SOTA methods are provided from the original papers whenever possible.
For all image restoration tasks, the images are resized to 256, and the networks are trained with a batch size of 1. We use shadow masks and shadow images as conditions for shadow removal (similar to~\citep{le2019shadow,zhu2022bijective}), while other image restoration tasks use the degraded image as condition inputs. For low-light enhancement, we use histogram equalization for pre-processing.
To cope with the varying tasks and dataset sizes, we only modified the number of training iterations, $\bar{\beta}_T^2$ and sampling steps (5 steps for shadow removal and deraining, 2 steps for low-light and deblurring) as shown in Table~\ref{table:Experimental_settings}. $\alpha_t$ is initialized using a linearly decreasing schedule (i.e., $P(1-x,1)$ in Eq.~\ref{Eq:20}), while $\beta_t^2$ is initialized using a linearly increasing schedule (i.e., $P(x,1)$).
The quantitative results were evaluated by Mean Absolute Error (MAE), Peak Signal to Noise Ratio (PSNR), Structural Similarity Index Measure (SSIM), and Learned Perceptual Image Patch Similarity (LPIPS)~\citep{zhang2018unreasonable}.

Notably, our RDDM uses an identical UNet architecture and is trained with a batch size of 1 for all these tasks. In contrast, SOAT methods often involve elaborate network architectures, such as multi-stage~\citep{fu2021auto,zamir2021multi,zhu2022efficient}, multi-branch~\citep{cun2020towards}, and GAN~\citep{wang2018stacked,kupyn2019deblurgan,qian2018attentive}, or sophisticated loss functions like the chromaticity~\citep{jin2021dc}, texture similarity~\citep{zhang2019kindling}, and edge loss~\citep{zamir2021multi}.

{\bf Image Inpainting and Image Translation.} To increase the diversity of the generated images, conditional input images were not fed into the deresidual and denoising network (see Fig.~\ref{fig:appendix_inpainting_wo_input}).

\begin{algorithm}[t]
   \SetKwData{Left}{left}\SetKwData{This}{this}\SetKwData{Up}{up}
   \SetKwFunction{Loss}{Loss}\SetKwFunction{return}{return}\SetKwFunction{pass}{pass}\SetKwFunction{Uniform}{Uniform}\SetKwFunction{abs}{abs}\SetKwFunction{InitWight}{InitWight}
   \SetKwInOut{Input}{Input}\SetKwInOut{Output}{Output}\SetKwFunction{Detach}{Detach}
   \Input{A degraded input image $I_{in}$ and its corresponding ground truth image $I_0$. Gaussian noise $\epsilon$. Time condition $t$. Coefficient schedules $\bar{\alpha}$ and $\bar{\beta}_t$. The initial learnable parameters $\lambda_{res}^{\theta}=0.5$. Network $G$ with parameters $\theta$. The initial learning rate $l$. $n$ is the training iterations number. $m$ is the iterations number of AOSA. The threshold of shifting training strategies, $\delta =0.01$.}
   \Output{Trained well parameters, $\theta,\lambda_{res}^{\theta}$.}
   \BlankLine
   $\theta \leftarrow$ \InitWight($G$)\Comment{initialize network parameters}\\
   \For{$i\leftarrow 1$ \KwTo $n+m$}{
   $t \sim$\Uniform ($\{1,2,...,T\}$), $\epsilon \sim \mathcal{N}(\mathbf{0},\mathbf{I})$, $I_{res} \leftarrow I_{in}-I_0$\\
   $I_t \leftarrow I_0+\bar{\alpha}_t I_{res}+\bar{\beta}_t \epsilon$ \Comment{synthesize $I_t$ by Eq.~\ref{Eq:8}}\\
   $I_{out} \leftarrow G(I_t,t,I_{in})$\\
   $I_{res}^{\theta} \leftarrow \lambda_{res}^{\theta}\times I_{out}+(1-\lambda_{res}^{\theta})\times f_{\epsilon \to res}(I_{out})$ \Comment{$f_{\epsilon \to res}(\cdot)$: from $\epsilon$ to $I_{res}$ using Eq.~\ref{Eq:17}}\\
   $\epsilon_{\theta} \leftarrow \lambda_{res}^{\theta}\times f_{res \to \epsilon}(I_{out})+(1-\lambda_{res}^{\theta})\times I_{out}$ \Comment{$f_{res \to \epsilon}(\cdot)$: from $I_{res}$ to $\epsilon$ using Eq.~\ref{Eq:17}}\\
   $L_{auto}\leftarrow$\Loss($I_{res}^{\theta}, I_{res}, \epsilon_{\theta}, \epsilon$) \Comment{based on Eq.~\ref{Eq:50}}\\
   $\theta,\lambda_{res}^{\theta}$ $\overset{+}{\leftarrow}$ $-\nabla _{\theta,\lambda_{res}^{\theta}}(\mathcal{L}_{auto}, l)$ \Comment{updating gradient}\\
   \uIf{\abs($\lambda_{res}^{\theta}-0.5)<\delta $}{
      \pass \Comment{adversarial-like training}
   }\Else{
      $\lambda_{res}^{\theta}\leftarrow$\Detach($\lambda_{res}^{\theta}$)\Comment{halt the gradient updates}\\
      $\theta \leftarrow$ \InitWight($G$)\Comment{reinitialize network parameters}
      \uIf{$\lambda_{res}^{\theta}>0.5$}{
         $\lambda_{res}^{\theta}\leftarrow 1$ \Comment{SM-Res}
      }\Else{
         $\lambda_{res}^{\theta}\leftarrow 0$ \Comment{SM-N}
      }
   }
   }
   \caption{Training Pipeline Using AOSA.}\label{Algorithm2}
\end{algorithm}

\begin{table}[t]
   \resizebox{0.9\columnwidth}{!}{
      \setlength{\abovecaptionskip}{0.0cm}
      \setlength{\belowcaptionskip}{0.0cm}
      \caption{Ablation studies of sampling methods and network structures on the ISTD dataset~\citep{wang2018stacked}. ``SM-Res-N-1Net+One network'' denotes to output 6 channels using a network, where the 0-3-th channels are residual and the 3-6-th channels are noise.}\label{table:AOSA_a}
      \setlength{\tabcolsep}{0.5cm}{
         \begin{tabular}{llccc}
            \toprule
            Sampling Method & Network                          & MAE($\downarrow$) & SSIM($\uparrow$)  & PSNR($\uparrow$)  \\
            \midrule
            SM-Res          & Residual network                 & 4.76              & 0.959             & 30.72             \\
            SM-Res-N-2Net   & Residual network+noise network   & \underline{4.67}  & \underline{0.962} & \underline{30.91} \\
            SM-Res-N-1Net   & One network, only shared encoder & 4.72              & 0.959             & 30.73             \\
            SM-Res-N-1Net   & One network                      & {\bf 4.57}        & {\bf 0.963}       & {\bf 31.10}       \\
            \bottomrule
         \end{tabular}
      }}
\end{table}

\subsection{SAMPLING Details}\label{Appendix:b.3}
In Section ~\ref{Sec:3.3},we described the empirical selection process for SM-Res or SM-N. Below, we give some additional methods in detail.

{\bf Automatic selection of SM-Res or SM-N by AOSA.} We present the motivation, conceptualization, and implementation pipeline (Algorithm~\ref{Algorithm2}) of the Automatic Objective Selection Algorithm (AOSA) as follows:

Step 1. At the initial simultaneous training (similar to SM-Res-N), we do not know whether the network output ($I_{out}$) is residual or noise. Therefore, we set $\lambda_{res}^{\theta}= 0.5$ to denote the probability that the output is residual ($I_{res}^{\theta}$), and $1-\lambda_{res}^{\theta}$ is the probability that the output is noise ($\epsilon _{\theta}$).

\begin{table*}[t]
   \setlength{\abovecaptionskip}{0.cm}
   \setlength{\belowcaptionskip}{0.cm}
   \caption{Generalization analysis of ``SM-Res-N-1Net+One network''.}
   \label{table:Generalization}
   \begin{center}
      \setlength{\tabcolsep}{5mm}{
         \begin{tabular}{lccccc}
            \toprule
            Sampling      & Shadow                              & Deraining                            & Deblurring                          & \multicolumn{2}{c}{Generation}                    \\
            Method        & PSNR/SSIM                           & PSNR/SSIM                            & PSNR/SSIM                           & FID($\downarrow $)             & IS($\uparrow $)  \\
            \midrule
            SM-Res        & 30.72/0.959                         & \underline{31.96}/\underline{0.9509} & \underline{32.32}/\underline{0.957} & 31.47                          & 1.73             \\
            SM-N          & 11.34/0.175                         & 19.15/0.7179                         & 9.49/0.087                          & {\bf 23.25}                    & {\bf 2.05}       \\
            SM-Res-N-2Net      & \underline{30.91}/\underline{0.962} & {\bf 32.51}/{\bf 0.9563}             & {\bf 32.40}/{\bf 0.963}             & 28.90                          & 1.78             \\
            SM-Res-N-1Net & {\bf 31.10}/{\bf 0.963}             & 31.79/0.9504                         & 31.69/0.951                         & \underline{28.57}              & \underline{1.81} \\
            \bottomrule
         \end{tabular}
      }
   \end{center}
   \vspace{-0.5cm}
\end{table*}

\begin{table*}[b]

   \setlength{\abovecaptionskip}{0.cm}
   \setlength{\belowcaptionskip}{0.cm}
   \caption{Shadow removal results on the ISTD dataset~\citep{wang2018stacked}.
      We report the MAE, SSIM and PSNR in the shadow area  (S), non-shadow area (NS), and whole image (ALL).
   }
   \label{table:shadow_all}
   \setlength{\tabcolsep}{1.9mm}{
      \begin{tabular}{l|ccc|ccc|ccc|c}
         \hline
         \multirow{2}*{Method}                 & \multicolumn{3}{c|}{MAE($\downarrow $)} & \multicolumn{3}{c|}{SSIM($\uparrow $)} & \multicolumn{3}{c|}{PSNR($\uparrow $)} & LPIPS($\downarrow $)                                                                                                                          \\
         \cline{2-10}
                                               & S                                       & NS                                     & ALL                                    & S                    & NS                & ALL               & S                 & NS                & ALL                                    \\
         \hline
         ST-CGAN~\citep{wang2018stacked}       & 10.33                                   & 6.93                                   & 7.47                                   & 0.981                & 0.958             & 0.929             & 33.74             & 29.51             & 27.44             & -                  \\
         DSC~\citep{hu2019direction} $\P$      & 9.48                                    & 6.14                                   & 6.67                                   & 0.967                & -                 & -                 & 33.45             & -                 & -                 & -                  \\
         DHAN~\citep{cun2020towards}           & 8.14                                    & 6.04                                   & 6.37                                   & 0.983                & -                 & -                 & 34.50             & -                 & -                 & -                  \\
         CANet~\citep{chen2021canet}           & 8.86                                    & 6.07                                   & 6.15                                   & -                    & -                 & -                 & -                 & -                 & -                 & -                  \\
         LG-ShadowNet~\citep{liu2021shadowTip} & 10.23                                   & 5.38                                   & 6.18                                   & 0.979                & 0.967             & 0.936             & 31.53             & 29.47             & 26.62             & -                  \\
         FusionNet~\citep{fu2021auto}          & 7.77                                    & 5.56                                   & 5.92                                   & 0.975                & 0.880             & 0.945             & 34.71             & 28.61             & 27.19             & 0.1204             \\
         UnfoldingNet~\citep{zhu2022efficient} & 7.87                                    & 4.72                                   & 5.22                                   & 0.987                & \underline{0.978} & 0.960             & {\bf 36.95}       & 31.54             & 29.85             & -                  \\
         BMNet~\citep{zhu2022bijective}        & 7.60                                    & 4.59                                   & 5.02                                   & \underline{0.988}    & 0.976             & 0.959             & 35.61             & 32.80             & 30.28             & 0.0377             \\
         DMTN~\citep{liu2023decoupled}         & \underline{7.00}                        & \underline{4.28}                       & \underline{4.72}                       & {\bf 0.990}          & {\bf 0.979}       & {\bf 0.965}       & 35.83             & \underline{33.01} & \underline{30.42} & \underline{0.0368} \\
         Ours (RDDM)                           & {\bf 6.67}                              & {\bf 4.27}                             & {\bf 4.67}                             & \underline{0.988}    & {\bf 0.979}       & \underline{0.962} & \underline{36.74} & {\bf 33.18}       & {\bf 30.91}       & {\bf 0.0305}       \\
         \hline
      \end{tabular}
   }
\end{table*}

Step 2. We then impose loss constraints on both residual and noise estimation weighted by the learned parameter ($\lambda_{res}^{\theta}$), as follows:
\begin{align}
   L_{auto}(\theta):=\lambda_{res}^{\theta}E\left[\left\|I_{res}-I_{res}^{\theta}(I_t,t,I_{in})\right\|^2\right]+(1-\lambda_{res}^{\theta})E\left[\left\|\epsilon-\epsilon_{\theta}(I_t,t,I_{in})\right\|^2\right].\label{Eq:50}
\end{align}
The joint loss functions $L_{auto}(\theta)$ drive the network to gradually tend to output residuals or noise based on the input. For example, in the image restoration task with deterministic input, it should be simpler for the network to estimate a clear image than noise. In contrast, for the image generation task with random noise input, it is simpler for the network to estimate the noise than a clear image.

Step 3. To enable learning of $\lambda_{res}^{\theta}$, we then include it in the network computation, allowing gradient transmission.
Since $\lambda_{res}^{\theta}$ denotes the probability that the output is residual, the estimated residual $I_{res}^{\theta}$ can be represented as $\lambda_{res}^{\theta}\times I_{out}+(1-\lambda_{res}^{\theta})\times f_{\epsilon \to res}(I_{out})$. $f_{\epsilon \to res}(\cdot)$ represents the transformation from $\epsilon$ to $I_{res}$ using Eq.~\ref{Eq:17}.
Similarly, $\epsilon_{\theta}$ can be represented as $\lambda_{res}^{\theta}\times f_{res \to \epsilon}(I_{out})+(1-\lambda_{res}^{\theta})\times I_{out}$. This is very similar to the cross-entropy loss function.

Step 4. As the training process is completed, our objective should be to estimate only noise (SM-N) or residuals (SM-Res). By utilizing the learned $\lambda_{res}^{\theta}$, we can determine when to switch from an adversarial-like process (residuals vs. noise in Step 2) to a single prediction (residuals or noise). This transition can be controlled, for instance, by setting a condition such as abs($\lambda_{res}^{\theta}-0.5)\geq 0.01$. When the network's tendency to estimate residuals surpasses 51$\%$ probability, we set $\lambda_{res}^{\theta}$ to 1 and halt the gradient updates for $\lambda_{res}^{\theta}$.

The experimental results were consistent with the empirical analysis in Section~\ref{Sec:3.3} and verified the effectiveness of AOSA.
For instance, the initial simultaneous training switches to residual learning (SM-Res) for shadow removal and low-light enhancement in approximately 300 iterations, and to denoising learning (SM-N) for image generation in approximately 1000 iterations.
To summarize, AOSA achieves the same inference cost as the current denoising-based diffusion methods~\citep{ho2020denoising} with the plug-and-play training strategy.

   {\bf SM-Res-N.} Both the residuals and the noise are predicted, which can be implemented with two or one networks.
   {\bf SM-Res-N-2Net.} If computational resources are sufficient, two separate networks can be trained for noise and residual predictions, and the optimal sampling method can be determined during testing. This setting easily obtains a well-suited network for the target task, and facilitates the exploration of the decoupled diffusion process and the partially path-independent generation process in Section~\ref{Sec:4.0}. {\bf SM-Res-N-1Net.} To avoid training two separate networks, another solution is to simply use a joint network (i.e., a shared encoder and decoder) to output 6 channels where the 0-3-th channels are residual and the 3-6-th channels are noise. This setting loses the decoupling property of RDDM, but can achieve dual prediction with a slight cost. Table~\ref{table:AOSA_a} shows that the joint network (i.e., SM-Res-N-1Net+One network) achieves the best shadow removal results (MAE 4.57), even better than two independent networks (4.67). A network with the shared encoder (MAE 4.72) has a slight performance degradation compared to the independent two networks (4.67). We conduct further generalization experiments on Table~\ref{table:Generalization}, indicating ``SM-Res-N-1Net+One network'' isn’t bad, but not always the best.

   \begin{figure*}[t]
      \setlength{\abovecaptionskip}{0.cm}
      \setlength{\belowcaptionskip}{0.cm}
      \centering
      \includegraphics[width=0.85\linewidth]{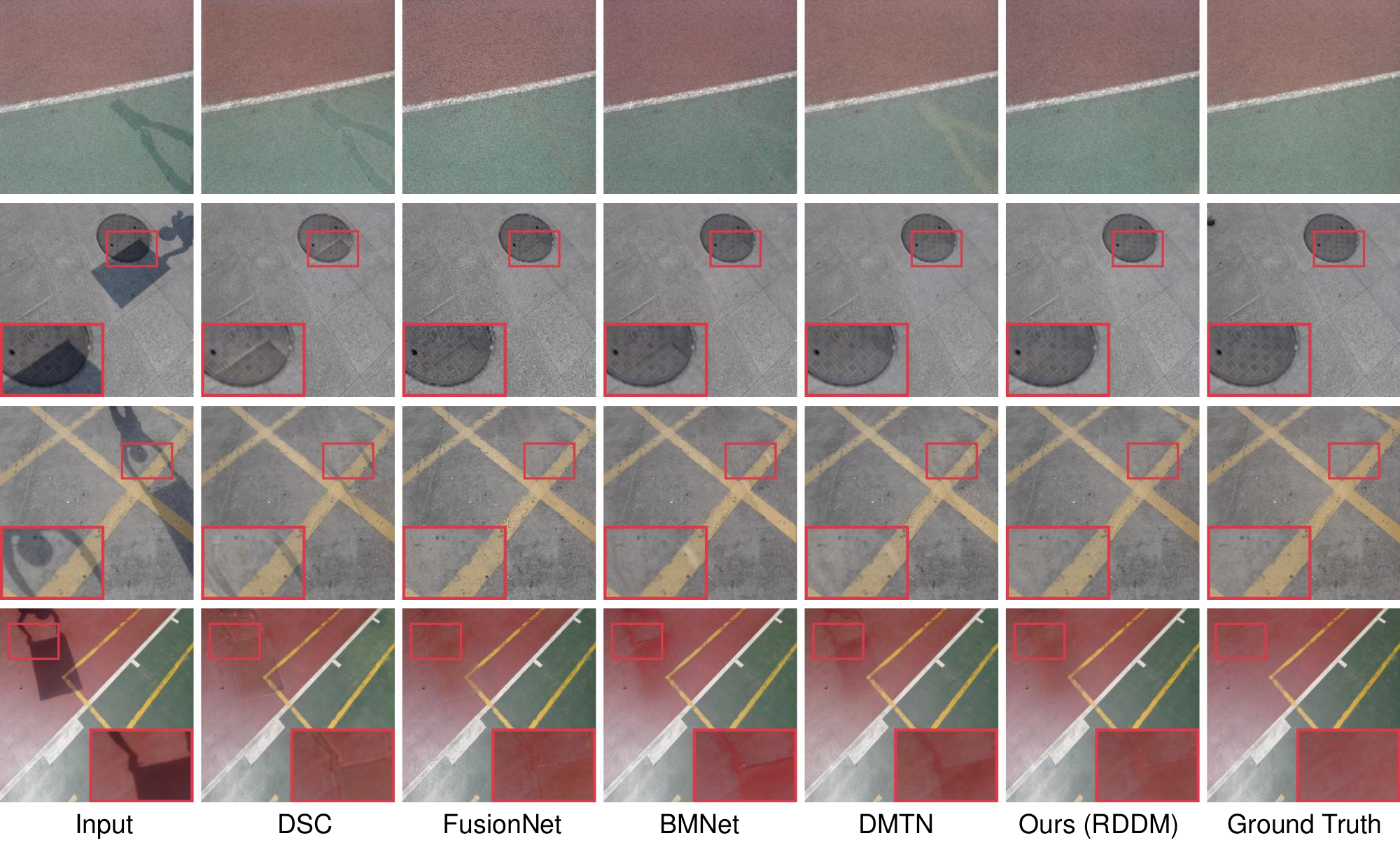}
      \caption{More visual comparison results for shadow removal on the ISTD dataset~\citep{wang2018stacked}.\vspace{-0.5cm}}
      \label{fig:appendix_fig1}
   \end{figure*}

\begin{figure*}[t]
   \setlength{\abovecaptionskip}{0.cm}
   \setlength{\belowcaptionskip}{0.cm}
   \centering
   \includegraphics[width=0.85\linewidth]{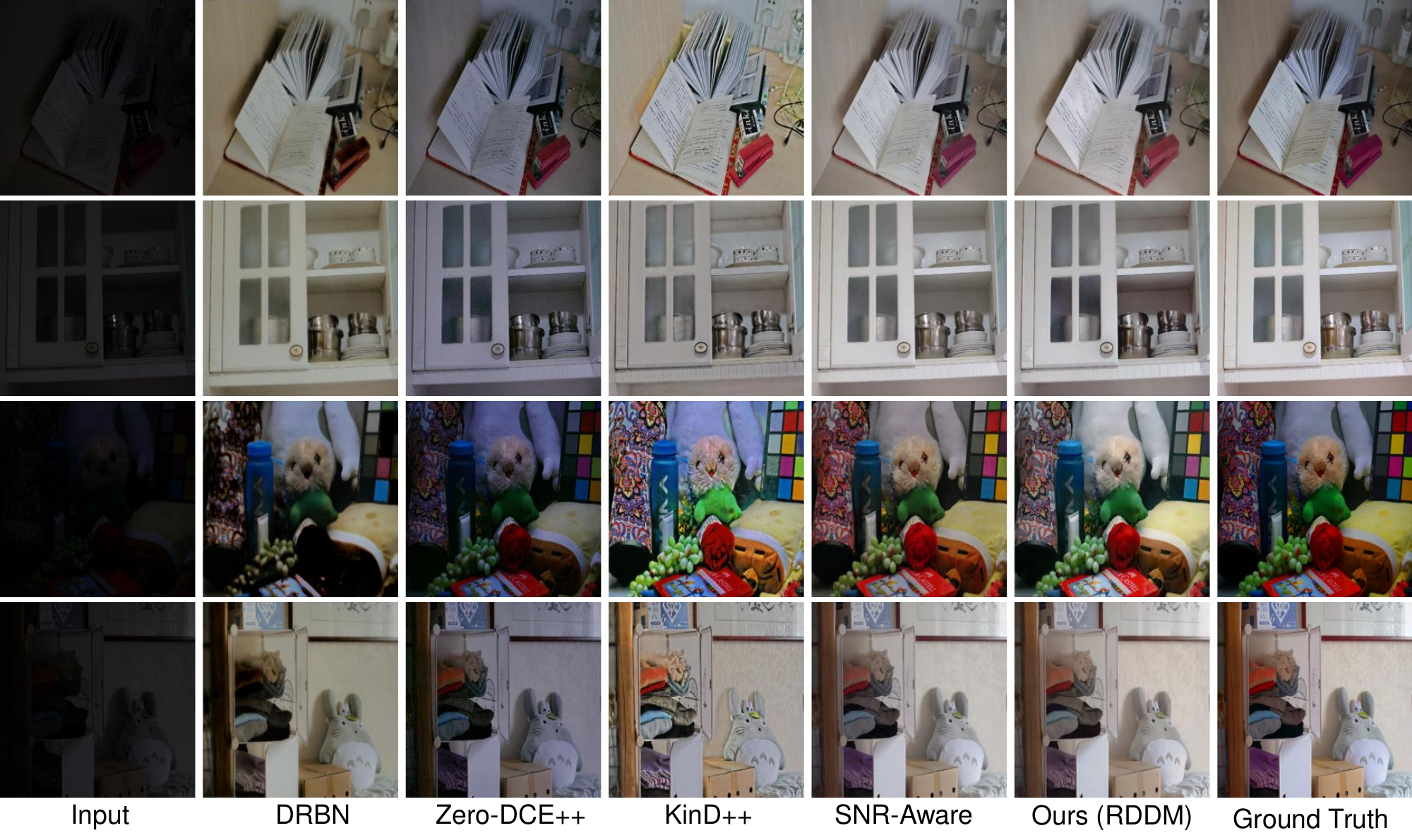}
   \caption{More visual comparison results for Low-light enhancement on the LOL dataset~\citep{wei2018deep}.\vspace{-0.5cm}}
   \label{fig:appendix_fig2}
\end{figure*}

\begin{figure*}[t]
   \setlength{\abovecaptionskip}{0.cm}
   \setlength{\belowcaptionskip}{0.cm}
   \centering
   \includegraphics[width=0.85\linewidth]{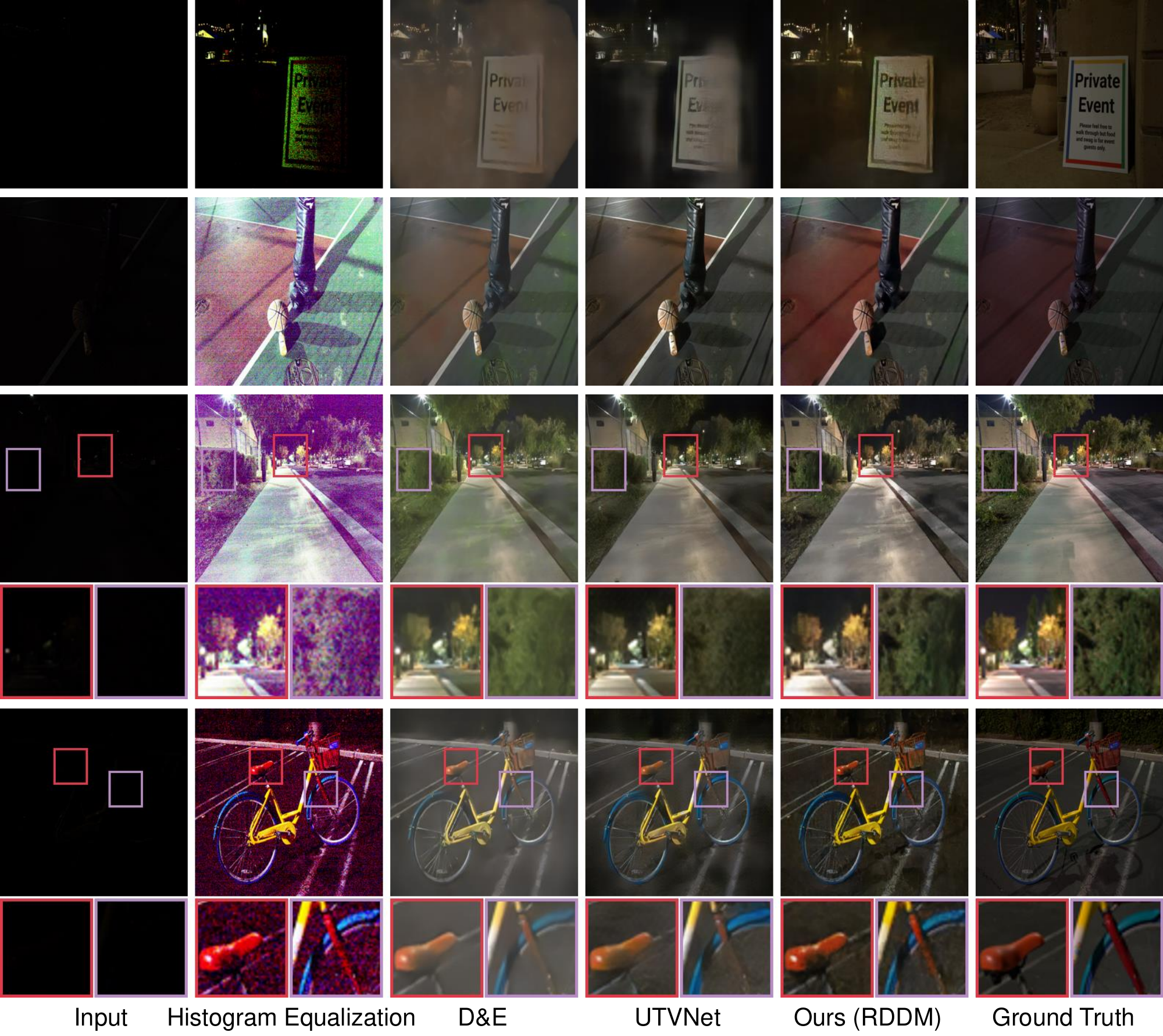}
   \caption{More visual comparison results for Low-light enhancement on the SID-RGB dataset~\citep{xu2020learning}.\vspace{-0.3cm}}
   \label{fig:appendix_fig4}
\end{figure*}

\begin{table*}[t]
   \setlength{\tabcolsep}{5mm}{
      \begin{tabular}{l|ccc}
         \toprule
         Low-light (SID-RGB)                                & PSNR($\uparrow $) & SSIM($\uparrow $)  & LPIPS($\downarrow $) \\
         \midrule
         SID \citep{chen2018learning}                       & 21.16             & 0.6398             & 0.4026               \\
         D\&E \citep{xu2020learning}                        & 22.13             & 0.7172             & 0.3794               \\
         MIR-Net\citep{zamir2020learning,zamir2022learning} & 22.34             & 0.7031             & 0.3562               \\ 
         UTVNet \citep{zheng2021adaptive}                   & 22.69             & \underline{0.7179} & 0.3417               \\
         SNR-Aware \citep{xu2022snr}                        & \underline{22.87} & 0.625              & -                    \\
         Our RDDM (2 step)                                  & \textbf{23.97}    & \textbf{0.8392}    & \underline{0.2433}   \\
         Our RDDM (5 step)                                  & \underline{23.80} & \underline{0.8378} & \textbf{0.2289}      \\
         \bottomrule
      \end{tabular}
   }
   {
      \caption{Quantitative comparison results of Low-light enhancement on the SID-RGB dataset~\citep{xu2020learning}. The results of MIR-Net are reported by \citep{zheng2021adaptive}.}
      \label{table:6a}
   }
\end{table*}

\begin{figure*}[t]
   \setlength{\abovecaptionskip}{0.cm}
   \setlength{\belowcaptionskip}{0.cm}
   \centering
   \includegraphics[width=0.75\linewidth]{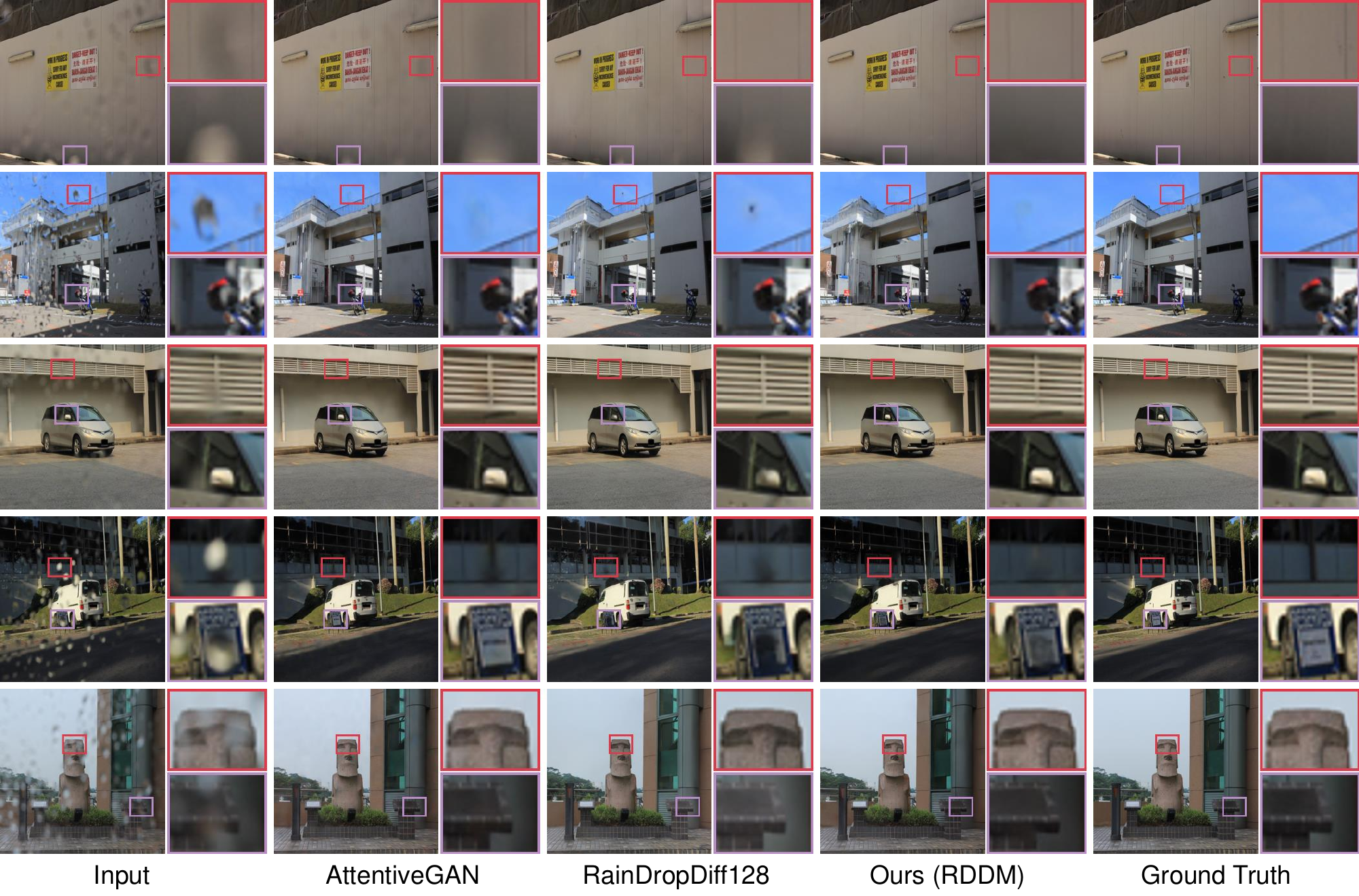}
   \caption{More visual comparison results for deraining on the RainDrop~\citep{qian2018attentive} dataset.\vspace{-0.5cm}}
   \label{fig:appendix_fig3}
\end{figure*}

\begin{figure*}[t]
   \setlength{\abovecaptionskip}{0.cm}
   \setlength{\belowcaptionskip}{0.cm}
   \centering
   \includegraphics[width=0.75\linewidth]{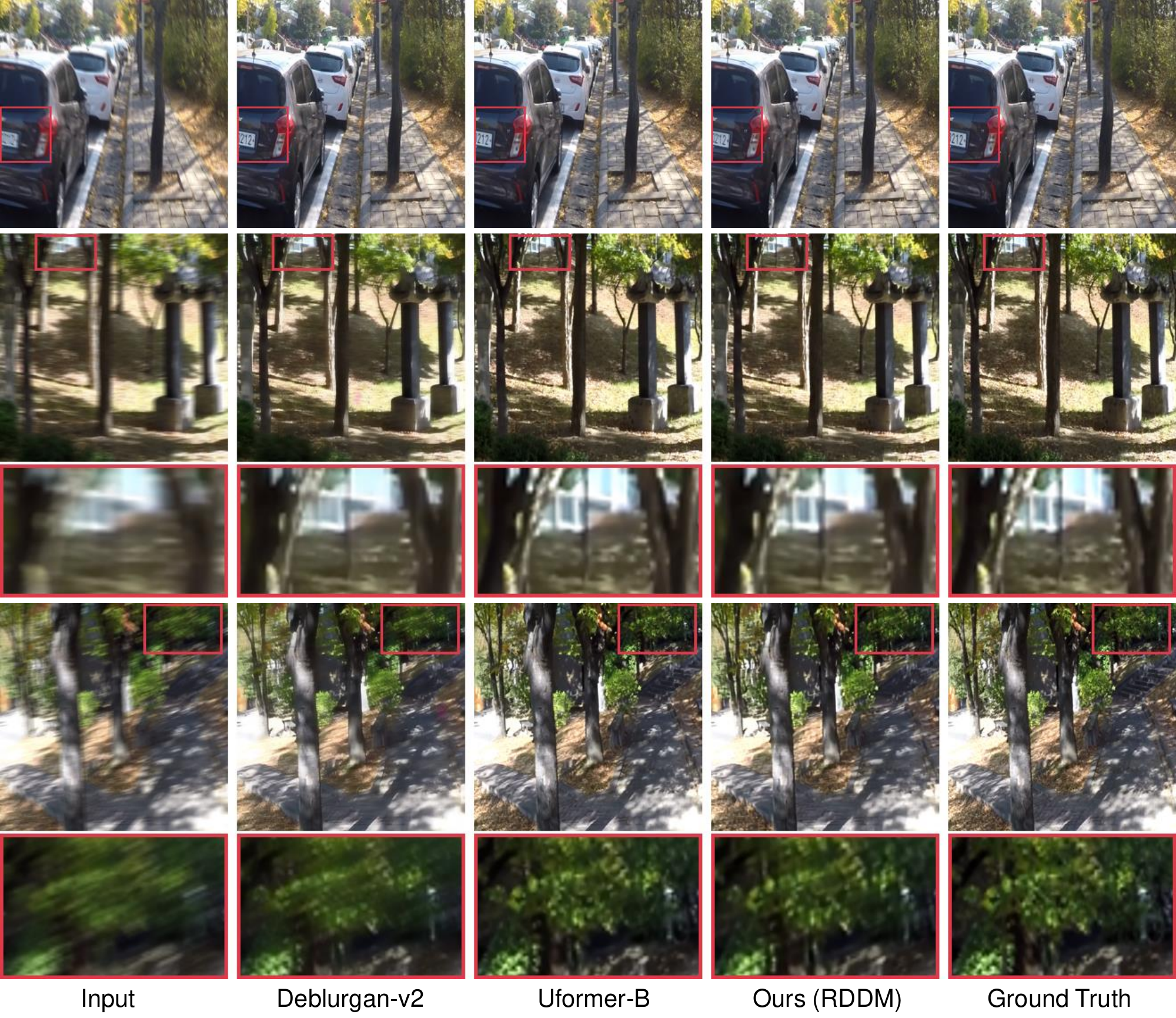}
   \caption{More visual comparison results for deblurring on the GoPro~\citep{nah2017deep} dataset.\vspace{-0.5cm}}
   \label{fig:appendix_fig5}
\end{figure*}

\begin{table*}[t]
   \setlength{\tabcolsep}{5mm}{
      \begin{tabular}{l|ccc}
         \toprule
         Deblurring (GoPro)                             & PSNR($\uparrow $) & SSIM($\uparrow $) & LPIPS($\downarrow $) \\
         \midrule
         Deblurgan-v2~\citep{kupyn2019deblurgan}        & 29.55             & 0.934             & 0.117                \\
         Suin \textit{et al.}~\citep{suin2020spatially} & 31.85             & 0.948             & -                    \\
         MPRNet~\citep{zamir2021multi}                  & \underline{32.66} & 0.959             & 0.089                \\ 
         DvSR~\citep{whang2022deblurring}               & 31.66             & 0.948             & 0.059                \\
         Uformer-B~\citep{wang2022uformer}              & {\bf 32.97}       & {\bf 0.967}       & {\bf 0.0089}         \\
         I2SB~\citep{liu20232}                          & 29.31             & 0.906             & 0.0961               \\
         InDI~\citep{delbracio2023inversion}            & 31.49             & 0.946 & 0.058                \\
         Our RDDM (2 step)                              & 32.40             & \underline{0.963}             & 0.0415               \\
         Our RDDM (10 step)                             & 31.67             & 0.950             & \underline{0.0379}   \\
         \bottomrule
      \end{tabular}
   }
   {
      \caption{Quantitative comparison results of deblurring on the GoPro dataset~\citep{nah2017deep}.}
      \label{table:6b}
   }
\end{table*}

\begin{figure*}[t]
   \setlength{\abovecaptionskip}{0.cm}
   \setlength{\belowcaptionskip}{0.cm}
   \centering
   \includegraphics[width=0.85\linewidth]{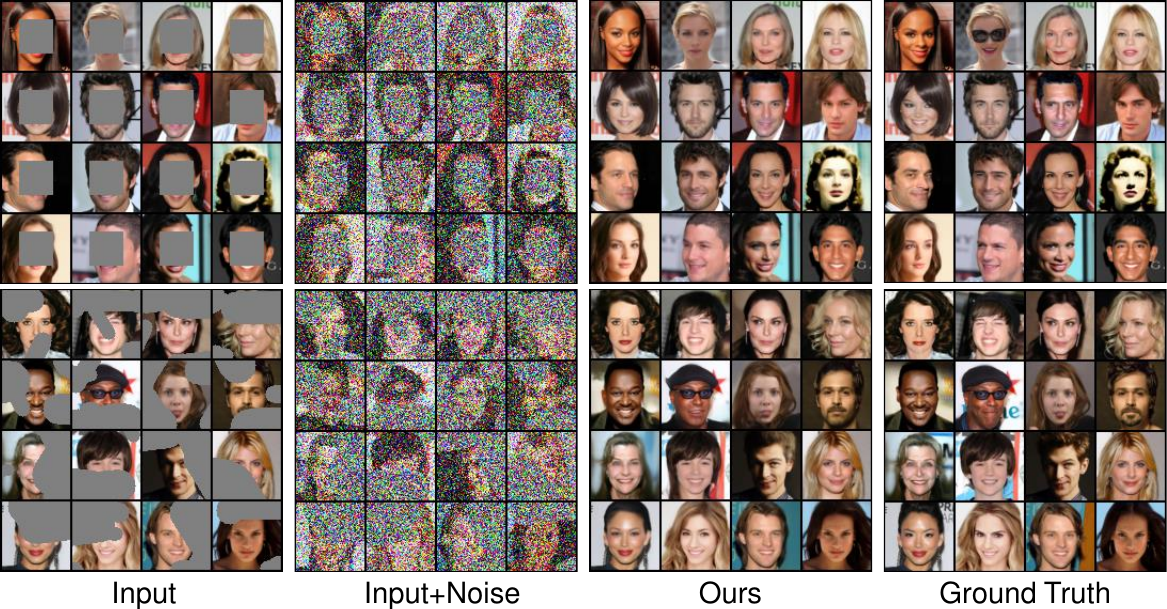}
   \caption{More visual results for image inpainting on the CelebA-HQ~\citep{karras2017progressive} dataset.\vspace{-0.5cm}}
   \label{fig:appendix_inpainting}
\end{figure*}

\begin{figure*}[t]
   \setlength{\abovecaptionskip}{0.cm}
   \setlength{\belowcaptionskip}{0.cm}
   \centering
   \includegraphics[width=1\linewidth]{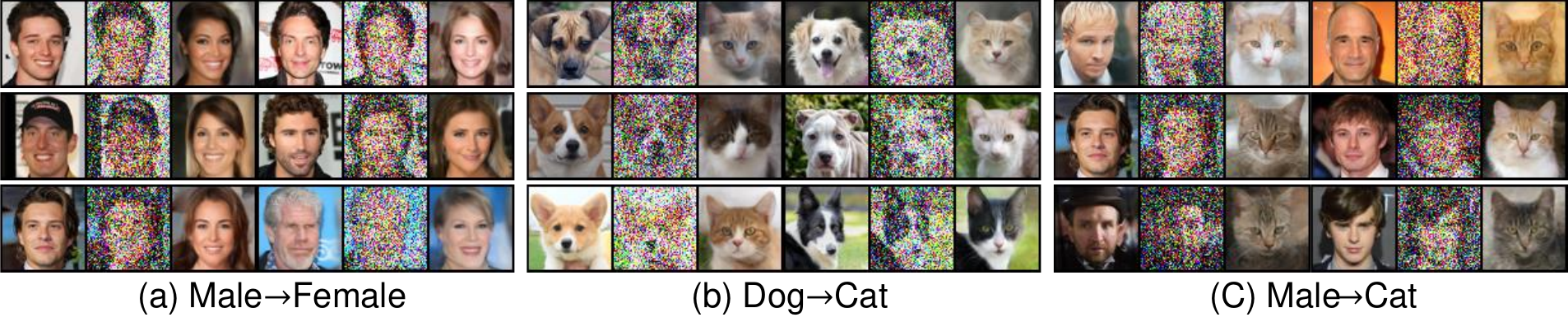}
   \caption{More visual results for image translation on the CelebA-HQ~\citep{karras2017progressive} and AFHQ~\citep{choi2020stargan} datasets.\vspace{-0.3cm}}
   \label{fig:appendix_translation}
\end{figure*}

\subsection{More Results}\label{Appendix:b.2}


{\bf Shadow removal.} We compare RDDM with DSC~\citep{hu2019direction}, FusionNet~\citep{fu2021auto}, BMNet~\citep{zhu2022bijective} and DMTN~\citep{liu2023decoupled} on the ISTD dataset~\citep{wang2018stacked}. The ISTD dataset~\citep{wang2018stacked} contains shadow images, shadow masks, and shadow-free image triplets (1,330 for training; 540 for testing). Table~\ref{table:all}(b), Fig.~\ref{fig:7}(b), and Fig.~\ref{fig:appendix_fig1} demonstrate the superiority of our method. In addition, we compare RDDM with more shadow removal methods (e.g., ST-CGAN~\citep{wang2018stacked}, DHAN~\citep{cun2020towards}, CANet~\citep{chen2021canet}, LG-ShadowNet~\citep{liu2021shadowTip}, UnfoldingNet~\citep{zhu2022efficient}) in Table~\ref{table:shadow_all}.

{\bf Low-light enhancement.} We evaluate our RDDM on the LOL datasets~\citep{wei2018deep} (500 images) and SID-RGB~\citep{xu2020learning} dataset (5,094 images), and compare our method with the current SOTA methods~\citep{zhang2021beyond,liu2021retinex,wu2023skf,zhang2022deep,xu2022snr,zamir2022learning,zheng2021adaptive}.
To unify and simplify the data loading pipeline for training, we only evaluate the RGB low-light image dataset~\citep{wei2018deep,xu2020learning}, not the RAW datasets (e.g., FiveK~\citep{fivek}). Table~\ref{table:all}(c), Fig.~\ref{fig:7}(c), and Fig.~\ref{fig:appendix_fig2} show that our RDDM achieves the best SSIM and LPIPS~\citep{zhang2018unreasonable} and can recover better visual quality on the LOL~\citep{wei2018deep} dataset. Table~\ref{table:6a} shows the low-light enhancement results on the SID-RGB~\citep{xu2020learning} dataset of different methods. Our RDDM outperforms the state-of-the-art SNR-Aware~\citep{xu2022snr} by a ${\bf 4.8\%}$ PSNR and a ${\bf 34.2\%}$ SSIM improvement on the SID-RGB~\citep{xu2020learning} dataset. Fig.~\ref{fig:appendix_fig4} shows that our RDDM outperforms competitors in detail recovery (sharper text of 1st row), and color vibrancy (2nd \& 3rd rows), avoiding issues like gray shading and detail blurring.

   {\bf Image deraining.} We make a fair comparison with the current SOTA diffusion-based image restoration method - RainDiff128~\citep{ozdenizci2023restoring} (``128'' denotes the 128$\times$128 patch size for training) on the RainDrop dataset~\citep{qian2018attentive} (1119 images). RainDiff128~\citep{ozdenizci2023restoring} feeds the degraded input image as a condition to the denoising network, which requires 50 sampling steps to generate a clear image from the noise, while our RDDM requires only 5 sampling steps to recover the degraded image from the noise-carrying input image and outperforms RainDiff128~\citep{ozdenizci2023restoring}, as shown in Table~\ref{table:all}(d) and Fig.~\ref{fig:appendix_fig3}.

{\bf Image deblurring.} We evaluate our method on the widely used deblurring dataset - GoPro~\citep{nah2017deep} (3,214 images). Table~\ref{table:6b} shows that our RDDM is the second best LPIPS.
Fig.~\ref{fig:appendix_fig5} shows that our method is competitive with the SOTA deblurring methods (e.g, Uformer-B~\citep{wang2022uformer}) in terms of visual quality.

   {\bf Image Inpainting and Image Translation.} We show more qualitative results of image inpainting (Fig.~\ref{fig:appendix_inpainting}) and translation (Fig.~\ref{fig:appendix_translation}).

\begin{figure*}[t]
   \setlength{\abovecaptionskip}{0.cm}
   \setlength{\belowcaptionskip}{0.cm}
   \centering
   \includegraphics[width=1\linewidth]{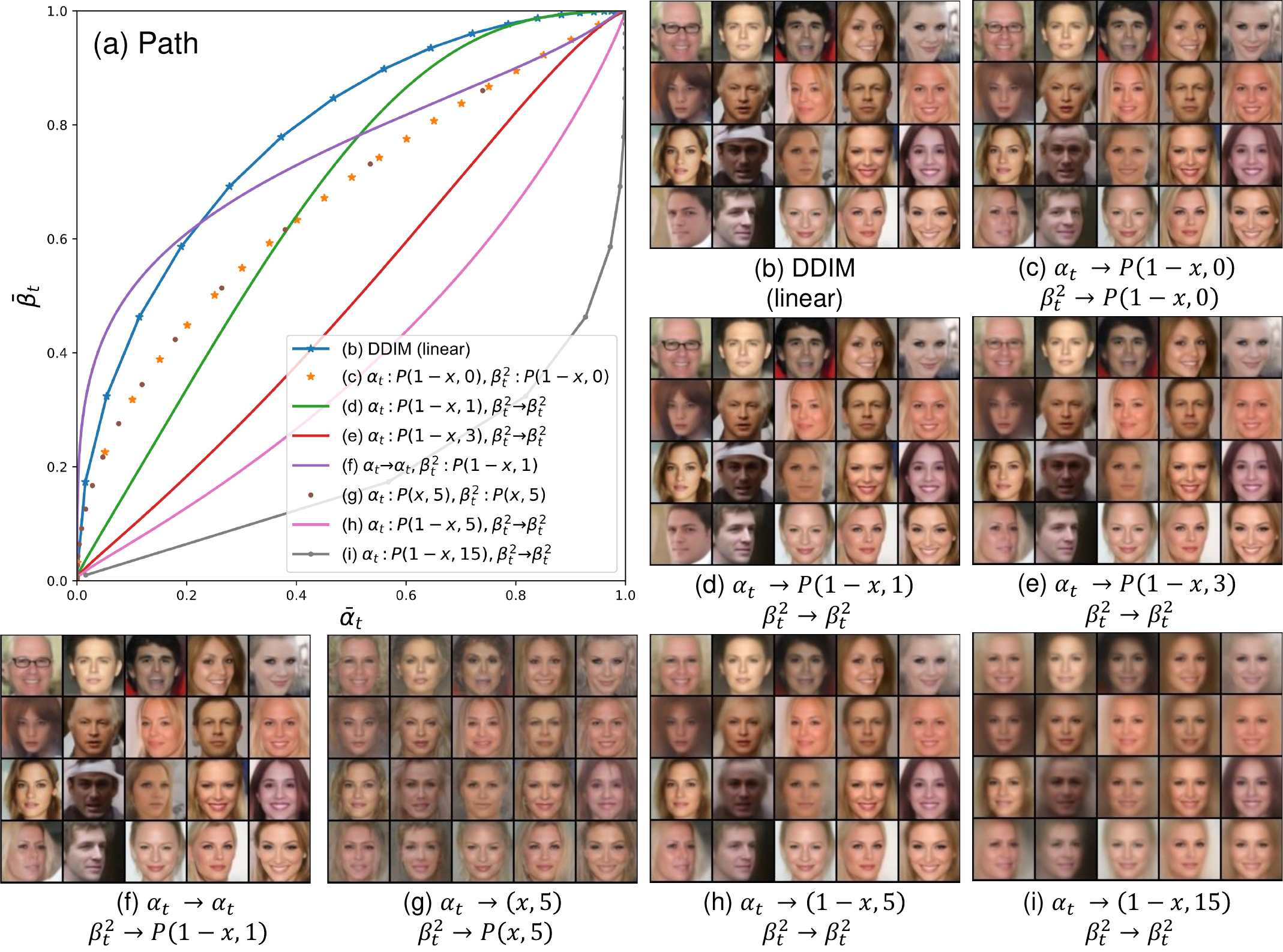}
   \caption{More visual results for the partially path-independent generation process. Two networks are used to estimate residuals and noise separately, i.e., $I_{res}^\theta(I_t,\bar{\alpha}_t\cdot T)$ and $\epsilon_\theta(I_t,\bar{\beta}_t\cdot T)$ ($\eta=0$).\vspace{-0.5cm}}
   \label{fig:appendix_fig7}
\end{figure*}
\begin{figure*}[t]
   \setlength{\abovecaptionskip}{0.cm}
   \setlength{\belowcaptionskip}{0.cm}
   \centering
   \includegraphics[width=1\linewidth]{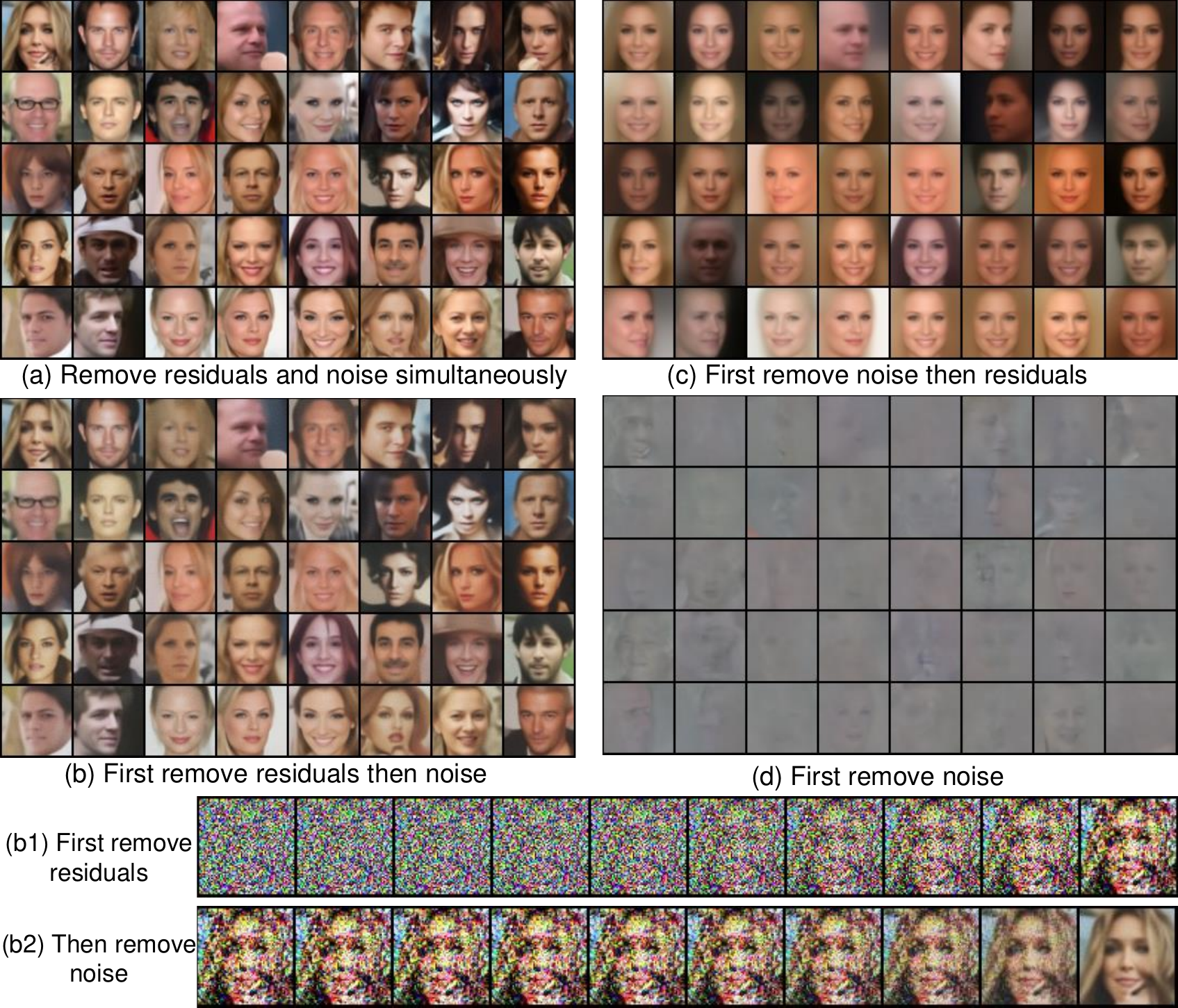}
   \caption{Special paths of the partially path-independent generation process. Two networks are used to estimate residuals and noise separately, i.e., $I_{res}^\theta(I_t,\bar{\alpha}_t\cdot T), \epsilon_\theta(I_t,\bar{\beta}_t\cdot T)$ ($\eta=0$).\vspace{-0.3cm}}
   \label{fig:appendix_fig8}
\end{figure*}

\subsection{Partially Path-independent Generation Process}\label{Appendix:c.1}
Fig.~\ref{fig:appendix_fig7}(b-f) provides evidence supporting the partially path-independent generation process, demonstrating the robustness of the generative process within a certain range of diffusion rates (step size per step) and path variation, e.g., converting DDIM~\citep{song2020denoising} to a uniform diffusion speed in Fig.~\ref{fig:appendix_fig7}(c). However, excessive disturbances can result in visual inconsistencies, as depicted in Fig.~\ref{fig:appendix_fig7}(h)(i).
Furthermore, Fig.~\ref{fig:appendix_fig7}(c) and Fig.~\ref{fig:appendix_fig7}(g) illustrate that even when the paths are the same, the variation in diffusion speed significantly impacts the quality of the generated images. This highlights the importance of carefully considering and controlling the diffusion speed and generation path during the generation process.

We also investigated two reverse paths to gain insight into the implications of the proposed partial path independence. In the first case, the residuals are removed first, followed by the noise: $I(T)\overset{-I_{res}}{\rightarrow} I(0)+\bar{\beta }_T\epsilon \overset{-\bar{\beta }_T\epsilon }{\rightarrow}I(0)$, as shown in Fig.~\ref{fig:appendix_fig8}(b1)(b2). The second case involves removing the noise first and then the residuals: $I(T)\overset{-\bar{\beta }_T\epsilon }{\rightarrow} I_{in} \overset{-I{res}}{\rightarrow} I(0)$. In the first case, images are successfully generated (as shown in Fig.~\ref{fig:appendix_fig8}(b)) which exhibit a striking similarity to the default images in Fig.~\ref{fig:appendix_fig8}(a).
However, the second case shown in Fig.~\ref{fig:appendix_fig8}(c) fails to go from $I_{in}$ to $I(0)$ due to $I_{in}=0$ in the generation task. Figure ~\ref{fig:appendix_fig8}(d) shows the intermediate visualization results of removing the noise first.

\subsection{Ablation Studies}\label{Appendix:b.4}
We have analyzed the sampling method in Table~\ref{table:7}, the coefficient schedule in Table~\ref{table:alpha_schedules}, and the network structure for SM-Res-N in Table~\ref{table:AOSA_a}.

\begin{figure}[t]
   \setlength{\abovecaptionskip}{0cm}
   \centering
   \includegraphics[width=0.7\linewidth]{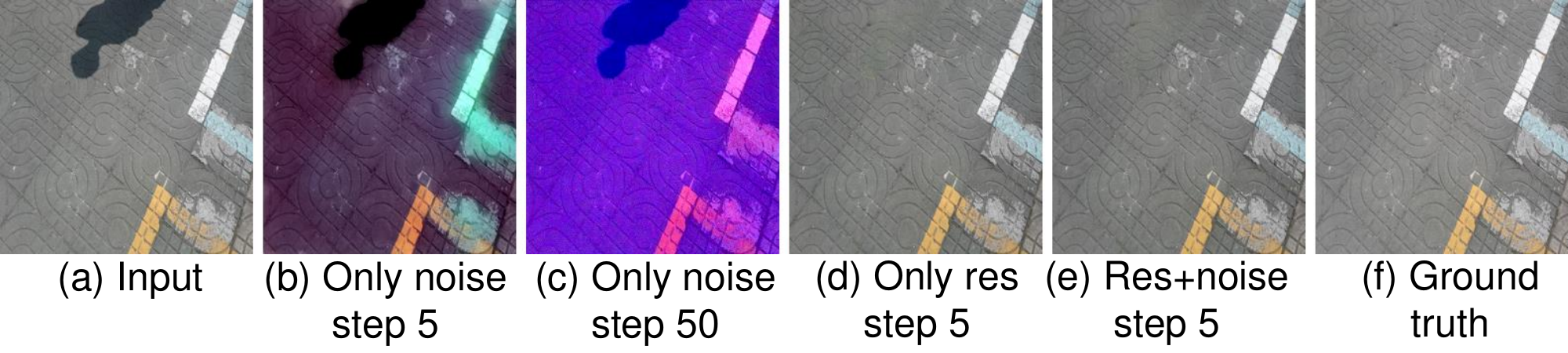}
   \caption{Visualizing ablation studies of sampling methods.}
   \label{fig:a1}
   \vspace{-0.3cm}
\end{figure}

\begin{figure*}[h]
   \setlength{\abovecaptionskip}{0.cm}
   \setlength{\belowcaptionskip}{0.cm}
   \centering
   \includegraphics[width=0.5\linewidth]{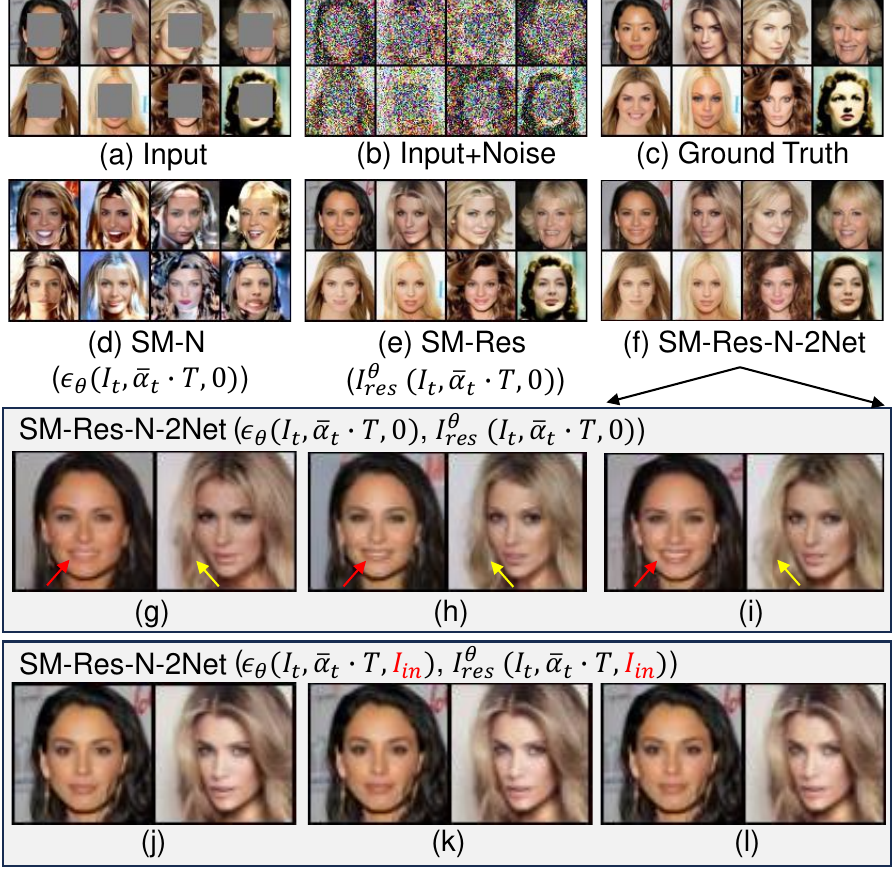}
   \caption{Visualizing ablation studies of sampling methods for image inpainting on the CelebA-HQ~\citep{karras2017progressive} dataset. (g-i) The conditional input image (a) is not used as an input to the deresidual and denoising network in the generation process from (b) to (f). Compared to (g-i), the diversity of the generated images in (j-l) decreases.\vspace{-0.5cm}}
   \label{fig:appendix_inpainting_wo_input}
\end{figure*}

{\bf Sampling Methods.} We present the results for noise predictions only (SM-N) in Fig.~\ref{fig:a1}. Fig.~\ref{fig:a1} (b) and (c) illustrate that estimating only the noise poses challenges as colors are distorted, and it becomes difficult to retain information from the input shadow image. We found that increasing sampling steps does not lead to improved results from Fig.~\ref{fig:a1} (b) to Fig.~\ref{fig:a1} (c), which may be an inherent limitation when estimating only the noise for image restoration. Actually, this is also reflected in DeS3~\citep{jin2022des3} (a shadow removal method based on a denoising diffusion model), where DeS3~\citep{jin2022des3} specifically designs the loss against color bias. Additionaly, training with batch size 1 may contribute to poor results of only predicting noise. However, estimating only the residuals (SM-Res) with batch size 1 does not exhibit such problems for image restoration, as demonstrated in Fig.~\ref{fig:a1} (d)\&(e) and Table~\ref{table:7}, further demonstrating the merits of our RDDM. For image inpainting, SM-Res-N-2Net can generate more realistic face images compared to SM-N and SM-Res, as shown in Fig.~\ref{fig:appendix_inpainting_wo_input}(d-f). If computational resources are sufficient, to obtain better image quality for an unknown task, we suggest that two separate networks can be trained for noise and residual predictions, and the optimal sampling method can be determined
during testing. If computational resources are limited, the sampling method can be determined empirically (see Section ~\ref{Sec:3.3}).

\begin{table}[h]
   \resizebox{0.85\columnwidth}{!}{
      \setlength{\tabcolsep}{6mm}{
         \begin{tabular}{lccccccc}
            \toprule
            RDDM (SM-Res-N-2Net)                       & metric                  & 1 step & 2 step      & 5 step       & 10 step    & 100 step \\
            \midrule
            \multirow{3}*{$\bar{\beta}_T^2=0.01$} & MAE-ALL ($\downarrow $) & 4.83   & 4.69        & {\bf 4.67}   & 4.72       & 4.90     \\
                                                  & PSNR-S ($\uparrow $)    & 36.83  & {\bf 36.98} & 36.74        & 36.59      & 36.41    \\
                                                  & LPIPS ($\downarrow $)   & 0.0344 & 0.0308      & {\bf 0.0305} & 0.0314     & 0.0334   \\
            \hline
            \multirow{3}*{$\bar{\beta}_T^2=1$}    & MAE-ALL ($\downarrow $) & 5.07   & 4.94        & 4.90         & {\bf 4.87} & 4.99     \\
                                                  & PSNR-S ($\uparrow $)    & 36.93  & {\bf 37.20} & 37.07        & 37.01      & 36.62    \\
                                                  & LPIPS ($\downarrow $)   & 0.0346 & 0.0314      & {\bf 0.0298} & 0.0300     & 0.0319   \\
            \bottomrule
         \end{tabular}
      }{
         \caption{Ablation studies of sampling steps for shadow removal on the ISTD dataset~\citep{wang2018stacked}.}
         \label{table:sampling_steps}
      }}
   \vspace{-0.3cm}
\end{table}

{\bf Sampling Steps.} Table~\ref{table:sampling_steps} shows that our RDDM performance improves as the number of sampling steps increases for shadow removal. Unlike the findings in InDI~\citep{delbracio2023inversion} where one step yields the best MSE reconstruction, RDDM achieved its best MAE at 5 steps for shadow removal, which may be due to differences in coefficient schedules and sampling methods.

{\bf Certainty and diversity.} Indeed, feeding conditional input images (Fig.~\ref{fig:appendix_inpainting_wo_input}(a)) into the deresidual and denoising network enhances the certainty of the generated images, while diminishing diversity, as shown in Fig.~\ref{fig:appendix_inpainting_wo_input}(j-l). Generating a clear target image directly from a noisy-carrying degraded image (Fig.~\ref{fig:appendix_inpainting_wo_input}(b)) without any conditions increases diversity, but changes non-missing regions (Fig.~\ref{fig:appendix_inpainting_wo_input}(g-i)).

\begin{figure*}[h]
   \setlength{\abovecaptionskip}{0cm}
   \centering
   \includegraphics[width=1\linewidth]{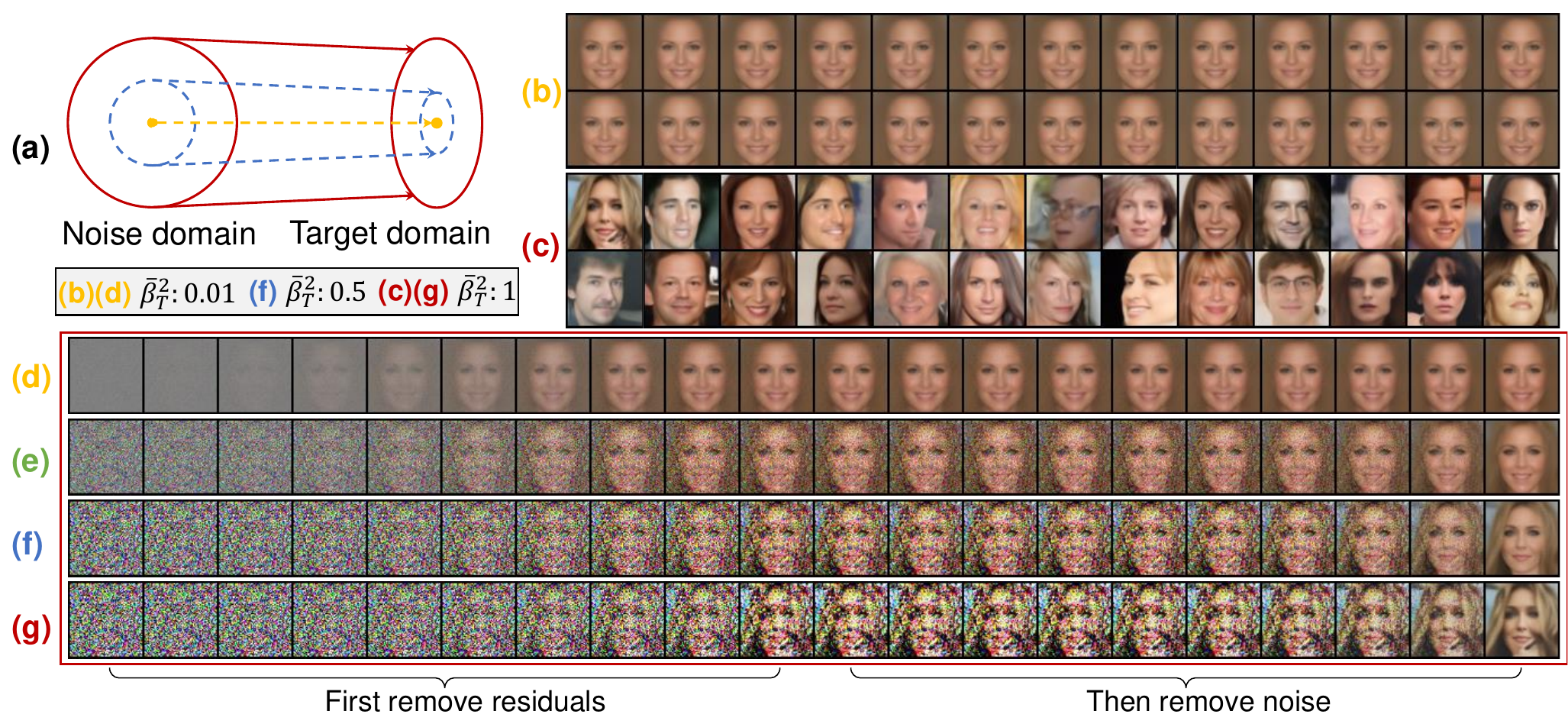}
   \caption{Visualizing ablation studies of noise perturbation intensity ($\bar{\beta}_T^2$). (a) We change the variance ($\bar{\beta}_T^2$) during testing, specifically by variance transformation via Algorithm~\ref{Algorithm1}. (b-c) When $\bar{\beta}_T^2$ decreases from 1 in (c) to 0.01 in (b), the diversity of the generated images decreases significantly. (d-g) We visualize each step in the generation process. $\bar{\beta}_T^2=0.01$ in (d), $\bar{\beta}_T^2=0.1$ in (e), $\bar{\beta}_T^2=0.5$ in (f), and $\bar{\beta}_T^2=1$ in (g). The sampling method is SM-Res-N-2Net with 10 sampling steps.}
   \label{fig:appendix_noise}
   \vspace{-0.3cm}
\end{figure*}

\begin{figure*}[h]
   \setlength{\abovecaptionskip}{0cm}
   \centering
   \includegraphics[width=1\linewidth]{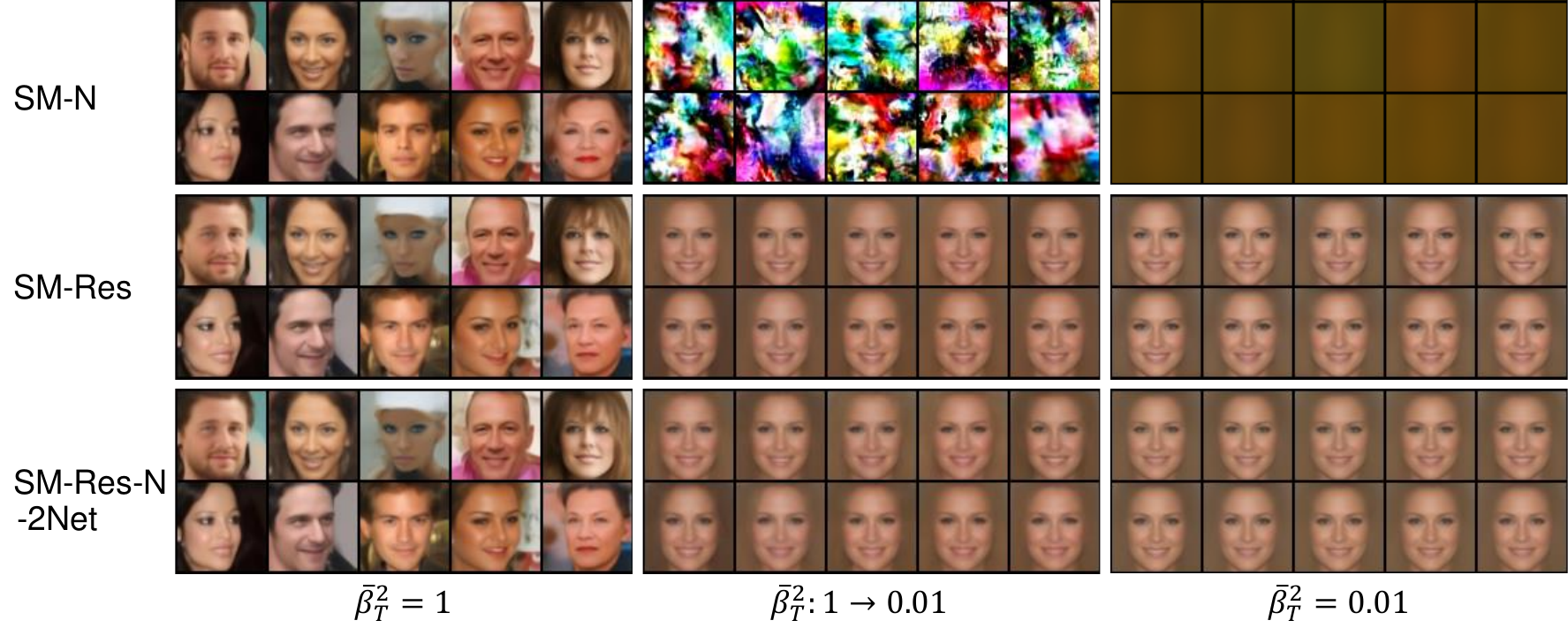}
   \caption{Visualizing ablation studies of sampling methods with different intensities of noise perturbation ($\bar{\beta}_T^2$). ``$\bar{\beta}_T^2:1\to0.01$'' denotes that the variance ($\bar{\beta}_T^2$) is changed during testing by variance transformation via Algorithm~\ref{Algorithm1}. The sampling steps are 10.}
   \label{fig:appendix_res_noise}
   \vspace{-0.3cm}
\end{figure*}

{\bf Noise Perturbation Intensity.} Table~\ref{table:sampling_steps} shows that noise might play a beneficial role in restoring image details and enhancing perceptual quality. For image generation, the diversity of the generated images decreases significantly as $\bar{\beta}_T^2$ decreases from 1 in Fig.~\ref{fig:appendix_noise}(c) to 0.01 in Fig.~\ref{fig:appendix_noise}(b). The experiment is related to the mean face~\citep{wilson2002synthetic,loffler2005fmri,hu2021domain,meng2021magface} and could provide useful insights to better understand the generative process.
Fig.~\ref{fig:appendix_res_noise} shows that modifying $\bar{\beta}_T^2$ during testing ($\bar{\beta}_T^2:1\to0.01$) causes SM-N to fail to generate meaningful faces. SM-Res-N-2Net including deresidual and denoising networks can generate meaningful face images like SM-Res, indicating that the denoising network can perform denoising when modifying $\bar{\beta}_T^2$, but cannot obtain robust residuals ($I_{res}^\theta$) in the sampling process by Eq.~\ref{Eq:17}. In summary, the deresidual network is relatively robust to noise variations compared to the denoising network.


\begin{table}[h]
   \setlength{\abovecaptionskip}{0.cm}
   \setlength{\belowcaptionskip}{0.cm}

   \caption{Resource efficiency and performance analysis by THOP. ``MAC'' means multiply-accumulate operation. (a) Low-light enhancement on the LoL dataset~\citep{wei2018deep}. (b) Shadow removal on the ISTD dataset~\citep{wang2018stacked}. For a fair comparison, a priori shadow mask are used in SR3 with a batch size of 1. (c) Deraining on the RainDrop dataset~\citep{qian2018attentive}.\label{table:Resource}}
   \vspace{-3mm}
   \begin{center}
      \resizebox{0.8\columnwidth}{!}{
         \setlength{\tabcolsep}{1.2mm}{
            \begin{tabular}{l|c|c|c|c|c|c}
               \toprule
               (a) Low-light & PSNR($\uparrow$)  & SSIM($\uparrow$) & LPIPS($\downarrow$) & Params(M)         & MACs(G)$\times$Steps             & Inference Time(s)                \\
               \midrule
               LLformer      & 23.649            & 0.816            & 0.169               & 24.51             & \textbf{22.0$\times$1 = 22.0}    & 0.09$\times$1 = 0.09             \\
               LLFlow        & \underline{25.19} & \underline{0.93} & \textbf{0.11}       & \underline{17.42} & 286.33$\times$1 = 286.3          & \underline{0.18$\times$1 = 0.18} \\
               Ours(RDDM)    & \textbf{25.392}   & \textbf{0.937}   & \underline{0.116}   & \textbf{7.73}     & \underline{32.9$\times$2 = 65.8} & \textbf{0.03$\times$2 = 0.06}    \\
               \bottomrule
            \end{tabular}
         }}
   \end{center}
   \begin{center}
      \resizebox{0.8\columnwidth}{!}{

         \setlength{\tabcolsep}{0.35mm}{

            \begin{tabular}{l|c|c|c|c|c|c}
               \toprule
               (b) Shadow Removal                              & MAE($\downarrow$) & PSNR($\uparrow$)  & SSIM($\uparrow$)  & Params(M)         & MACs(G) $\times$ Steps            & Inference Time(s)                \\
               \midrule
               Shadow Diffusion~\citep{guo2023shadowdiffusion} & \textbf{4.12}     & \textbf{32.33}    & \textbf{0.969}    & -                 & -                                 & -                                \\
               SR3 \cite{saharia2022image}  (80k)              & 14.22             & 25.33             & 0.780             & 155.29            & 155.3$\times$100=15530.0          & 0.02$\times$100 = 2.00           \\
               SR3 \cite{saharia2022image} (500K)              & 13.38             & 26.03             & 0.820             & 155.29            & 155.3$\times$100=15530.0          & 0.02$\times$100 = 2.00           \\
               SR3 \cite{saharia2022image} (1000K)             & 11.61             & 27.49             & 0.871             & 155.29            & 155.3$\times$100=15530.0          & 0.02$\times$100 = 2.00           \\
               Ours (only res, 80k)                            & 4.76              & 30.72             & 0.959             & \textbf{7.74}     & \textbf{33.5$\times$5 = 167.7}    & \textbf{0.03$\times$5 = 0.16}    \\
               Ours (80k)                                      & \underline{4.67}  & \underline{30.91} & \underline{0.962} & \underline{15.49} & \underline{67.1$\times$5 = 335.5} & \underline{0.06$\times$5 = 0.32} \\
               \bottomrule
            \end{tabular}

         }}
   \end{center}
   \begin{center}
      \resizebox{0.8\columnwidth}{!}{

         \setlength{\tabcolsep}{1.8mm}{
            \begin{tabular}{l|c|c|c|c|c}
               \toprule
               (c) Deraining   & PSNR($\uparrow$)  & SSIM($\uparrow$)   & Params(M)         & MACs(G) $\times$ Steps            & Inference Time(s)                \\
               \midrule
               RainDiff64[28]  & 32.29             & 0.9422             & 109.68            & 252.4$\times$10 = 2524.2          & 0.03$\times$10 = 0.38            \\
               RainDiff128[28] & \underline{32.43} & 0.9334             & 109.68            & 248.4$\times$50 = 12420.0         & 0.038$\times$50 = 1.91           \\
               Ours (only res) & 31.96             & \underline{0.9509} & \textbf{7.73}     & \textbf{32.9$\times$5 = 164.7}    & \textbf{0.032$\times$5 = 0.16}   \\
               Ours            & \textbf{32.51}    & \textbf{0.9563}    & \underline{15.47} & \underline{65.8$\times$5 = 329.3} & \underline{0.07$\times$5 = 0.35} \\
               \bottomrule
            \end{tabular}
         }}
   \end{center}
   \vspace{-0.3cm}
\end{table}

\begin{figure}[t]
   \setlength{\abovecaptionskip}{0cm}
   \centering
   \includegraphics[width=0.55\linewidth]{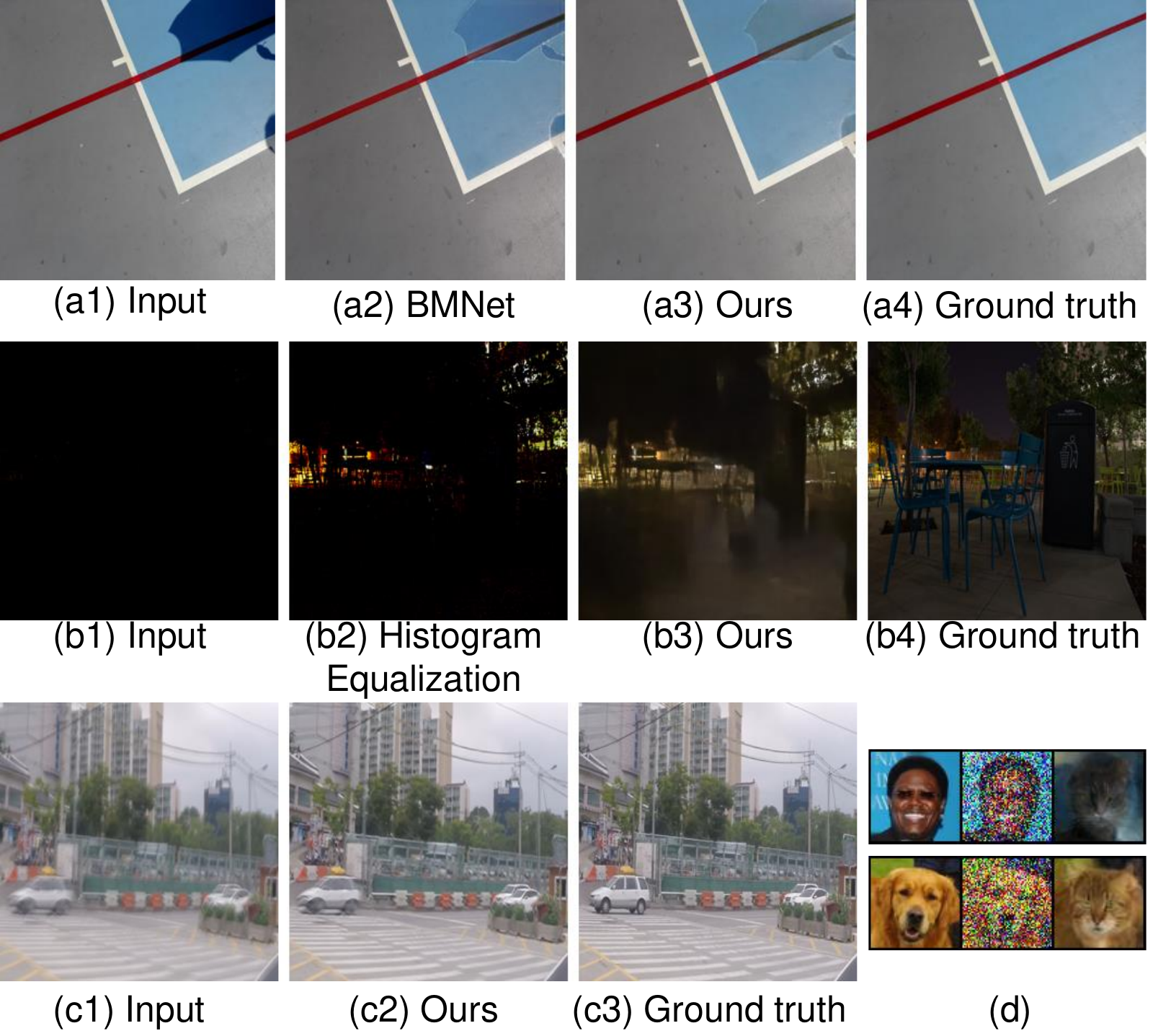}
   \caption{Failure cases. (a1-a4) Shadow removal on the ISTD dataset~\citep{wang2018stacked}. (b1-b4) Low-light enhancement on the SID-RGB dataset~\citep{xu2020learning}. (c1-c3) Deblurring on the GoPro~\citep{nah2017deep} dataset. (d) Image translation (male/dog$\to$cat) on the CelebA-HQ~\citep{karras2017progressive} and AFHQ~\citep{choi2020stargan} datasets.}
   \label{fig:a2}
\end{figure}

{\bf Resource efficiency.} Due to fewer sampling steps, our RDDM inference time and performance is comparable to lllflow~\citep{wang2022low}, and LLFormer~\citep{wang2023ultra} (not diffusion-based). Compared to SR3~\citep{saharia2022image}, our RDDM (only res in Table~\ref{table:Resource}(b)) has 10x fewer training iterations, 10x fewer parameters, 10x faster inference time, and 10\% improvement in PSNR and SSIM on the ISTD~\citep{wang2018stacked} dataset (shadow removal). For a fair comparison, priori shadow masks are used in SR3~\citep{saharia2022image} with a batch size of 1. ShadowDiffusion~\citep{guo2023shadowdiffusion} uses SR3~\citep{saharia2022image} and Uformer~\citep{wang2022uformer}, which has a higher PSNR but is also expected to be more computationally expensive. {\bf Our RDDM with SM-Res requires only 4.8G of GPU memory for training.} Experiments in low-light enhancement, shadow removal, and deraining demonstrate the effectiveness of RDDM, enabling computationally-constrained researchers to utilize diffusion modeling for image restoration tasks.

   {\bf Accelerating Convergence.} The residual prediction in our RDDM helps the diffusion process to be more certain, which can accelerate the convergence process, e.g., fewer training iterations and higher performance in Table~\ref{table:Resource}(b).

{\bf Failure case.} We present some failure cases in Fig.~\ref{fig:a2}.

\section{Discussions, Limitations, and Further Work}\label{Appendix:c.2}
{\bf Limitations.}
Our primary focus has been on developing a unified prototype model for image restoration and generation, which may result in certain performance limitations when compared to task-specific state-of-the-art methods.
To further improve the performance of a specific task, potential avenues for exploration include using a UNet with a larger number of parameters, increasing the batch size, conducting more iterations, and implementing more effective training strategies, such as learning rate adjustments customized for different tasks.
For the image generation task, although Table~\ref{table:alpha_schedules}
showcases the development of an improved coefficient schedule, attaining state-of-the-art performance  in image generation necessitates further investigation and experimentation.
In summary, while we recognize the existing performance limitations for specific tasks, we are confident that our unified prototype model serves as a robust foundation for image restoration and generation.

   {\bf Further Work.} Here are some interesting ways to extend our RDDM.

\begin{enumerate}
   \item In-depth analysis of the relationship between RDDM and curve/multivariate integration.
   \item Development of a diffusion model trained with one set of pre-trained parameters to handle several different tasks.
   \item Implementing adaptive learning coefficient schedules to reduce the sampling steps while improving the quality of the generated images.
   \item Constructing interpretable multi-dimensional latent diffusion models for multimodal fusion, e.g., generating images using text and images as conditions.
   \item Adaptive learning noise intensity ($\beta_T^2$) for an unknown new task.
   \item Exploring residuals in distillation (e.g., introducing dual diffusion into consistency models~\citep{song2023consistency}).
\end{enumerate}



\end{document}